\documentclass{bmvc2k}

\usepackage{amsmath}
\usepackage{amssymb}
\usepackage{array}
\usepackage{multirow}
\usepackage{float}
\usepackage{epstopdf}

\title{Generalized Content-Preserving Warps for Image Stitching}

\addauthor{Kai Chen\\Jingmin Tu\\Jian Yao}{{chenkai,jingmin.tu,jian.yao}@whu.edu.cn}{0}

\addinstitution{
	School of Remote Sensing and Information Engineering\\
	Wuhan University\\
	Wuhan, Hubei Province, China
}

\runninghead{}{Generalized Content-Preserving Warping}

\def\eg{\emph{e.g}\bmvaOneDot}

\def\etal{\emph{et al}\bmvaOneDot}

\def \originalimage{I}
\def \sourceimage{I_s}
\def \targetimage{I_t}

\def \originvertex{\mathbf{V}}
\def \resultvertex{\hat{\mathbf{V}}}
\def \onevertex{\mathbf{v}}
\def \resultonevertex{\hat{\mathbf{v}}}

\begin{document}

\maketitle

\begin{abstract}
    Local misalignment caused by global homography is a common issue in image stitching task. Content-Preserving Warping (CPW) is a typical method to deal with this issue, in which geometric and photometric constraints are imposed to guide the warping process. One of its essential condition however, is colour consistency, and an elusive goal in real world applications. In this paper, we propose a Generalized Content-Preserving Warping (GCPW) method to alleviate this problem. GCPW extends the original CPW by applying a colour model that expresses the colour transformation between images locally, thus meeting the photometric constraint requirements for effective image stitching. We combine the photometric and geometric constraints and jointly estimate the colour transformation and the warped mesh vertexes, simultaneously. We align images locally with an optimal grid mesh generated by our GCPW method. Experiments on both synthetic and real images demonstrate that our new method is robust to colour variations, outperforming other state-of-the-art CPW-based image stitching methods.
\end{abstract}

\vspace{-15pt}
\section{Introduction}
\label{sec:introduction}

Given two images capturing a same scene, the objective of image stitching is to combine them into a panorama with a wider field of view (FOV). At the same time, the stitching result must be free from obvious misalignment artifacts. In order to achieve this purpose, Brown and Lowe~\cite{brown2007automatic} proposed to stitch images through global homography, which was estimated by Direct Linear Transform (DLT) using matched keypoints in an overlapping region. Li~\etal~\cite{li2015dual} utilized dual-feature correspondences to estimate global homography, which achieved a more accurate and more robust estimation result. Chang~\etal~\cite{chang2017clkn} proposed to estimate the global homography by deep learning. They adopted features learned from convolutional neural network (CNN), and designed a neural network layer to perform the inverse compositional Lucas-Kanade algorithm~\cite{baker2004lucas} to estimate the homography parameters. Methods based on global homography however, are prone to suffer from misalignment artifacts, as these methods assume that scenes are located on a single plane or the motion of camera is pure rotation, conditions that are hard to meet in many practical applications.

\emph{Content-Preserving Warping} (CPW) methods therefore were proposed to overcome the shortcoming of methods using global homography. The CPW framework was first proposed by Liu and Jin~\cite{liu2009content} for the purpose of video stabilization and then widely applied in image stitching. These methods formulated the local alignment of images as a mesh deformation problem and applied various types of constraints to guide the mesh deformation process. Generally, two types of constraints were considered: geometric and photometric constraints. CPW methods can be divided into two categories based on the type of constraint.

\begin{figure}[h]
	\centering
        \includegraphics[width=1.0\textwidth]{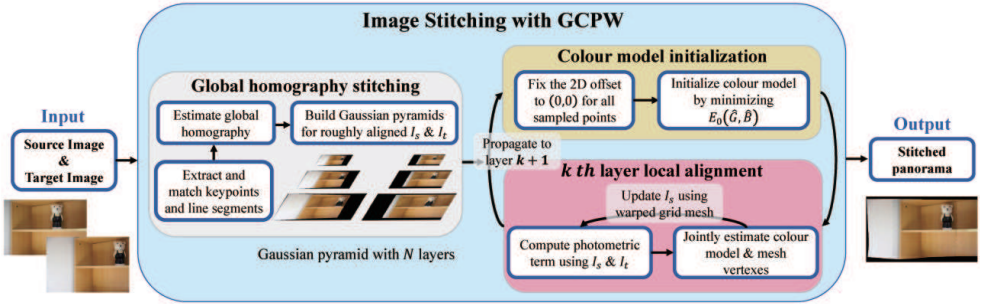}
	\caption{An overview of our image stitching method with GCPW. From given image pair, we first roughly align them with a global homography. Next, we combine photometric and geometric constraints in the GCPW framework and design a robust optimization scheme to estimate the colour model and the mesh vertexes, simultaneously. At last, images are locally aligned based on the optimized grid mesh to create a stitched panorama.}
	\label{fig:flowchart}\vspace{-10pt}
\end{figure}

For methods with geometric constraint, research~\cite{zhang2014parallax,hu2015multi,lin2016seagull} detected and matched keypoints in an overlapping region and constrained each pair of matched keypoints to be warped to the same position during mesh deformation. Guo~\etal~\cite{guo2016joint} proposed self-adaptive thresholds to detect keypoints in order to collect sufficient amount of point correspondences for adequate warping guidance. Similarly, Zhang~\etal~\cite{zhang2016multi} designed a local-homography-based outlier rejection procedure to preserve point correspondences located on different planes, while Chen and Chuang~\cite{chen2016natural} took the aligned mesh vertexes generated by APAP~\cite{zaragoza2013projective} as uniformly distributed matched keypoints in an overlapping region. These methods however, still fail when scenes were lack of textures. To deal with this problem, Li~\etal~\cite{li2015dual} and Xiang~\etal~\cite{xiang2017image} resorted to line segment correspondences. They combined point and line constraints in a CPW framework to obtain better stitching results than methods using only point constraints.

Geometric constraints however, are not robust enough. Although the combination of keypoints and line segments has significantly improved alignment quality, these two kinds of geometric constraints have limitations. For one, keypoints are often distributed with spatial bias in an overlapping region~\cite{joo2015line}. For another, the extraction of line segments is difficult in images with small structures or tiny gradients.

Lin~\etal~\cite{lin2017direct} therefore proposed to introduce photometric constraint into the CPW framework to stitch images. They sampled a set of points at a fixed interval in an overlapping region and required overall photometric differences of all sampled points to be minimized by mesh deformation. This process is just similar to the one of optical flow estimation~\cite{fortun2015optical}, which usually requires images to obey the colour consistency assumption~\cite{demetz2014learning}. Images with significant brightness or colour variations are common however, which would be resulted by complex scenes or various illumination conditions. Under these circumstances, we must correct colour differences before adopting a photometric constraint to stitch images. Many techniques can be applied to compensate image colour differences. These includes: invariant feature extraction~\cite{he2013visual} and colour consistency correction~\cite{yao2017color}, but these methods usually work as pre-processing operations, and may fail to correct the differences thoroughly or may deteriorate the original image content, which would affect the effectiveness of the photometric constraint and produce poor stitching results.

In this paper, we propose a \emph{Generalized Content-Preserving Warping} (GCPW) method to stitch images with colour variations. The proposed GCPW extends the original CPW by appending an affine colour model that expresses the colour transformation from a source image to a target image locally. Our extension to CPW makes the photometric constraint robust to significant colour variations. Figure~\ref{fig:flowchart} presents an overview of our image stitching scheme incorporating GCPW. We first roughly align images with a global homography that is estimated from dual features~\cite{li2015dual}. Next, we locally align them using the proposed GCPW. Specifically, we combine geometric and photometric constraints in the GCPW framework. Furthermore, we designed a robust optimization procedure that jointly estimates the affine colour model and the warped mesh vertexes. At last, images are locally aligned according to the optimized grid mesh. We tested our proposed method on both synthetic and real images. The results from these experiments show that the proposed method is robust to colour variations, producing better alignment quality than other state-of-the-art methods.

\vspace{-5pt}
\section{Image Stitching with GCPW}
\label{sec:Stitching}\vspace{-10pt}
Given two images $\originalimage_1$ and $\originalimage_2$, we first roughly align them with a global homography to obtain $\sourceimage$ and $\targetimage$. In the classical CPW framework, the local alignment of $\sourceimage$ and $\targetimage$ is then formulated into a problem of mesh deformation, in which various geometric and photometric constraints are imposed to align images and preserve image content at the same time. Lin~\etal~\cite{lin2017direct} demonstrated that the photometric constraints can improve the alignment quality, but this method requires that images obey the colour consistency assumption. We therefore propose the GCPW framework (Section~\ref{subsec:GCPW}), and design a photometric constraint beyond colour consistency (Section~\ref{subsec:PhotoError}). We combine the proposed photometric constraint with geometric constraints in the GCPW framework to locally align images by mesh deformation (Section~\ref{subsec:MeshDeformation}). Considering the high degree of freedom (DoF) of our proposed GCPW model, we designed an optimization scheme to estimate the colour model and the mesh vertexes simultaneously (Section~\ref{subsec:Optimization}).

\vspace{-5pt}
\subsection{The Proposed GCPW}
\label{subsec:GCPW}
\begin{figure}[h]
	\centering
		\begin{minipage}[t]{0.48\columnwidth}
			\includegraphics[width=1.0\textwidth]{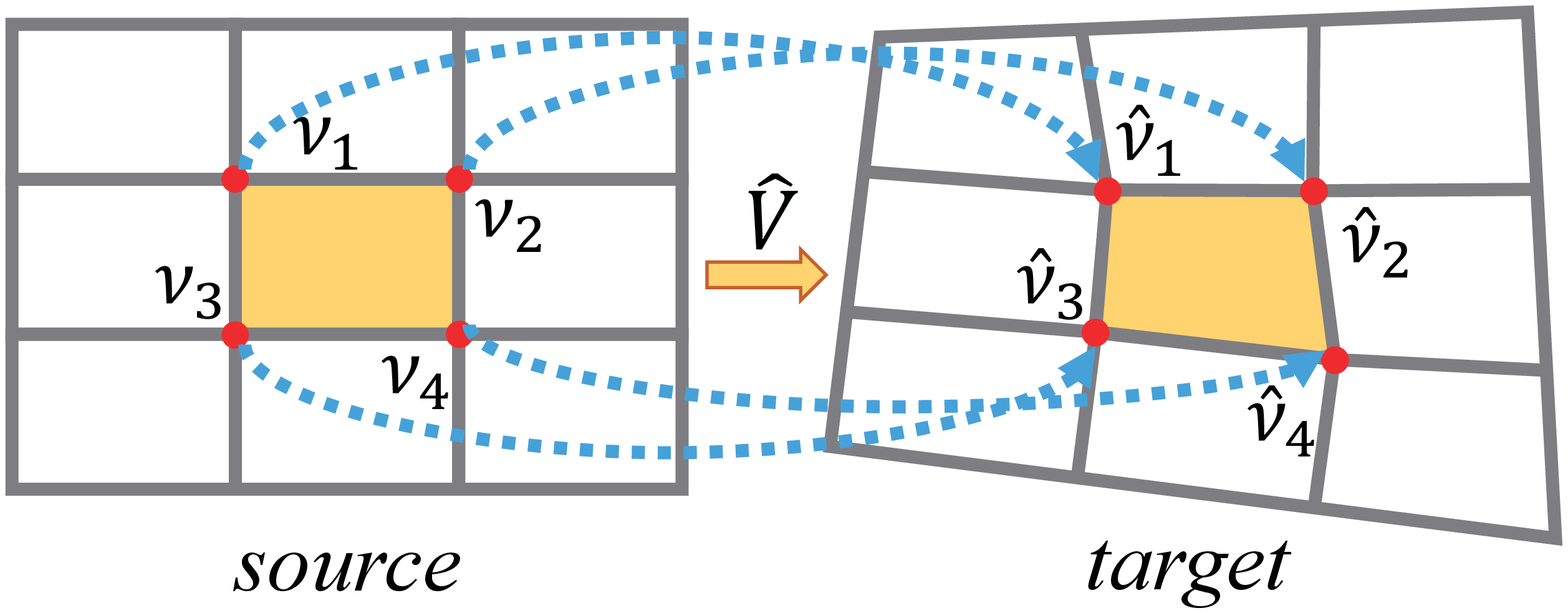}
            \centering{(a) The CPW framework~\cite{liu2009content}}
		\end{minipage}
		\hfill
		\begin{minipage}[t]{0.48\columnwidth}
			\includegraphics[width=1.0\textwidth]{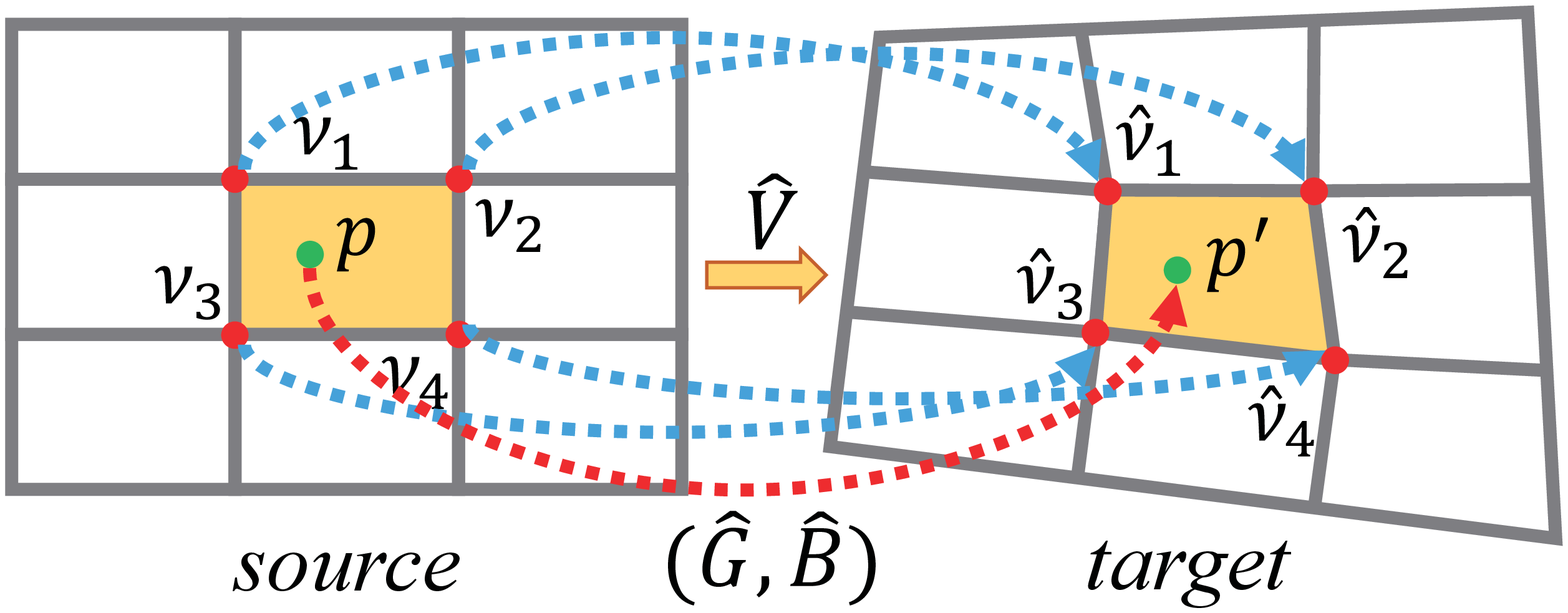}
            \centering{(b) The proposed GCPW framework}
		\end{minipage}
	\caption{A comparison between CPW and GCPW. (a) The original CPW only estimates warped mesh vertexes $\resultvertex$. (b) Our proposed GCPW jointly estimates warped mesh vertexes $\resultvertex$ and the local colour model $(\hat{\mathbf{G}},\hat{\mathbf{B}})$. The colour model in GCPW expresses the colour transformation from source image to traget image (\eg, from $p$ to $p^{\prime}$).}
	\label{fig:GCPW}\vspace{-12pt}
\end{figure}

\vspace{-7pt}
In this section, we introduce the proposed GCPW model. Figure~\ref{fig:GCPW} shows a comparison between CPW and GCPW. Specifically, $\sourceimage$ is first divided into an $m\times n$ uniform grid mesh ($m=n=32$ in this paper), whose vertexes are denoted as $\originvertex$. In contrast to CPW, which only estimates warped mesh vertexes $\resultvertex$, our proposed GCPW optimizes $\resultvertex$ and the local colour transformation from $\sourceimage$ to $\targetimage$ ($\hat{\mathbf{G}}$,$\hat{\mathbf{B}}$) at the same time. We assume that pixels within a quad share a same colour transformation. Similar to~\cite{hacohen2011non}, we use an affine model to express the colour transformation from $\sourceimage$ to $\targetimage$ within a quad, where $\hat{\mathbf{G}}$ and $\hat{\mathbf{B}}$ denote the colour gain and the colour bias respectively. For typical \emph{RGB} images, we convert them into $YCbCr$ colour space and compute colour models for each image channel separately.

The proposed GCPW typically has two advantages. Firstly, it considers the colour transformation from $\sourceimage$ to $\targetimage$. Therefore, it measures the photometric error normally even when $\sourceimage$ and $\targetimage$ have significant colour differences. Secondly, the photometric error computed by GCPW retains the quadratic property, which ensures that the photometric constraint is easily combined with other widely used constraints to align images locally.


\vspace{-5pt}
\subsection{Photometric Constraints Beyond Colour Constancy}
\label{subsec:PhotoError}\vspace{-7pt}

For two roughly aligned images $\sourceimage$ and $\targetimage$, our objective is to locally align them by warping $\sourceimage$ to $\targetimage$ via mesh deformation. Lin~\etal~\cite{lin2017direct} proved that photometric constraint is an important complementary to guide this process. The required colour constancy condition however, is not always satisfied in practice. We therefore propose a photometric constraint beyond colour consistency. Specifically, we uniformly sample a set of points at a fixed interval (three pixels in this paper) both horizontally and vertically in an overlapping region. For each sampled point $\mathbf{q}$ in $\sourceimage$, we want to find a $2D$ offset $\tau(\mathbf{q})$ that minimizes the photometric difference between $\mathbf{q}$ in $\sourceimage$ and $\mathbf{q}+\tau(\mathbf{q})$ in $\targetimage$. Considering the colour variation between $\sourceimage$ and $\targetimage$, we measure their photometric error as:
\vspace{-3pt}
\begin{equation}\label{eq:photoerror1}
    \|(\mathbf{G}^i(\mathbf{q})\times \sourceimage^i(\mathbf{q})+\mathbf{B}^i(\mathbf{q}))-\targetimage^i(\mathbf{q}+\tau(\mathbf{q}))\|^2,\vspace{-3pt}
\end{equation}
where $i$ indexes the image channel. $\sourceimage^i(\mathbf{q})$ is the intensity of $\sourceimage^i$ at $\mathbf{q}$, and $\targetimage^i(\mathbf{q}+\tau(\mathbf{q}))$ is the intensity of $\targetimage^i$ at $\mathbf{q}+\tau(\mathbf{q})$. $\mathbf{G}^i(\mathbf{q})$ and $\mathbf{B}^i(\mathbf{q}))$ denote the colour gain and bias of the quad that $\mathbf{q}$ is located. Since $\sourceimage$ and $\targetimage$ are roughly aligned, we assume that $\tau(\mathbf{q})$ is small and set its initial value as $(0,0)$. Using the first-order Taylor expansion of $\targetimage^i(\mathbf{q}+\tau(\mathbf{q}))$, the photometric error defined by Eq.~\ref{eq:photoerror1} can be expressed as:
\vspace{-3pt}
\begin{equation}\label{eq:photoerror2}
    E_c(\tau(\mathbf{q}))=\|(\mathbf{G}^i(\mathbf{q})\times \sourceimage^i(\mathbf{q})+\mathbf{B}^i(\mathbf{q}))-(\targetimage^i(\mathbf{q})+\nabla \targetimage^i(\mathbf{q})\tau(\mathbf{q}))\|^2,\vspace{-3pt}
\end{equation}
where $\nabla \targetimage^i(\mathbf{q})$ is approximated by image gradient of $\targetimage^i$ at $\mathbf{q}$.

\vspace{-5pt}
\subsection{Local Alignment by Mesh Deformation}
\label{subsec:MeshDeformation}\vspace{-7pt}
In the proposed GCPW framework, we combine the proposed photometric constraint with other widely used constraints to align images locally, and define the objective function as follows:
\vspace{-3pt}
\begin{equation}\label{eq:totalfunction}
    E(\resultvertex,\hat{\mathbf{G}},\hat{\mathbf{B}})=\lambda_1 E_p(\resultvertex,\hat{\mathbf{G}},\hat{\mathbf{B}})+\lambda_2 E_g(\resultvertex)+\lambda_3 E_{s}(\resultvertex)+\lambda_4 E_c(\hat{\mathbf{G}},\hat{\mathbf{B}}),\vspace{-3pt}
\end{equation}
where $E_p$ is the proposed photometric constraint, and $E_g$ is the geometric constraint and similar to the one in~\cite{li2015dual}. $E_s$ and $E_c$ are smoothness constraints, in which $E_s$ is used to constrain the similarity transformation of the warped mesh~\cite{liu2009content} and $E_c$ is the term designed to preserve the smoothness of the local colour model in GCPW. The associated weights are denoted by $\lambda_1$, $\lambda_2$, $\lambda_3$ and $\lambda_4$ respectively. ($\lambda_1=100.0$, $\lambda_2=\lambda_4=1.0$ and $\lambda_3=0.5$ in our implementation).


\noindent\textbf{Photometric term}\quad For each sampled point $\mathbf{q}$, we express its corresponding photometric error according to Eq.~\ref{eq:photoerror2}. In order to associate it with mesh deformation, we further parameterize the $2D$ offset $\tau(\mathbf{q})$ by:
\vspace{-3pt}
\begin{equation}\label{eq:sampleoffset}
    \tau(\mathbf{q})=\hat{\mathbf{q}}-\mathbf{q},\hat{\mathbf{q}}=\begin{matrix}\sum_{k=1}^{4}w_k\resultonevertex_{\mathbf{q}}^{k}\end{matrix},\vspace{-3pt}
\end{equation}
where $\hat{\mathbf{q}}$ is the warped position of $\mathbf{q}$, $w_k$ are bilinear interpolation coefficients computed by expressing $\mathbf{q}$ with four initial mesh vertexes $\onevertex_{\mathbf{q}}^{k}$ that enclose $\mathbf{q}$, and $\resultonevertex_{\mathbf{q}}^{k}$ are the warped positions of $\onevertex_{\mathbf{q}}^{k}$. $E_p(\resultvertex,\hat{\mathbf{G}},\hat{\mathbf{B}})$ is computed by summing up $E_c(\tau(\mathbf{q}))$ over all sampled points and all image channels:
\vspace{-3pt}
\begin{equation}\label{eq:phototerm}
    E_p(\resultvertex,\hat{\mathbf{G}},\hat{\mathbf{B}})=\begin{matrix}
                       \sum_{i=1}^{3}\sum_{\mathbf{q}}E_c(\tau(\mathbf{q}^i))
                       \end{matrix},\vspace{-3pt}
\end{equation}

\noindent\textbf{Geometric term}\quad Similar to~\cite{li2015dual}, we adopt geometric constraint based on dual-feature correspondences. Thus, our geometric term can be divided into the point component $E_f(\resultvertex)$ and the line component $E_l(\resultvertex)$.

For each pair of matched keypoints ($\mathbf{p}$, $\mathbf{p}^\prime$), we restrict the warped position of $\mathbf{p}$ to being close to $\mathbf{p}^\prime$, which is achieved by minimizing $D_p=\|\hat{\mathbf{p}}-\mathbf{p}^\prime\|^2$. For each pair of matched line segment $\mathbf{l}=[a,b,c]$ and $\mathbf{l}^\prime=[a^\prime,b^\prime,c^\prime]$, we uniformly sample keypoints along $\mathbf{l}$ at a fixed interval (five pixels in this paper) and stipulate that the total distance from these keypoints to $\mathbf{l}^\prime$ is minimized. We traverse all pairs of keypoints and line segments to get the point component $E_f(\resultvertex)$ and line component $E_l(\resultvertex)$. More details about $E_f(\resultvertex)$ and $E_l(\resultvertex)$ can be found in~\cite{li2015dual}. The overall geometric term is:
\vspace{-3pt}
\begin{equation}\label{eq:geometerm}
    E_g(\resultvertex)=E_f(\resultvertex)+E_l(\resultvertex).\vspace{-3pt}
\end{equation}

\noindent\textbf{Similarity transformation term}\quad The same similarity transformation constraint~\cite{liu2009content} is used in our implementation. Specifically, each quad $[\onevertex_1,\onevertex_2,\onevertex_3,\onevertex_4]$ is divided into two triangulations $\triangle\onevertex_1\onevertex_2\onevertex_3$ and $\triangle\onevertex_2\onevertex_3\onevertex_4$. We express $\onevertex_1$ of $\triangle\onevertex_1\onevertex_2\onevertex_3$ in the local coordinate system defined by $\onevertex_2$ and $\onevertex_3$ as follows:
\vspace{-7pt}
\begin{equation}\label{eq:st1} \onevertex_1=\onevertex_2+u(\onevertex_3-\onevertex_2)+v\mathbf{R}_{90}(\onevertex_3-\onevertex_2),\quad \mathbf{R}_{90}=\begin{bmatrix}
	0 & 1 \\ -1 & 0
	\end{bmatrix},\vspace{-7pt}
\end{equation}
where $u$ and $v$ are computed from original positions of $\onevertex_1$, $\onevertex_2$ and $\onevertex_3$. We require that $\onevertex_1$ be represented by $\onevertex_2$ and $\onevertex_3$ using the same local coordinates $(u,v)$ before and after warping. The overall similarity transformation term is defined as:
\vspace{-3pt}
\begin{equation}\label{eq:similarityterm}
	E_{s}(\resultvertex)=\begin{matrix}
	\sum_{i}^{N}\|\resultonevertex_1^{i}-(\resultonevertex_2^i+u(\resultonevertex_3^i-\resultonevertex_2^i)+v\mathbf{R}_{90}(\resultonevertex_3^i-\resultonevertex_2^i)) \|^2
	\end{matrix},\vspace{-6pt}
\end{equation}
where $N$ is the total number of triangulations in the grid mesh.\vspace{3pt}

\noindent\textbf{Colour smoothness term}\quad A colour smoothness term is designed to constrain the smoothness of colour models within an eight-connected neighboring region. Let $Q$ be a quad of the grid mesh, and $Q^\prime$ be one of its neighboring quads. Let $y=gx+b$ and $y=g^\prime x+b^\prime$ be their affine colour models on one image channel. We sample a set of intensity values in normalized intensity range (\eg, sample values within $[0,1]$ at an interval of $0.1$), and require the sampled values to remain close after the neighboring two affine transformations, which is achieved by minimizing $D_c=\sum_i\|(gx_i+b)-(g^\prime x_i+b^\prime)\|^2$, where $x_i$ denotes sampled intensity value.

For each quad of the grid mesh, we constrain the similarities between its colour model with those of all neighboring quads over three image channels. The colour smoothness term is defined as:
\vspace{-4pt}
\begin{equation}\label{eq:colorterm}
    E_c(\hat{\mathbf{G}},\hat{\mathbf{B}})=\begin{matrix}
                 \sum_{n=1}^{3}\sum_{i}\sum_{j\in \Omega_i}D_c(i,j,n)
             \end{matrix},\vspace{-4pt}
\end{equation}
where $n$ indexes image channels, $\Omega_i$ denotes the eight-connected neighborhood of quad $Q_i$.\vspace{3pt}

\vspace{-13pt}
\subsection{Robust Optimization}
\label{subsec:Optimization}\vspace{-7pt}
Since the proposed image stitching method with GCPW estimates warped mesh vertexes $\resultvertex$ and a local colour model ($\hat{\mathbf{G}}$, $\hat{\mathbf{B}}$) at the same time, the high degree of freedom (DoF) makes the optimization result unstable. Therefore, a two-step optimization pipeline is designed: We first initialize ($\hat{\mathbf{G}}$, $\hat{\mathbf{B}}$) using sampled points. After that, $\resultvertex$ and ($\hat{\mathbf{G}}$, $\hat{\mathbf{B}}$) are jointly estimated by minimizing Eq.~\ref{eq:totalfunction}. As for the colour model initialization, we fix the $\tau(\mathbf{q})$ in Eq.~\ref{eq:phototerm} at $(0,0)$ and initialize the colour model by minimizing the following function:
\vspace{-7pt}
\begin{equation}\label{eq:initialfunction}
    E_0(\hat{\mathbf{G}},\hat{\mathbf{B}})=E_p(\hat{\mathbf{G}},\hat{\mathbf{B}})+ E_c(\hat{\mathbf{G}},\hat{\mathbf{B}})+E_r(\hat{\mathbf{G}},\hat{\mathbf{B}}),\vspace{-7pt}
\end{equation}
where $E_p$ and $E_c$ have been defined in Eq.~\ref{eq:phototerm} and Eq.~\ref{eq:colorterm}. $E_r$ is an regular term that we add for initialization, and defined as: $\sum_{Q_i\cap (\sourceimage\cap \targetimage)=\varnothing}\|(\hat{\mathbf{G}}(Q_i)-1.0)\|^2+\|\hat{\mathbf{B}}(Q_i)-0.0\|^2$. $E_r$ restricts the colour gain to being close to $1.0$ and the colour bias to being close to $0.0$ when the quad $Q_i$ is not overlapped with an overlapping region in the image.

Both the initialization function $E_0(\hat{\mathbf{G}},\hat{\mathbf{B}})$ and the final objective function $E(\resultvertex,\hat{\mathbf{G}},\hat{\mathbf{B}})$ are quadratic and can be easily minimized by any sparse linear system. In order to handle images with large displacements, we built a three-layer Gaussian pyramid and adopted a coarse-to-fine optimization scheme. For each layer of the pyramid, as shown in Figure~\ref{fig:flowchart}, we minimize $E_0(\hat{\mathbf{G}},\hat{\mathbf{B}})$ to initialize the colour model, and then iteratively optimize $E(\resultvertex,\hat{\mathbf{G}},\hat{\mathbf{B}})$. As the grid mesh gets updated after each iteration, the image content of $\sourceimage$ is re-rendered according to the updated mesh. The iteration terminates when the grid mesh becomes stable. We warp $\sourceimage$ based on the optimized grid mesh to locally align $\sourceimage$ and $\targetimage$.

\vspace{-10pt}
\section{Experiments and Results}
\label{sec:experiment}\vspace{-5pt}
In this section, we conduct experiments on both synthetic and real images to demonstrate the superiority of our proposed image stitching method with GCPW. All experiments are conducted with default parameter settings as discussed in previous sections of this paper. For convenience, in these comparative experiments, we do not adopt any post-processing methods but simply blend the warped $\sourceimage$ and $\targetimage$ by the intensity average to obtain the final stitching result.

\vspace{-10pt}
\subsection{Experiments on Synthetic Images}
\label{subsec:experimentA}\vspace{-5pt}
We performed experiments on synthetic images with different degrees of colour variation to demonstrate the effectiveness and robustness of our proposed image stitching method. We collected images that have nearly no colour differences, which are obtained from publicly available datasets~\cite{zaragoza2013projective,li2015dual,chen2016natural} or captured by ourselves. In order to synthesize image pairs with different colour variations, for each pair of images, we used a colour model similar to the model in~\cite{park2016efficient} to generate $16$ pairs of images, which explicitly have different degrees of colour variation (from $1$ to $16$, the colour difference varies from smallest to largest). More details about this generation process are presented in the supplementary material.

We stitched these synthetic image pairs with our method and compared the results with those produced by three different methods: DF-W, LSH+MF-W, and ROS+MF-W. The dual-feature warping (DF-W) method~\cite{li2015dual} locally aligns $\sourceimage$ and $\targetimage$ using dual-feature correspondences, and we tested it with our own re-implementation. For the second and third tested methods, we combined the photometric constraint proposed in~\cite{lin2017direct} with DF-W in the CPW framework to get the multiple-feature warping (MF-W) method. As the photometric constraint in~\cite{lin2017direct} requires images to obey colour consistency assumption that are not satisfied in our synthetic images, we therefore tested MF-W cooperated with two types of pre-processing operations to get our another two tested methods. In the first operation, we detected invariant features using the locality sensitive histogram (LSH)~\cite{he2013visual} and extracted photometric constraint based on LSH feature images. We refer to the second tested method as LSH+MF-W. In the second operation, we performed colour consistency correction by the colour re-mapping optimization scheme (ROS)~\cite{yao2017color} and imposed photometric constraint using corrected images. We refer to the third tested method as ROS+MF-W.

In the experiment, we use the photometric term as the only variable, and fix point and line correspondences among image pairs in different degrees of colour variations. In order to measure alignment quality quantitatively, we use the same error metric as in~\cite{li2015dual,lin2017direct}, which computes the RMSE of one minus normalized cross correlation over a $W \times W$ window ($W=5$ in this paper) for pixels in the overlapping region. As the quality measurement is relevant to specific intensities of image content, for the sake of fairness, all alignment error values were computed based on original images. We estimated the optimal grid mesh using the synthetic images and warped the original images with the optimized grid mesh to compute the alignment error.

\begin{figure}[t]
	\centering
	\begin{minipage}[t]{1.0\columnwidth}
		\begin{minipage}[t]{0.32\columnwidth}
			\includegraphics[width=1.0\textwidth]{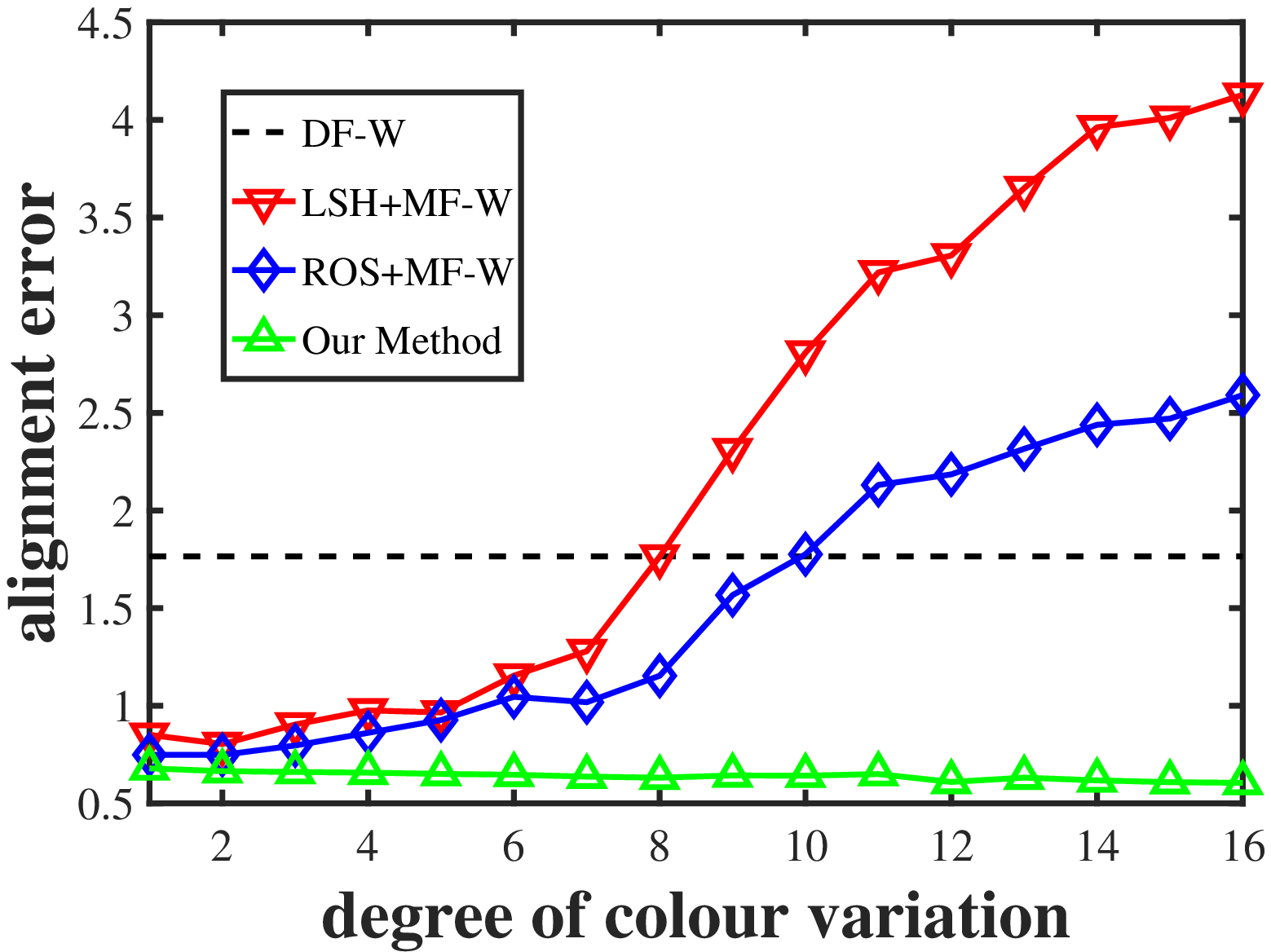}
		\end{minipage}
		\hfill
		\begin{minipage}[t]{0.32\columnwidth}
			\includegraphics[width=1.0\textwidth]{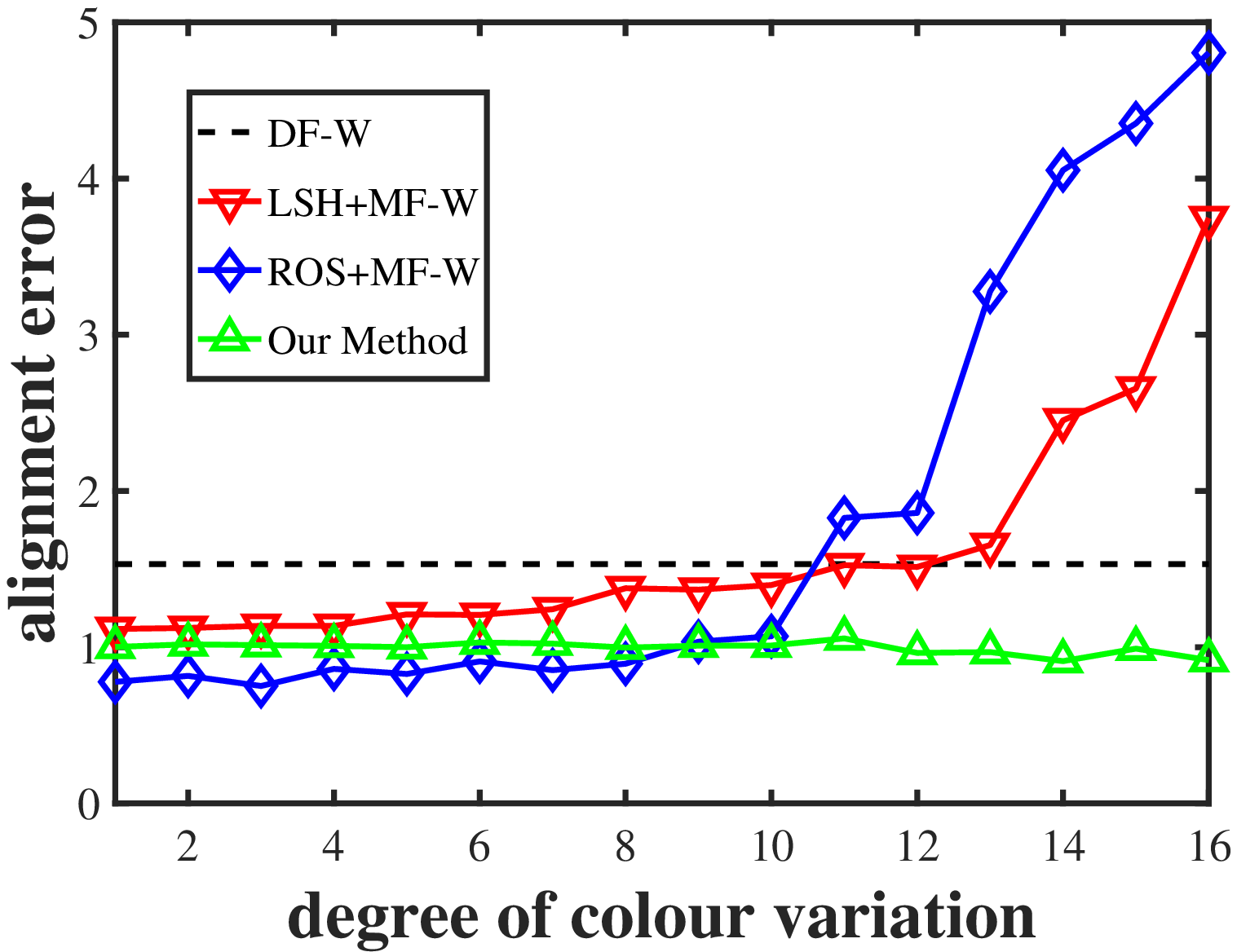}
		\end{minipage}
		\hfill
		\begin{minipage}[t]{0.32\columnwidth}
			\includegraphics[width=1.0\textwidth]{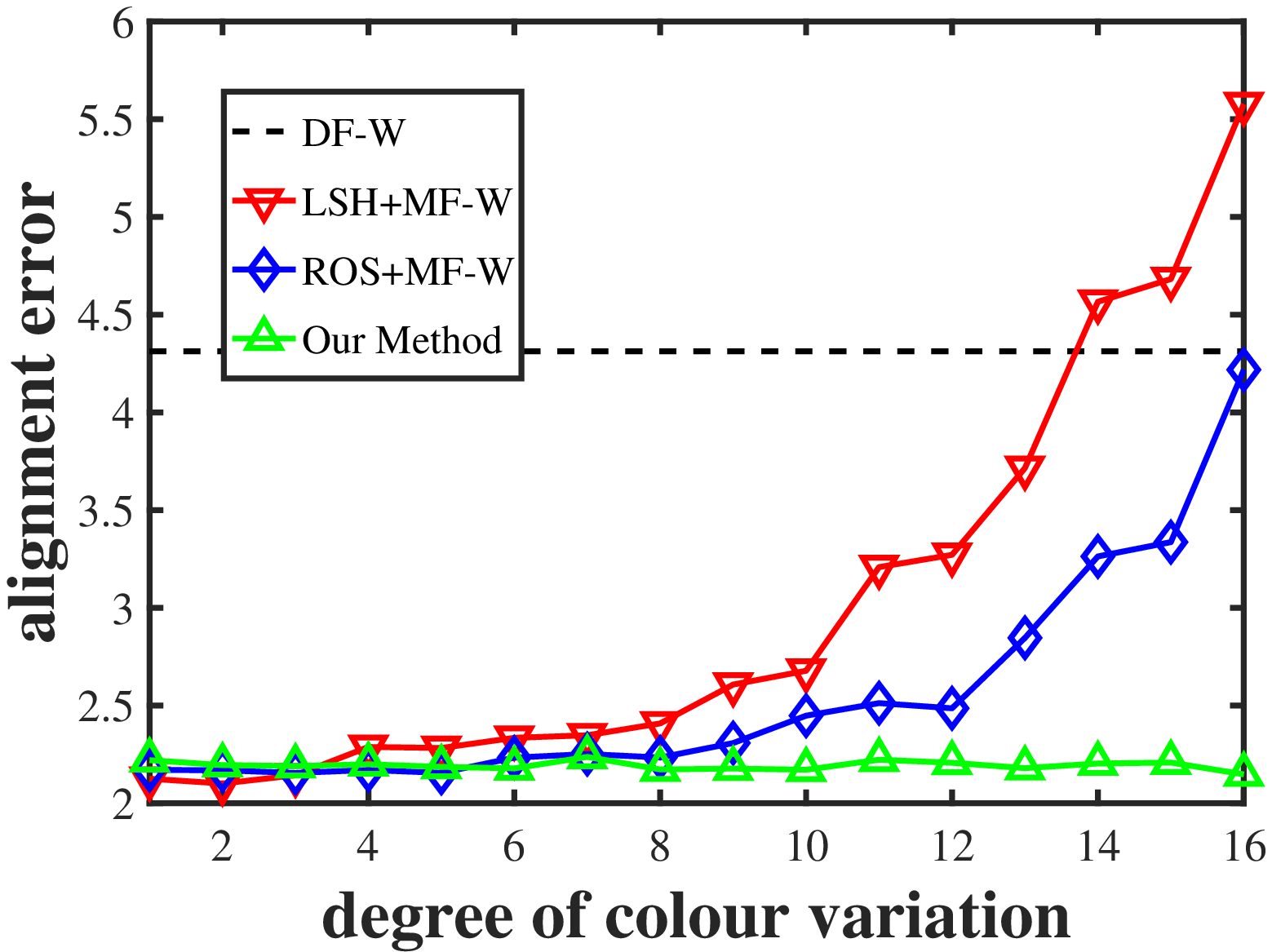}
		\end{minipage}
	\end{minipage}
	\vfill
	\begin{minipage}[t]{1.0\columnwidth}
		\begin{minipage}[t]{0.32\columnwidth}
			\includegraphics[width=1.0\textwidth]{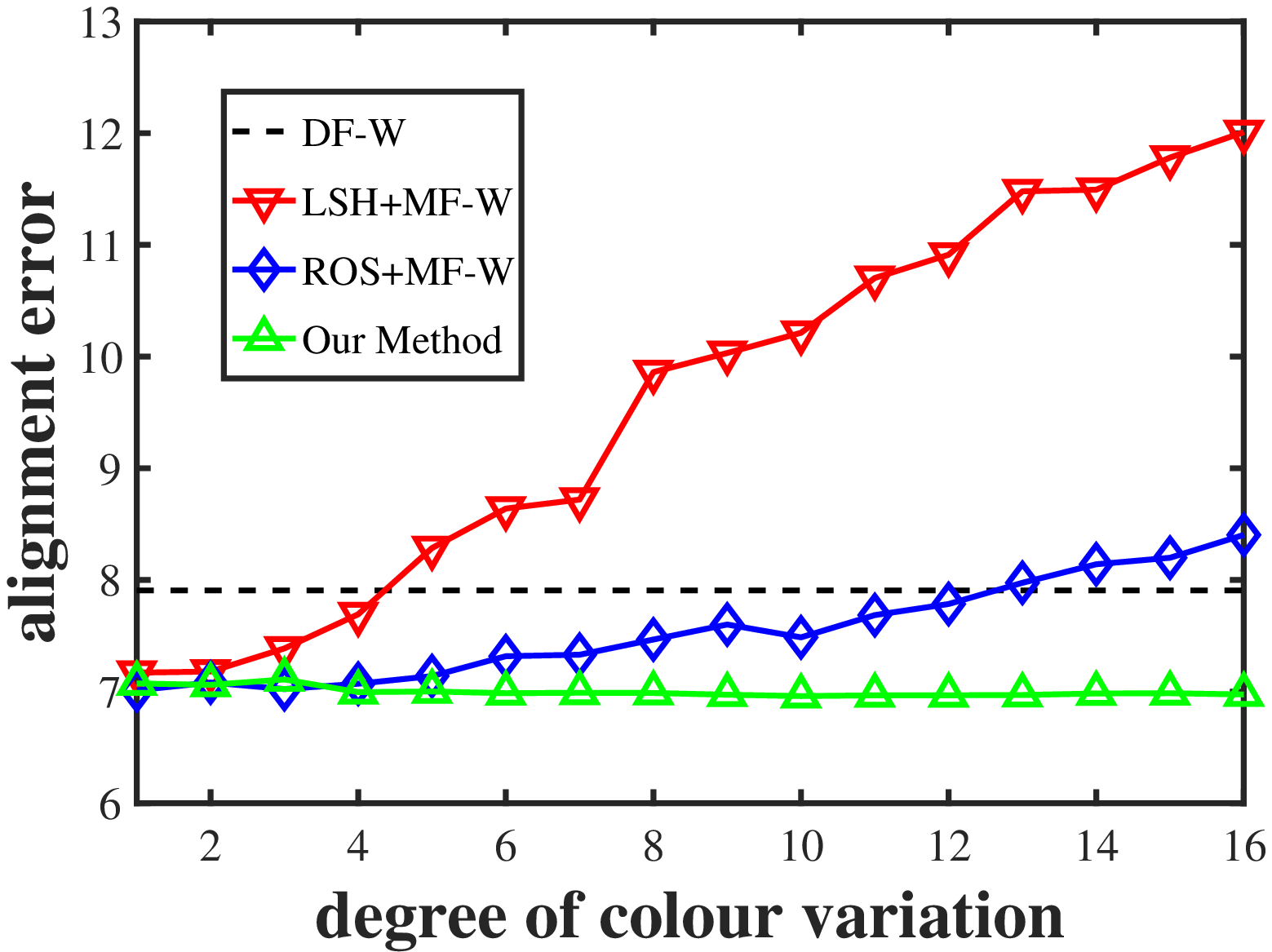}
		\end{minipage}
		\hfill
		\begin{minipage}[t]{0.32\columnwidth}
			\includegraphics[width=1.0\textwidth]{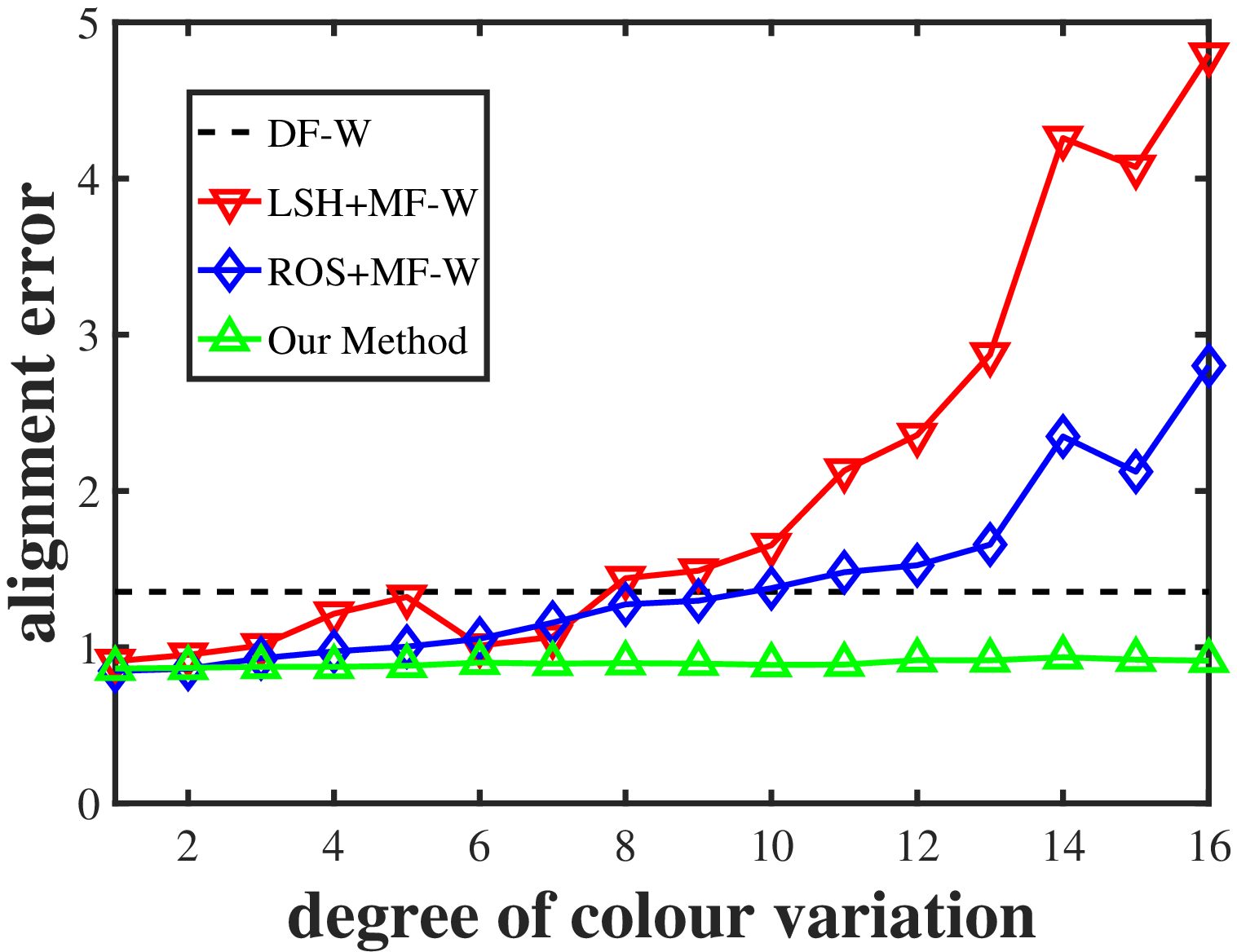}
		\end{minipage}
		\hfill
		\begin{minipage}[t]{0.32\columnwidth}
			\includegraphics[width=1.0\textwidth]{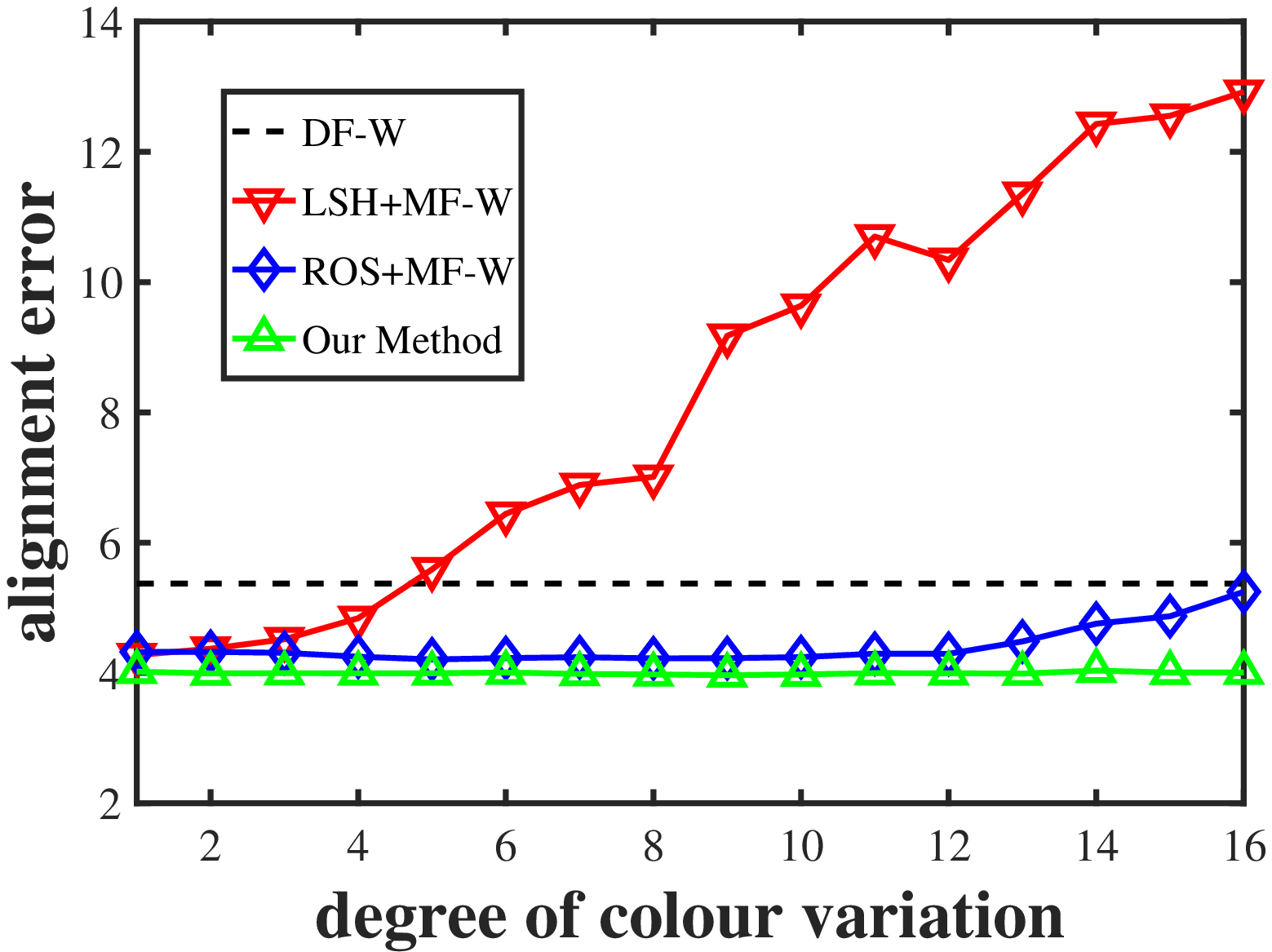}
		\end{minipage}
	\end{minipage}
    \vspace{-15pt}
	\caption{Comparisons on alignment errors of stitching results produced by different methods on synthetic images with different degrees of colour variations.}
	\label{fig:result_experimentA}\vspace{-10pt}
\end{figure}

Figure~\ref{fig:result_experimentA} presents six groups of comparative results, and more results are presented in our supplementary material. When colour difference is modest, LSH+MF-W, ROS+MF-W and our method usually produce fewer errors than DF-W, demonstrating that the photometric constraint is beneficial to better alignment quality. However, the errors in LSH+MF-W and ROS+MF-W results become larger as the colour difference increases, as LSH and ROS fail to compensate colour differences between images when the colour variation becomes larger and more complex. In contrast, as the degree of colour variation varies from the smallest to the largest, the alignment error in our method are stable, demonstrating that our proposed method is robust to significant colour differences. Our proposed method combines photometric and geometric constraints to produce better stitching results than DF-W, and it stably obtains good alignment results even for images with complex colour variations.
\vspace{-5pt}
\subsection{Experiments on Real Images}
\label{subsec:experimentB}

We conducted experiments on real images with significant colour differences, which were directly collected from publicly available datasets~\cite{zaragoza2013projective,li2015dual,chen2016natural}. In the experiment, The proposed method was compared with three geometric-based methods (global homography, CPW~\cite{liu2009content}, and DF-W~\cite{li2015dual}) as well as three photometric-based methods (MF-W, LSH+MF-W, and ROS+MF-W). Table~\ref{table:result} presents the alignment error results on ten pairs of images. The proposed GCPW method usually yielded the lowest alignment error values.
\vspace{-2pt}
\begin{table}[h]
\centering
\begin{tabular}{|c|c|c|c|c|c|c|c|}
\hline
& \multicolumn{3}{c|}{only geometric constraint} & \multicolumn{4}{c|}{geometric and photometric constraints}\\
\hline
method     & Global & CPW & DF-W & MF-W & LSH+MF-W & ROS+MF-W & \textbf{GCPW}\\
\hline
\hline
\textbf{01}      & 12.26        & 12.22    & 11.54    & 16.41    & 10.73        & 8.40         & \textbf{5.61}     \\ \cline{1-8}
\textbf{02}      & 7.60         & 9.41     & 4.19     & 5.57     & 4.10         & 3.18         & \textbf{1.71}     \\ \cline{1-8}
\textbf{03}      & 10.42        & 9.66     & 4.71     & 10.70    & 4.29         & 3.80         & \textbf{3.32}     \\ \cline{1-8}
\textbf{04}      & 10.26        & 7.56     & 5.17     & 8.23     & 4.64         & 4.16         & \textbf{3.74}     \\ \cline{1-8}
\textbf{05}      & 10.63        & 10.04    & 8.59     & 8.27     & 8.31         & \textbf{7.46}& 8.07              \\ \cline{1-8}
\textbf{06}      & 7.15         & 9.53     & 4.14     & 7.80     & 6.49         & 5.04         & \textbf{3.46}     \\ \cline{1-8}
\textbf{07}      & 5.28         & 5.15     & 2.66     & 2.19     & 4.05         & 2.26         & \textbf{1.91}     \\ \cline{1-8}
\textbf{08}      & 17.13        & 18.87    & 13.19    & 10.97    & 9.74         & 10.58        & \textbf{9.18}     \\ \cline{1-8}
\textbf{09}      & 10.90        & 9.31     & 5.52     & 10.37    & 7.68         & 4.59         & \textbf{4.44}     \\ \cline{1-8}
\textbf{10}      & 20.30        & 17.99    & 11.98    & 15.77    & 12.96        & 12.38        & \textbf{10.39}    \\ \cline{1-8}
\end{tabular}\vspace{2pt}
\caption{Alignment errors on ten image pairs for each evaluated methods. Our method usually yields the lowest alignment error.}
\label{table:result}\vspace{-5pt}
\end{table}

\begin{figure}[H]
\begin{center}
\renewcommand{\tabcolsep}{1pt}
\begin{tabular} {m{0.02\textwidth} m{0.19\textwidth} m{0.13\textwidth} m{0.13\textwidth} m{0.005\textwidth} m{0.19\textwidth} m{0.13\textwidth} m{0.13\textwidth}}
 \rotatebox{90}{Global}  & \includegraphics[width=0.19\textwidth]{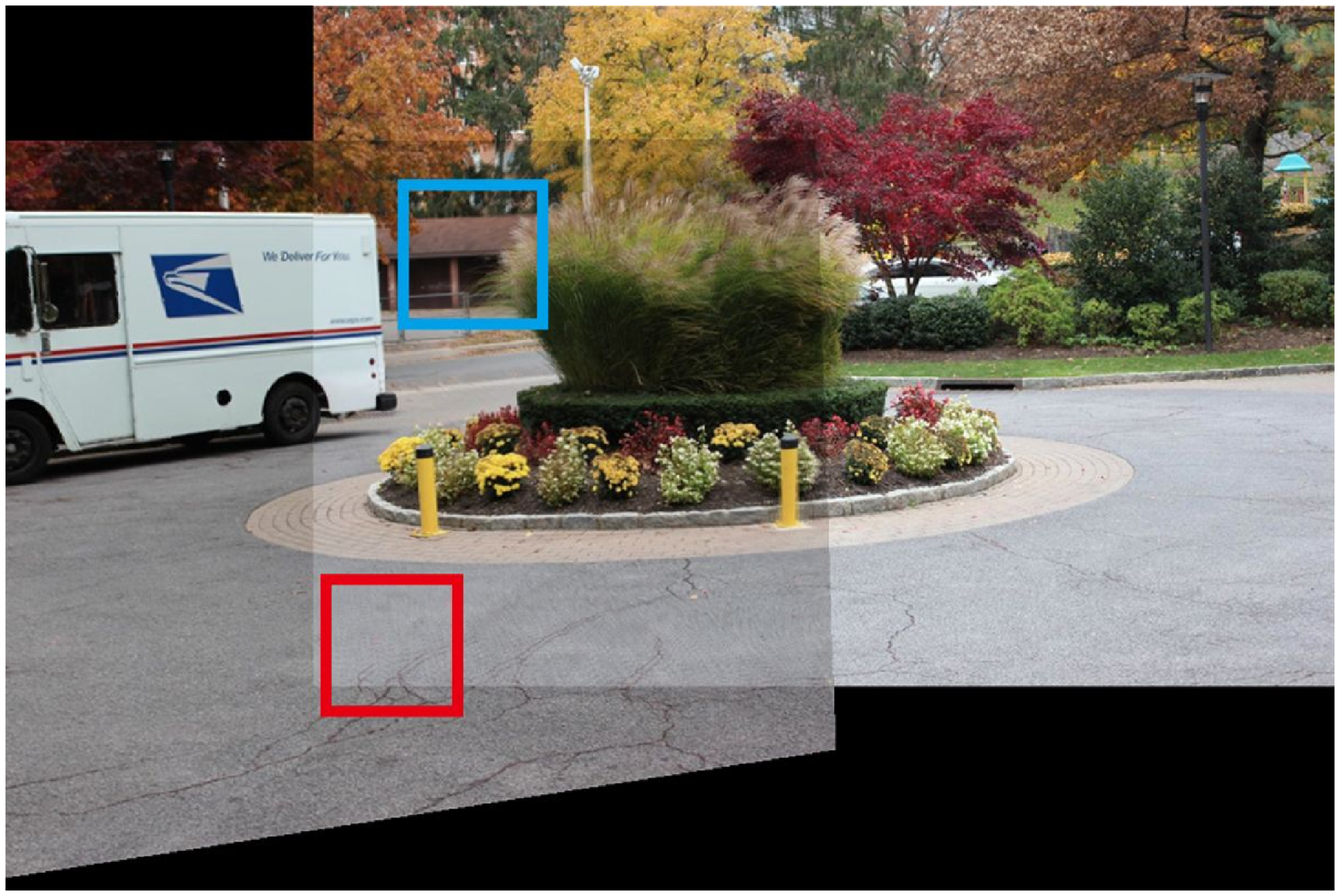} & \includegraphics[width=0.13\textwidth]{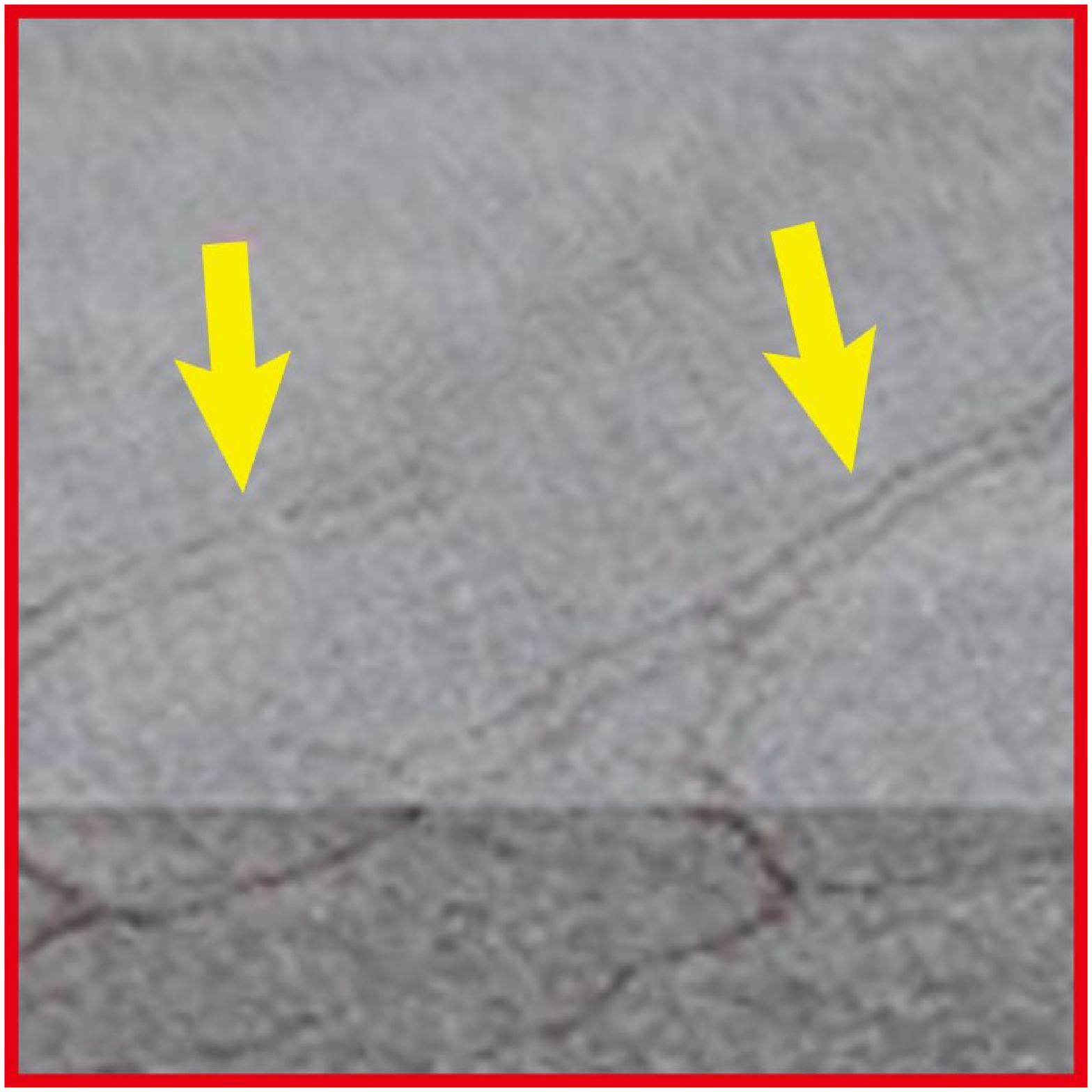}
      & \includegraphics[width=0.13\textwidth]{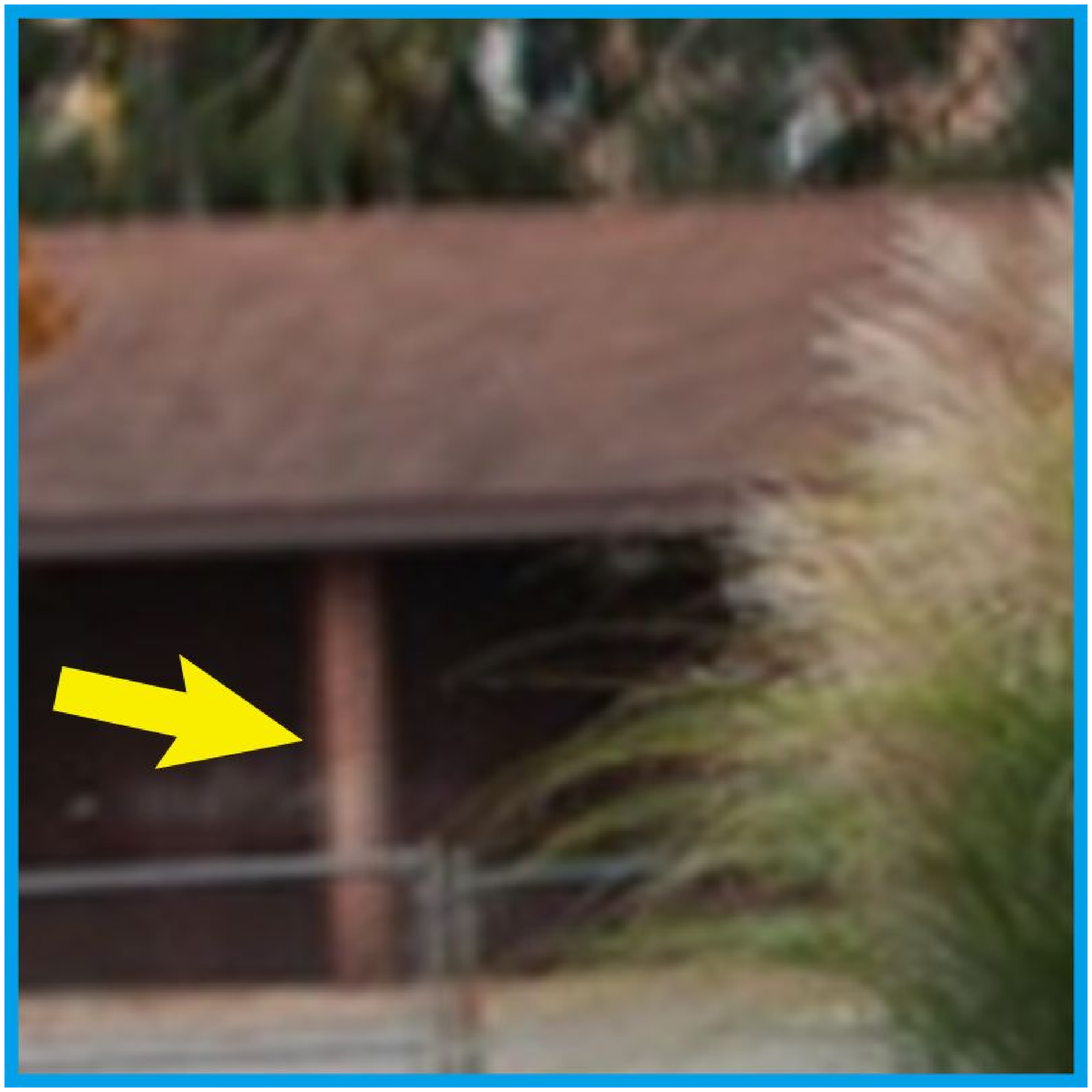} & & \includegraphics[width=0.19\textwidth]{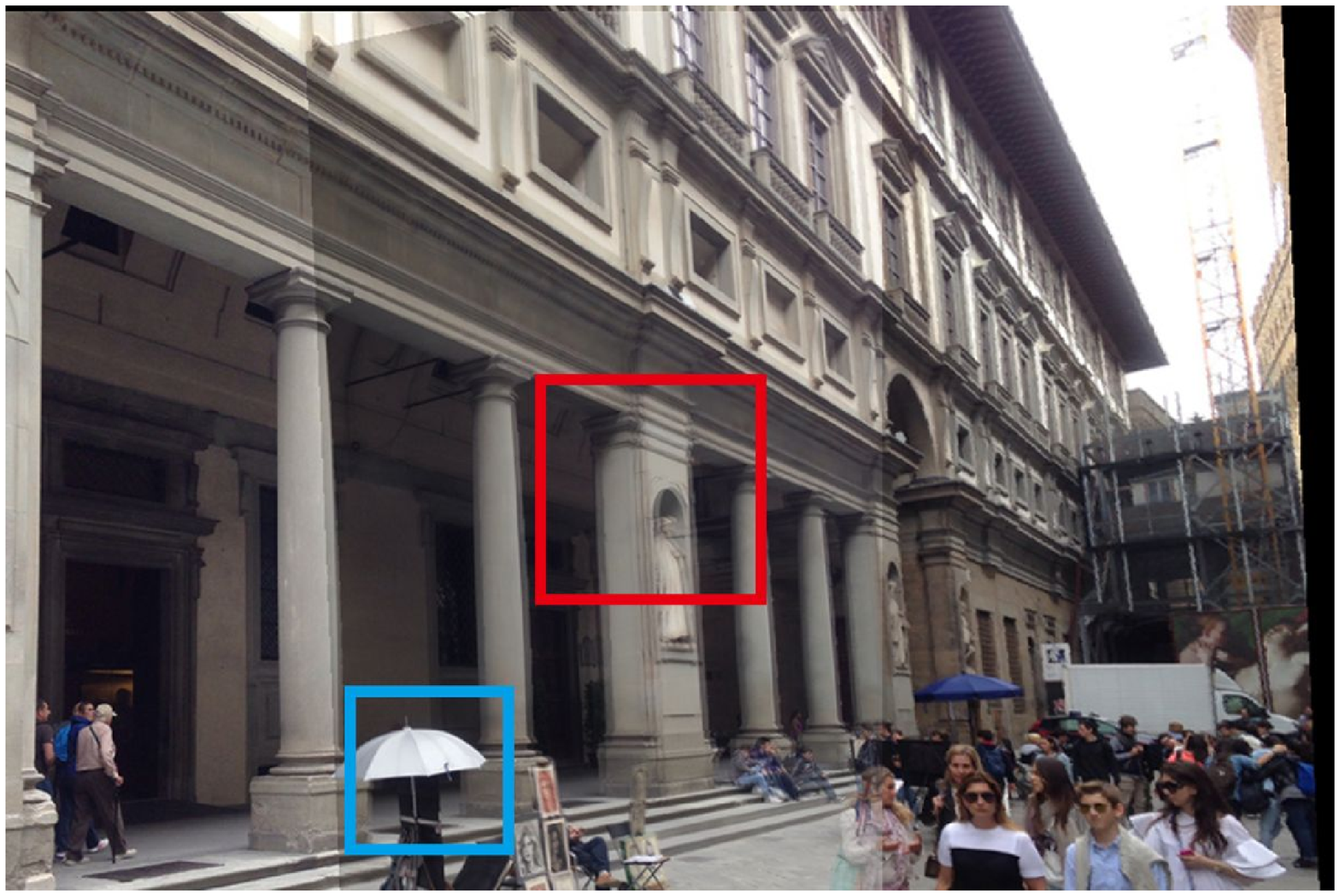}
      & \includegraphics[width=0.13\textwidth]{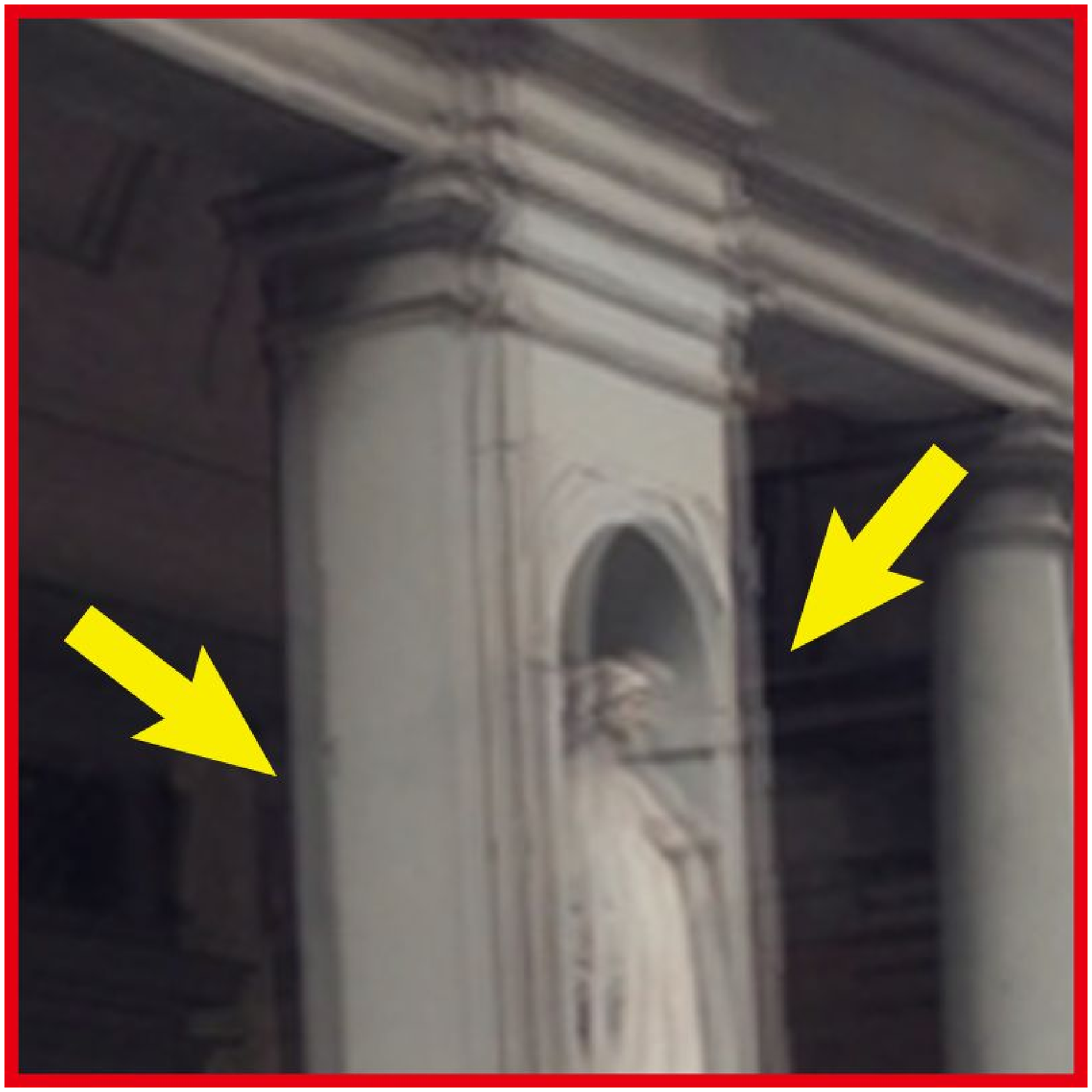} & \includegraphics[width=0.13\textwidth]{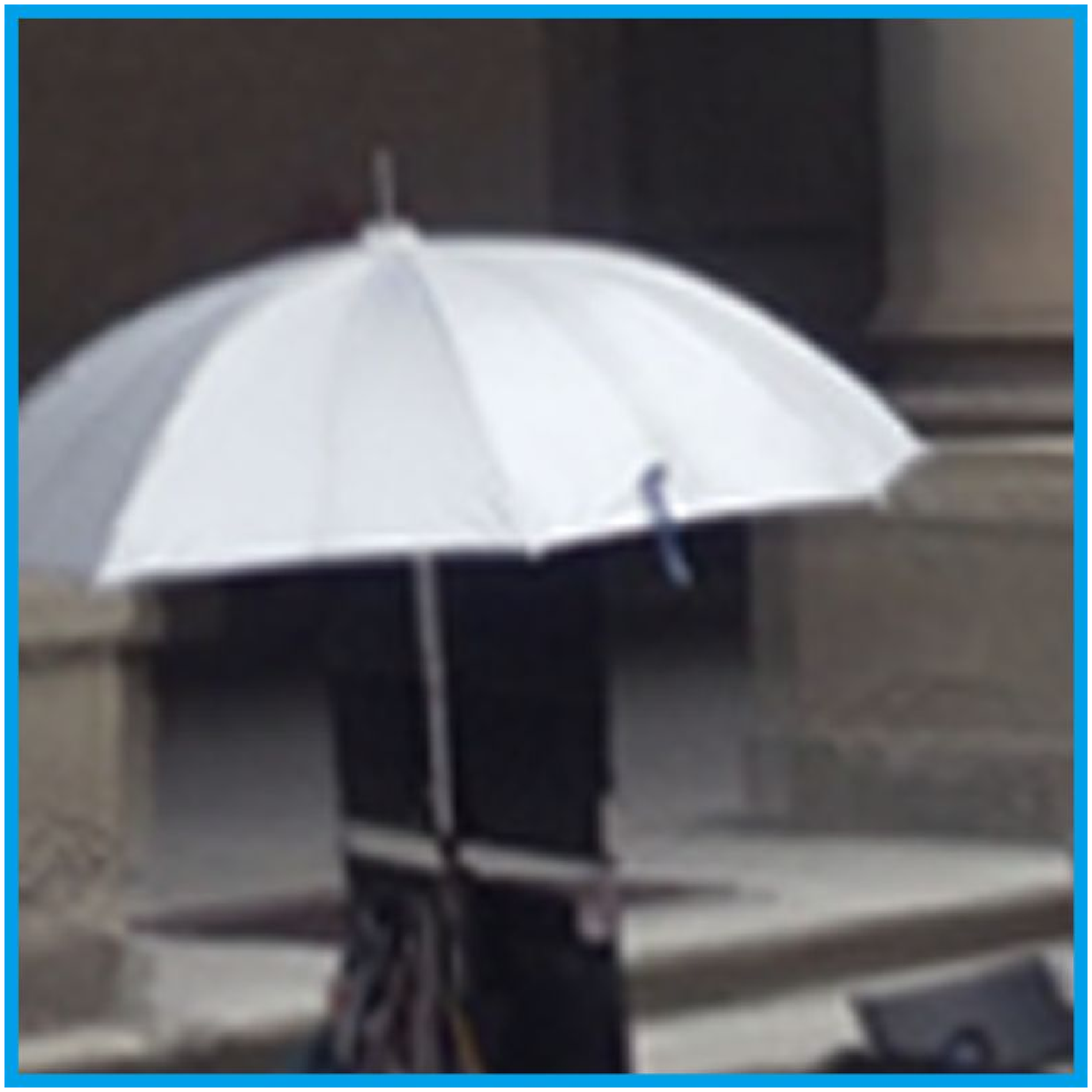}\\
 \rotatebox{90}{CPW}  & \includegraphics[width=0.19\textwidth]{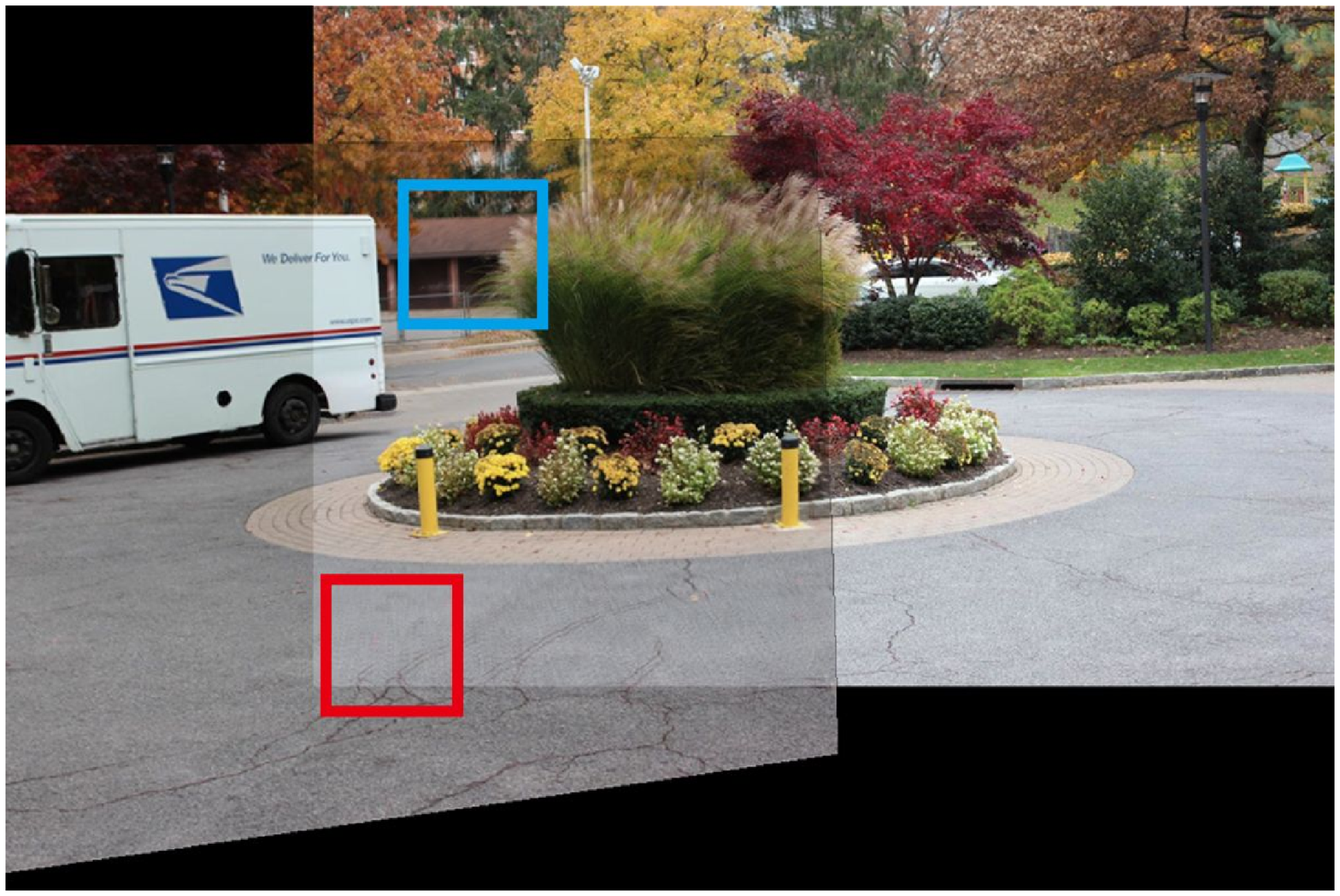} & \includegraphics[width=0.13\textwidth]{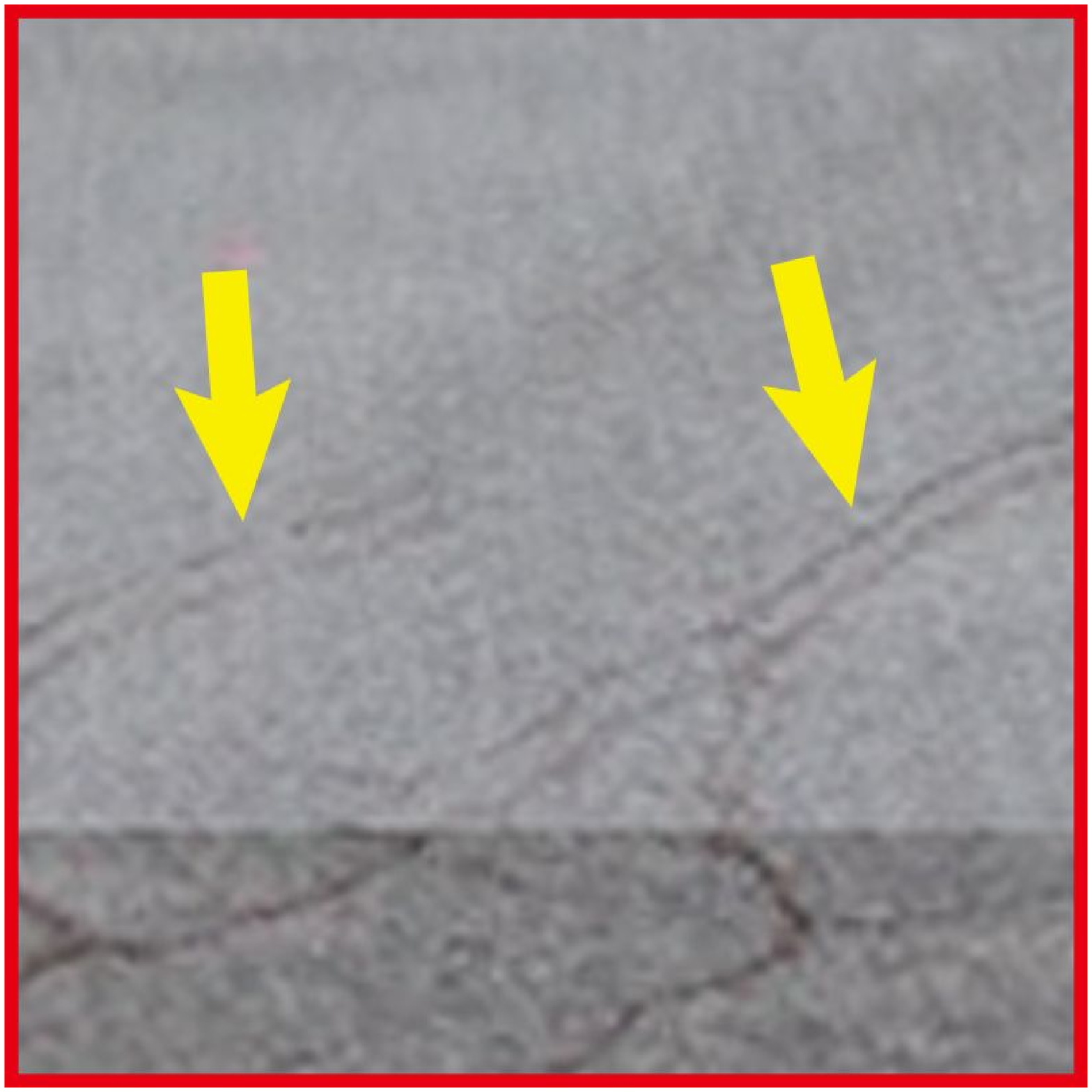}
      & \includegraphics[width=0.13\textwidth]{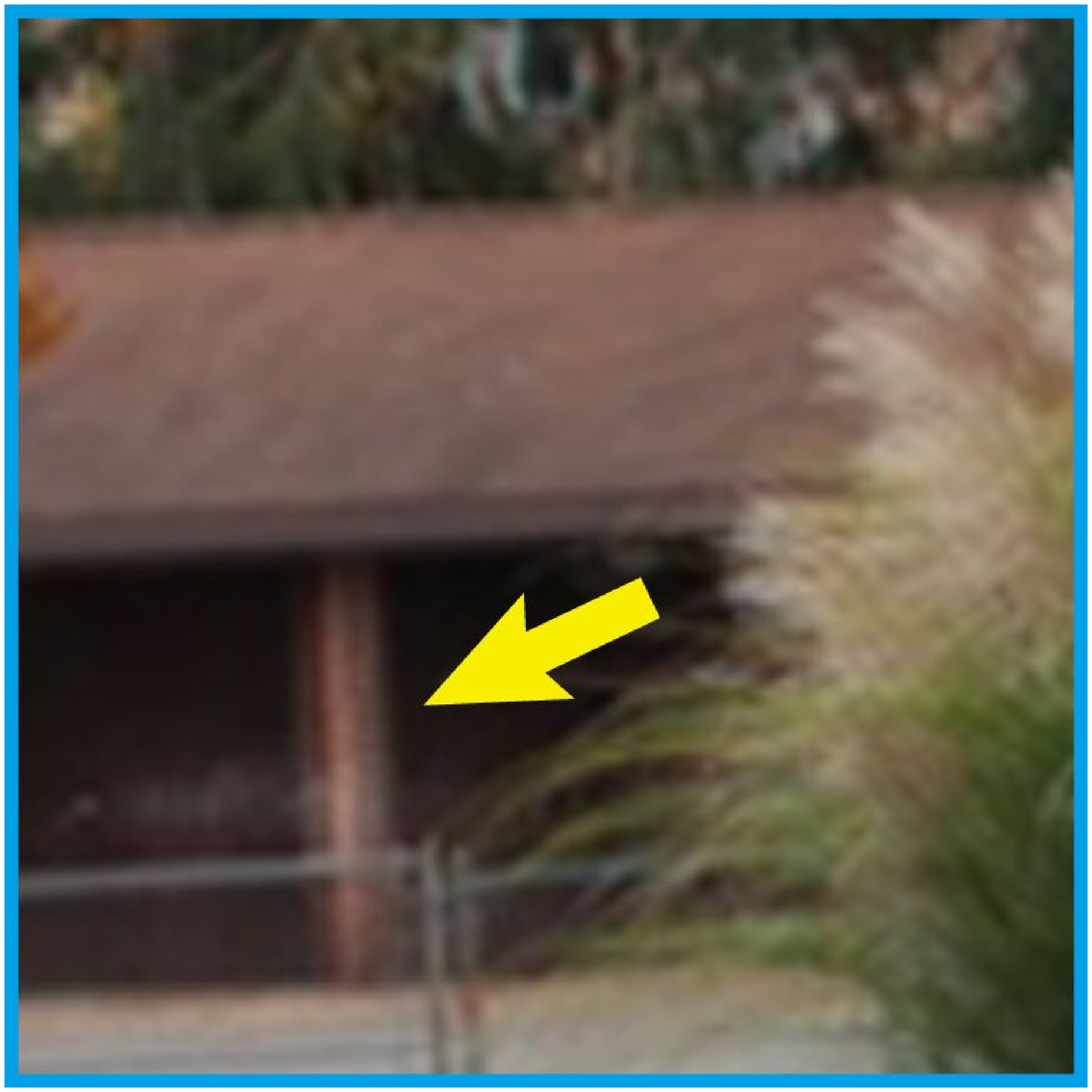} & & \includegraphics[width=0.19\textwidth]{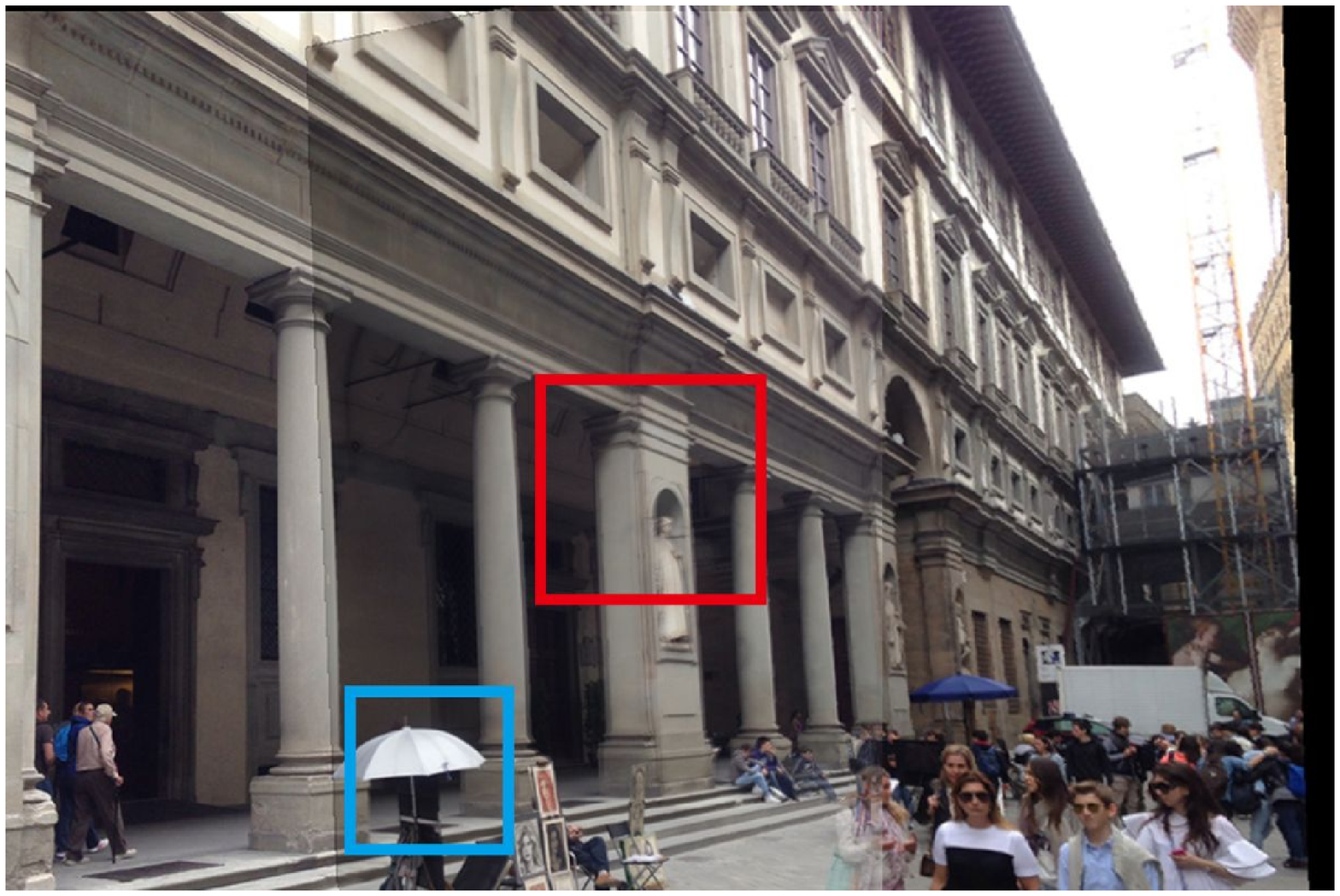}
      & \includegraphics[width=0.13\textwidth]{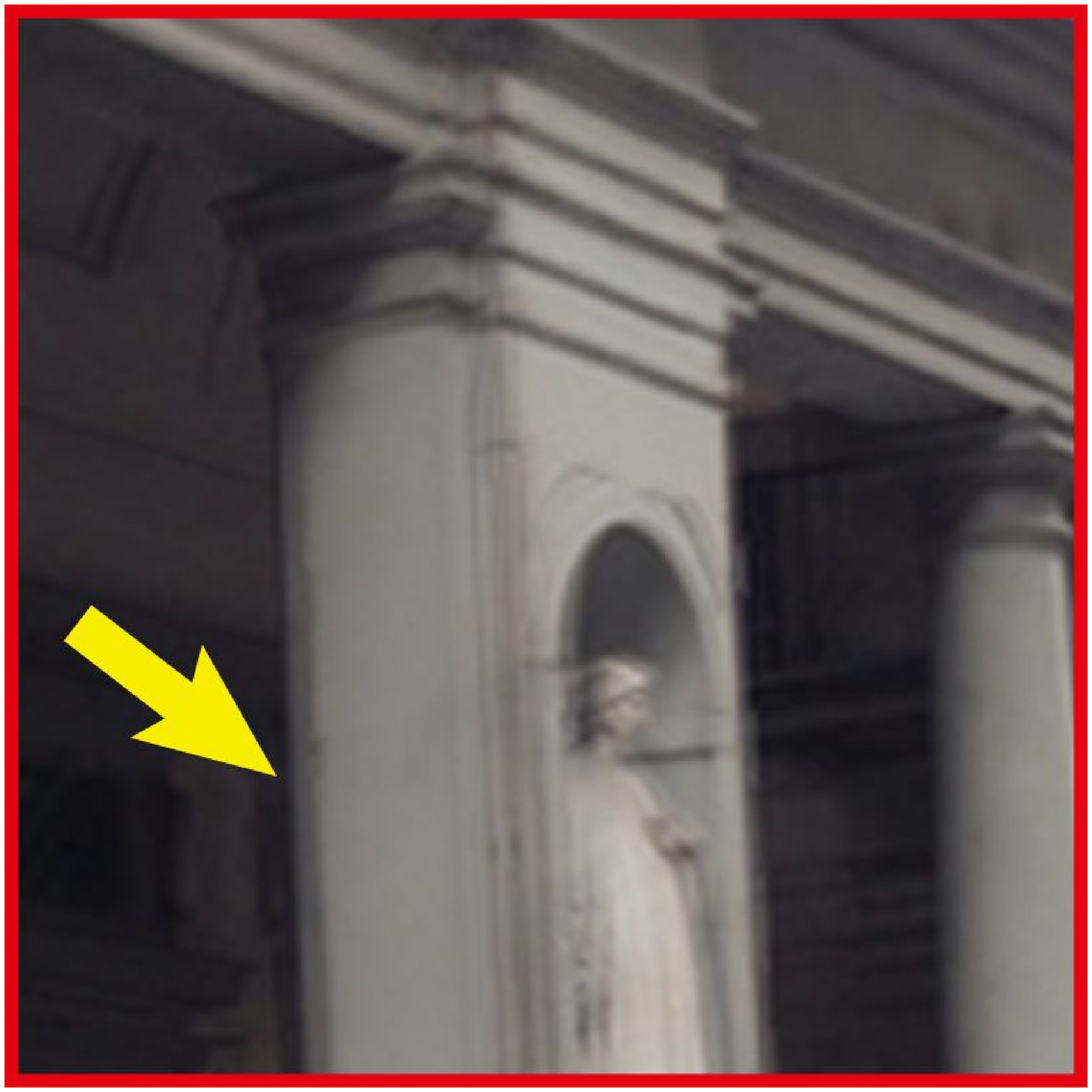} & \includegraphics[width=0.13\textwidth]{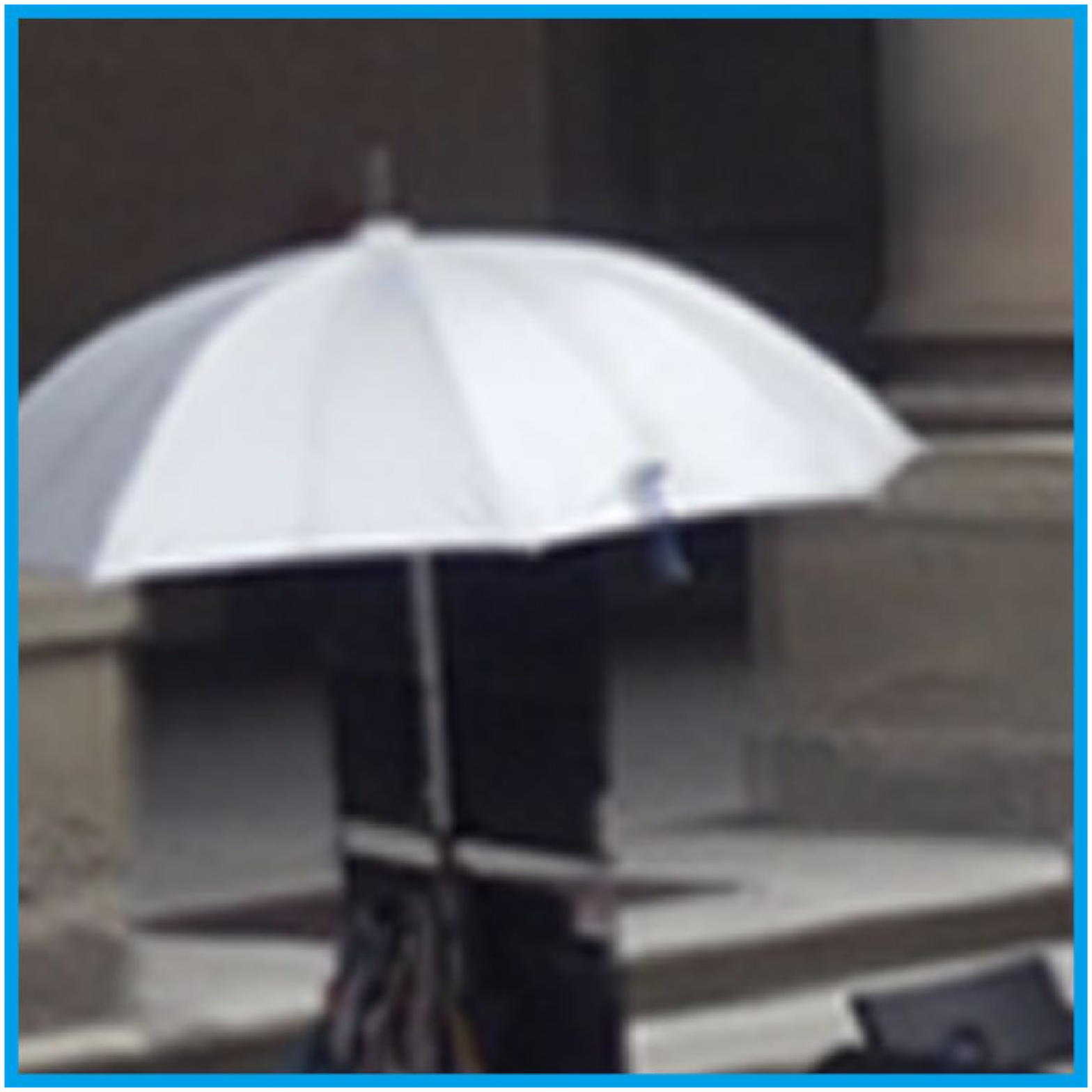}\\
 \rotatebox{90}{DF-W}  & \includegraphics[width=0.19\textwidth]{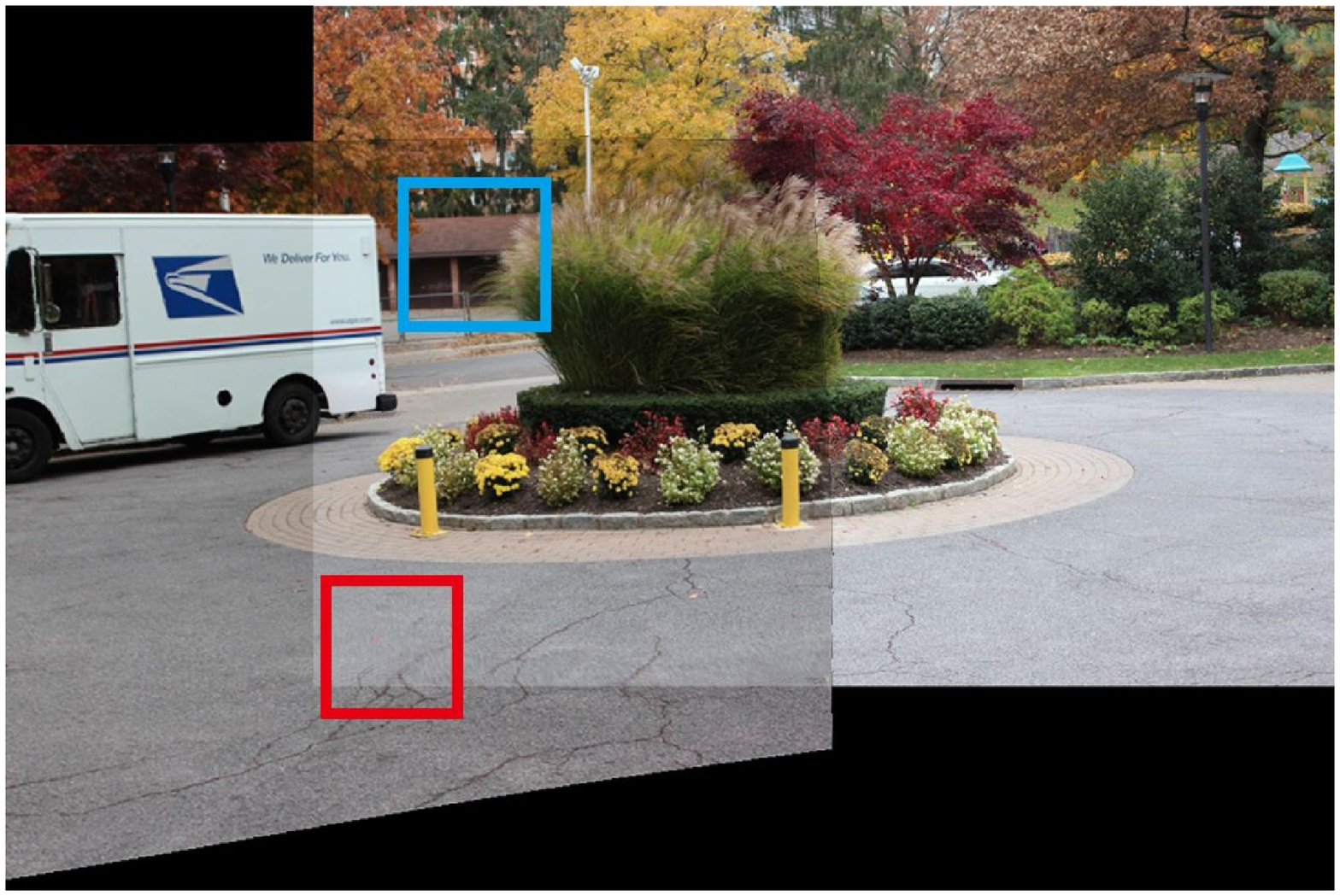} & \includegraphics[width=0.13\textwidth]{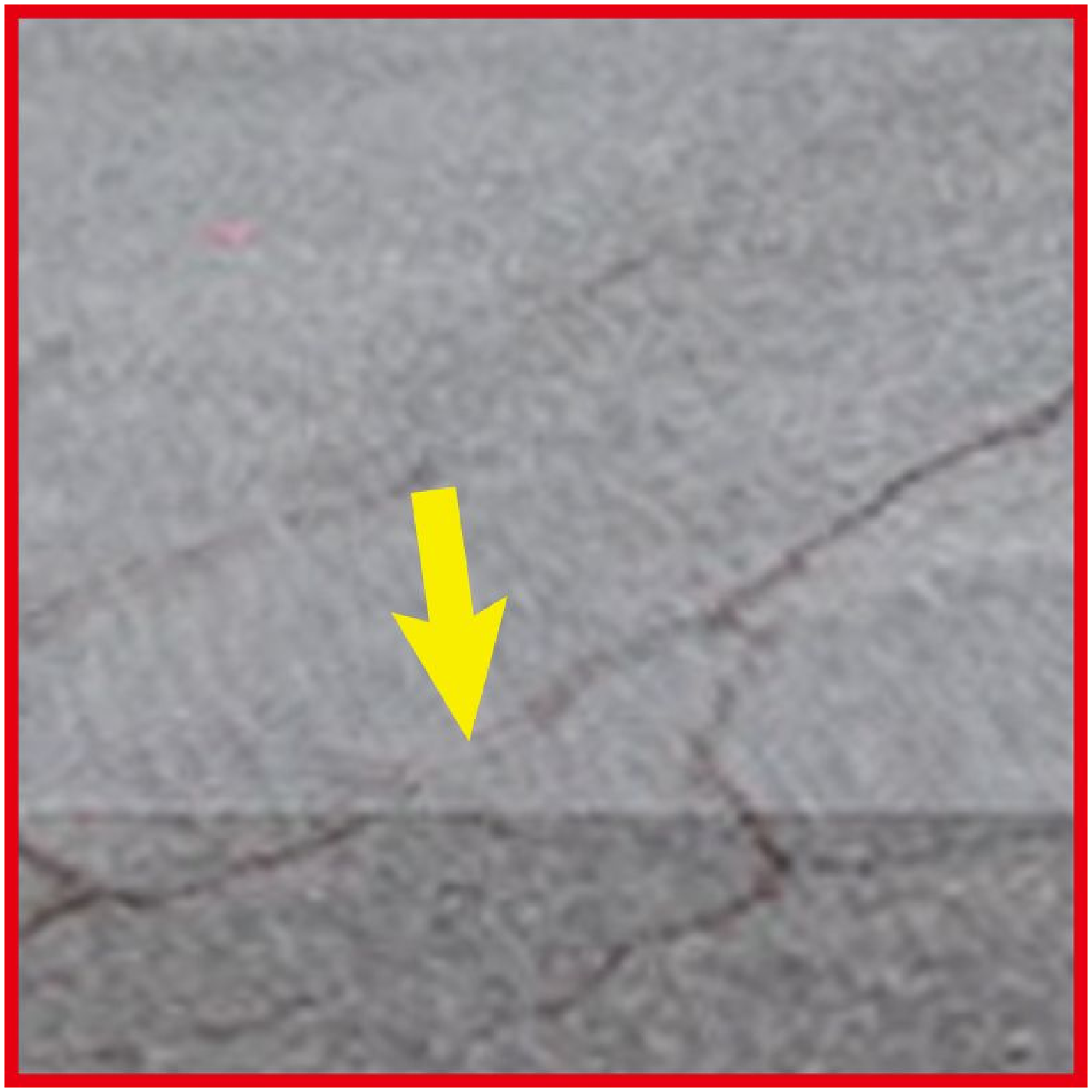}
      & \includegraphics[width=0.13\textwidth]{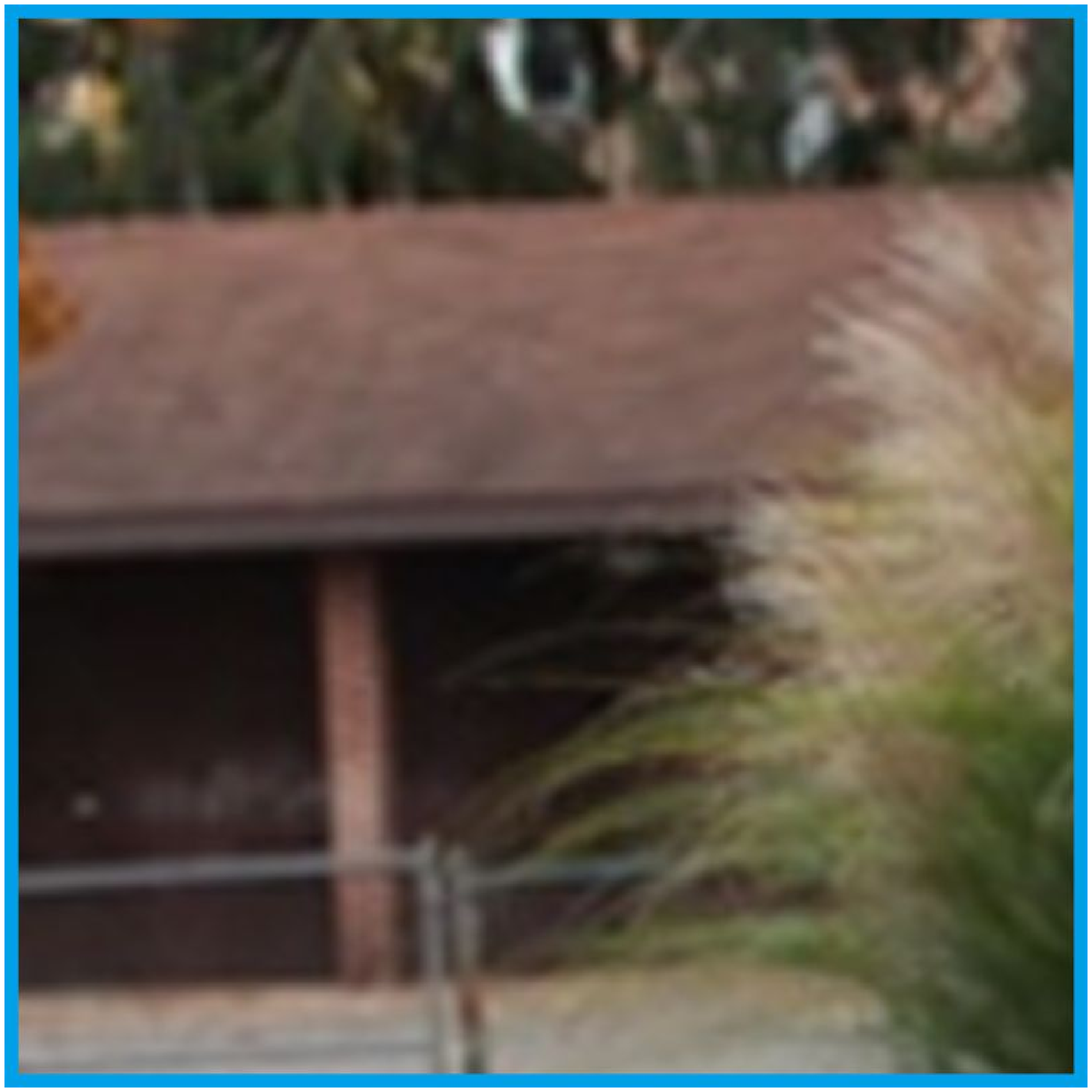} & & \includegraphics[width=0.19\textwidth]{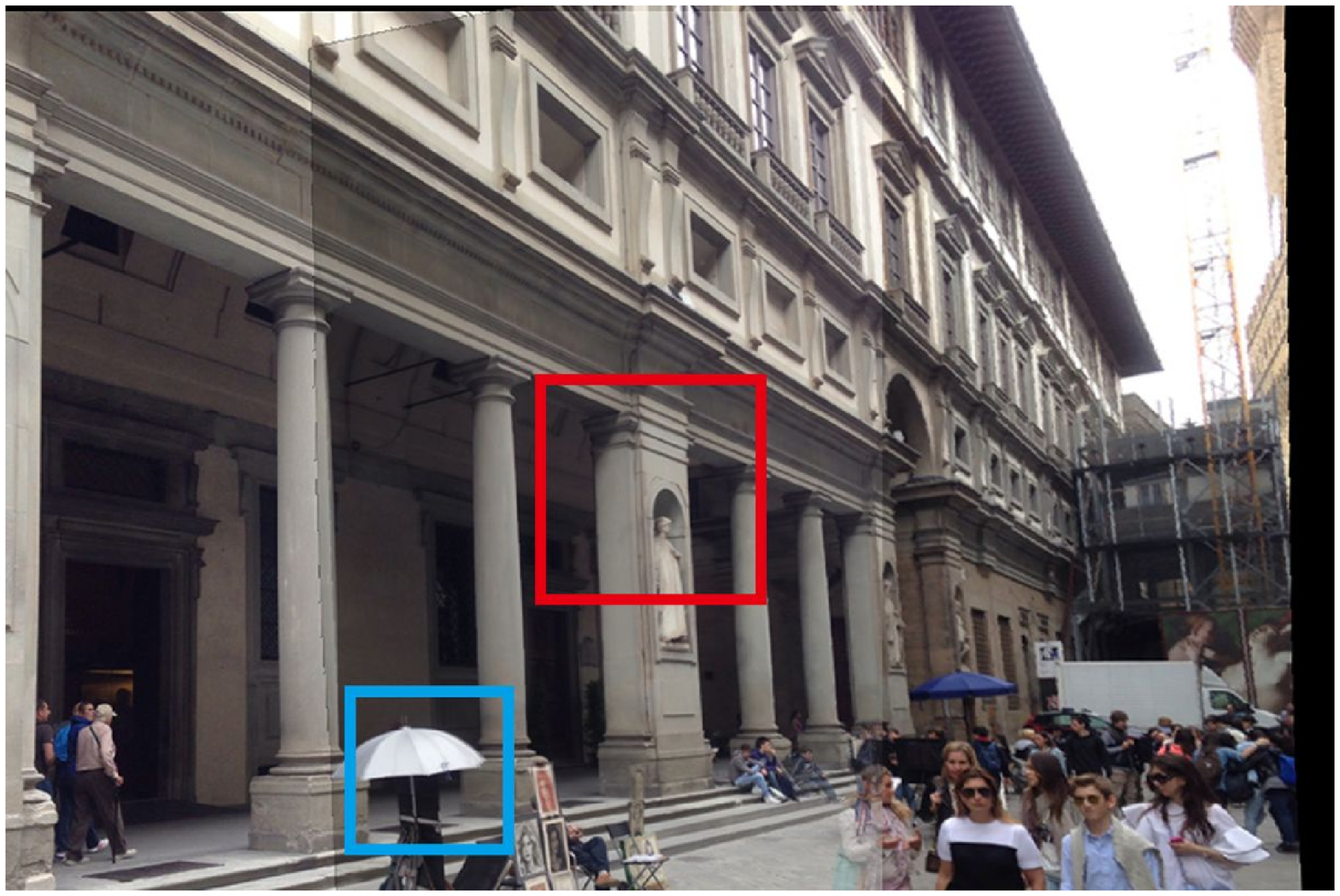}
      & \includegraphics[width=0.13\textwidth]{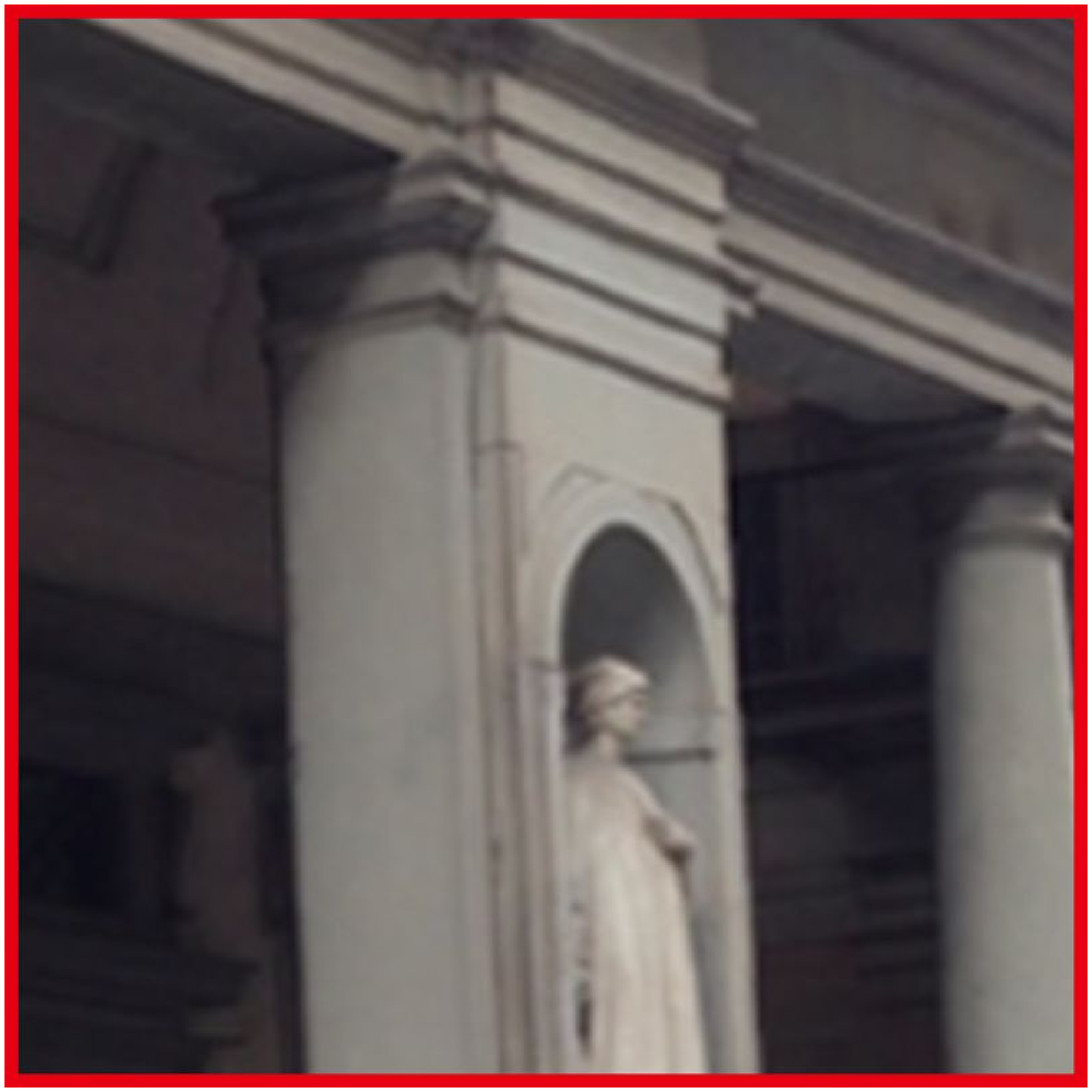} & \includegraphics[width=0.13\textwidth]{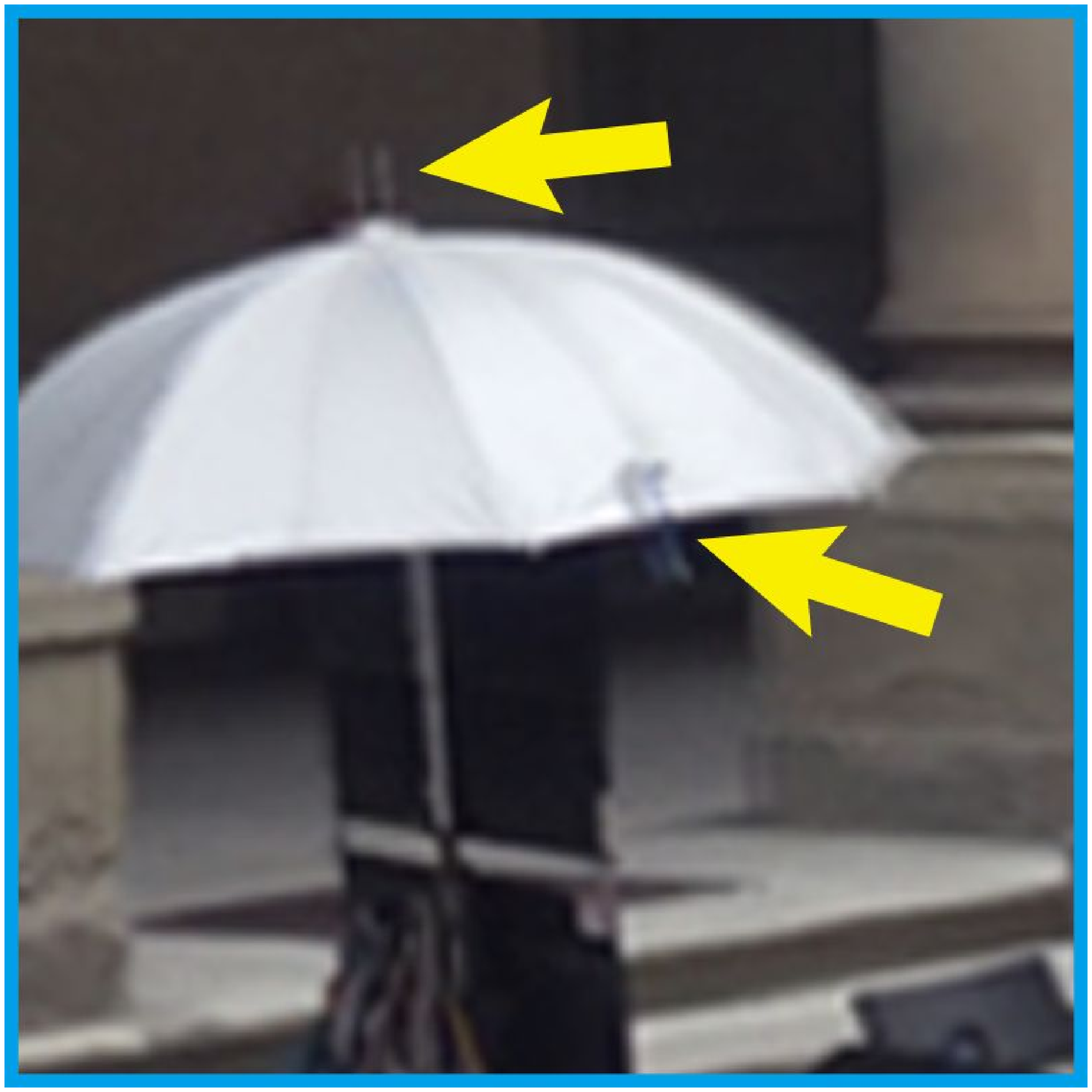}\\
 \rotatebox{90}{GCPW}  & \includegraphics[width=0.19\textwidth]{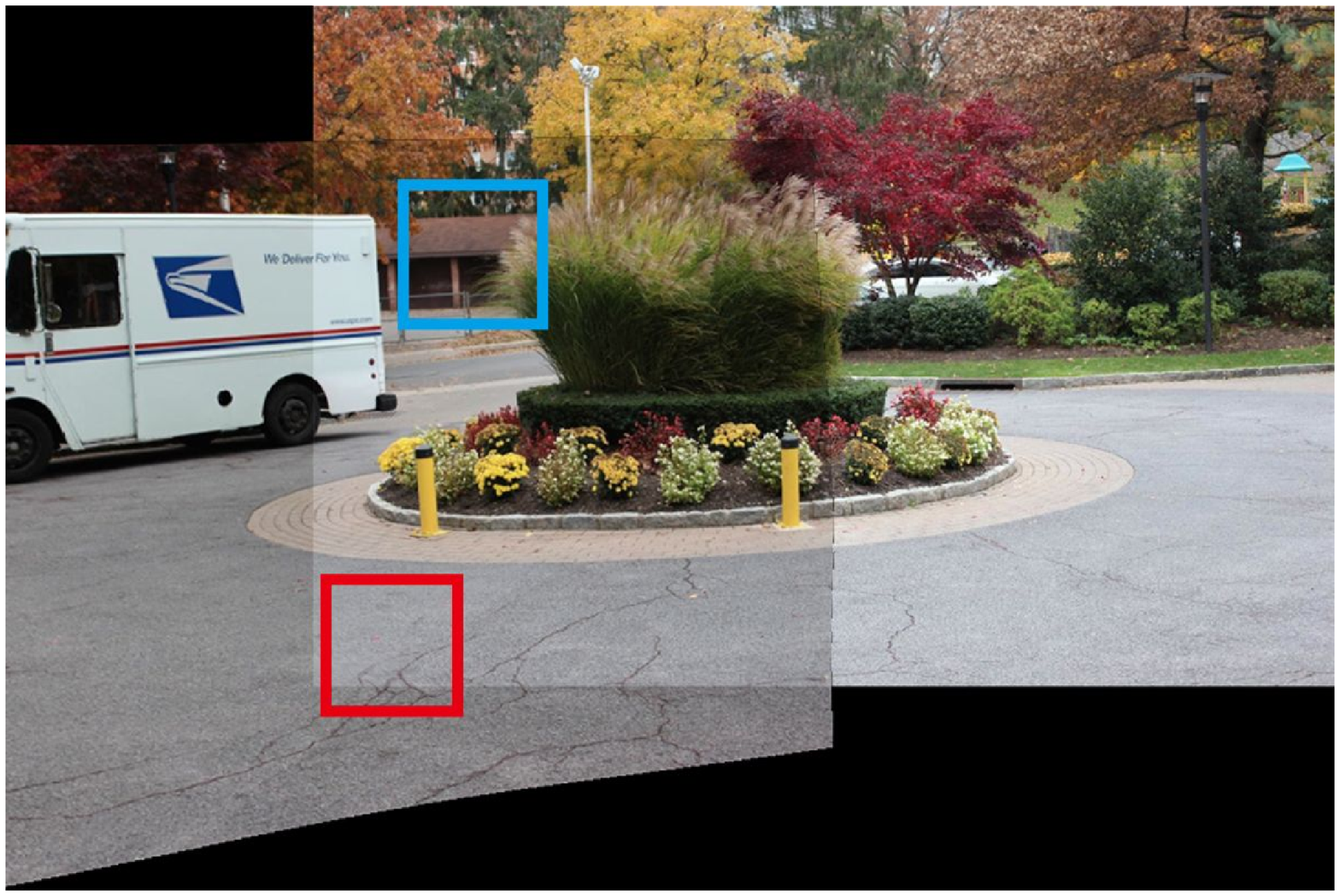} & \includegraphics[width=0.13\textwidth]{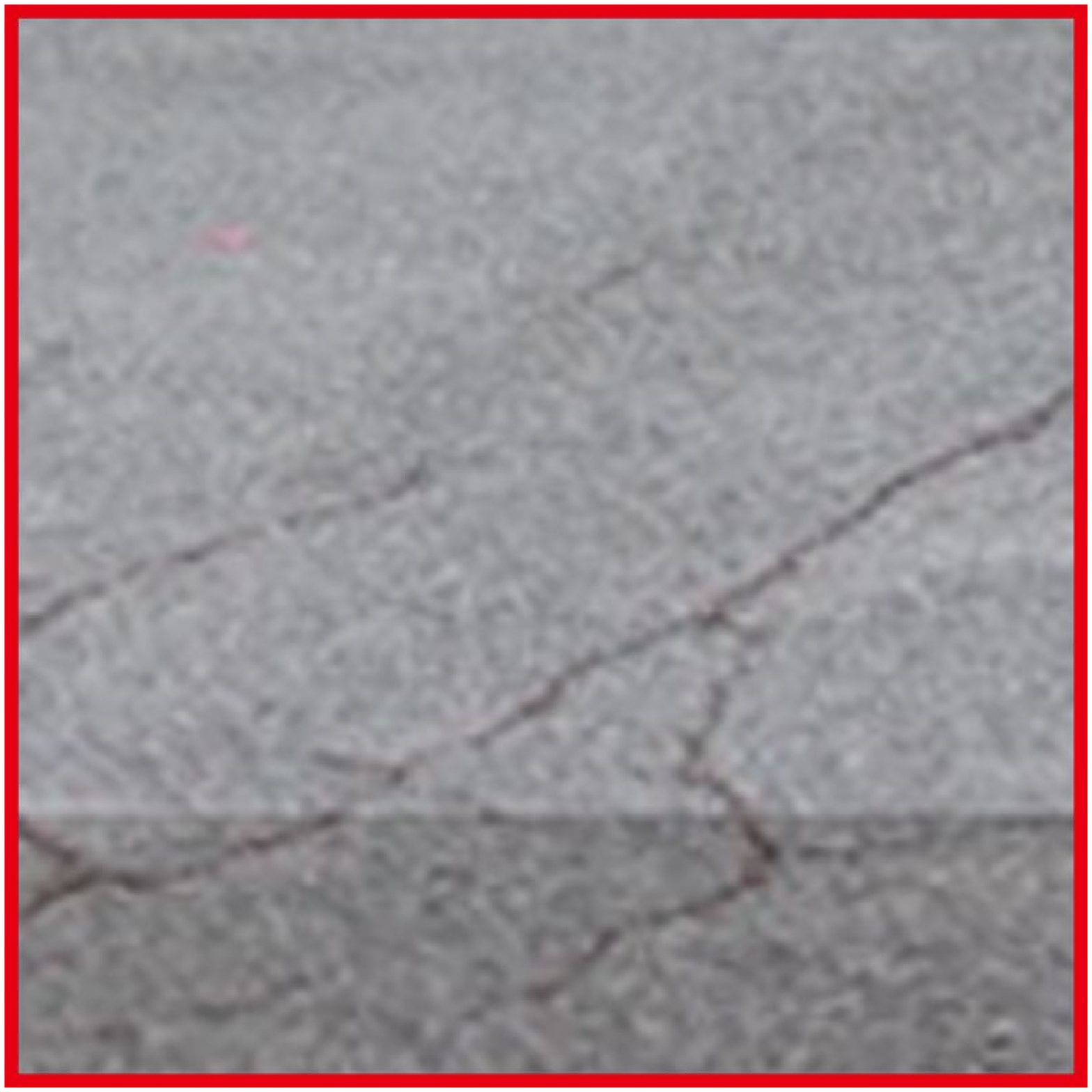}
      & \includegraphics[width=0.13\textwidth]{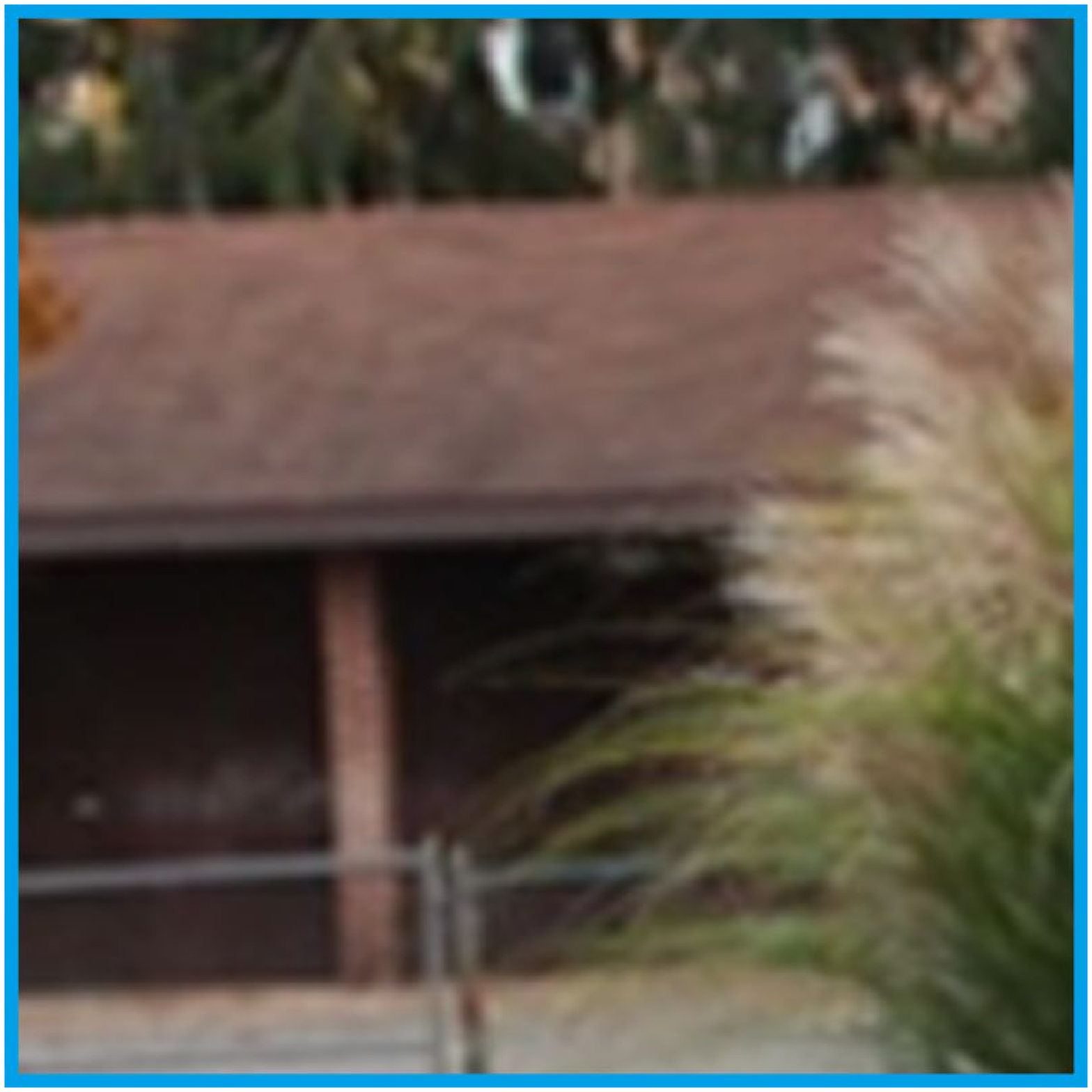} & & \includegraphics[width=0.19\textwidth]{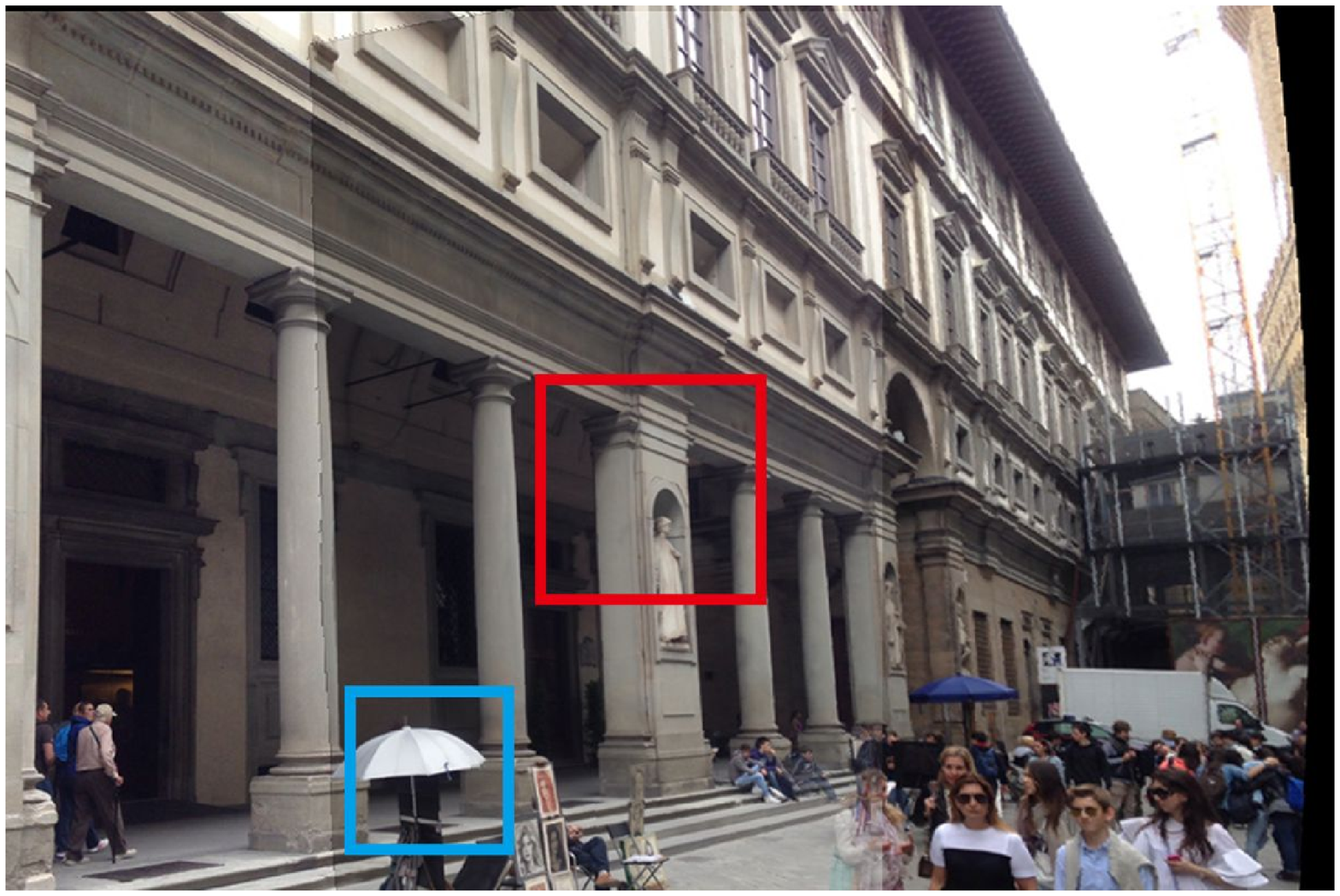}
      & \includegraphics[width=0.13\textwidth]{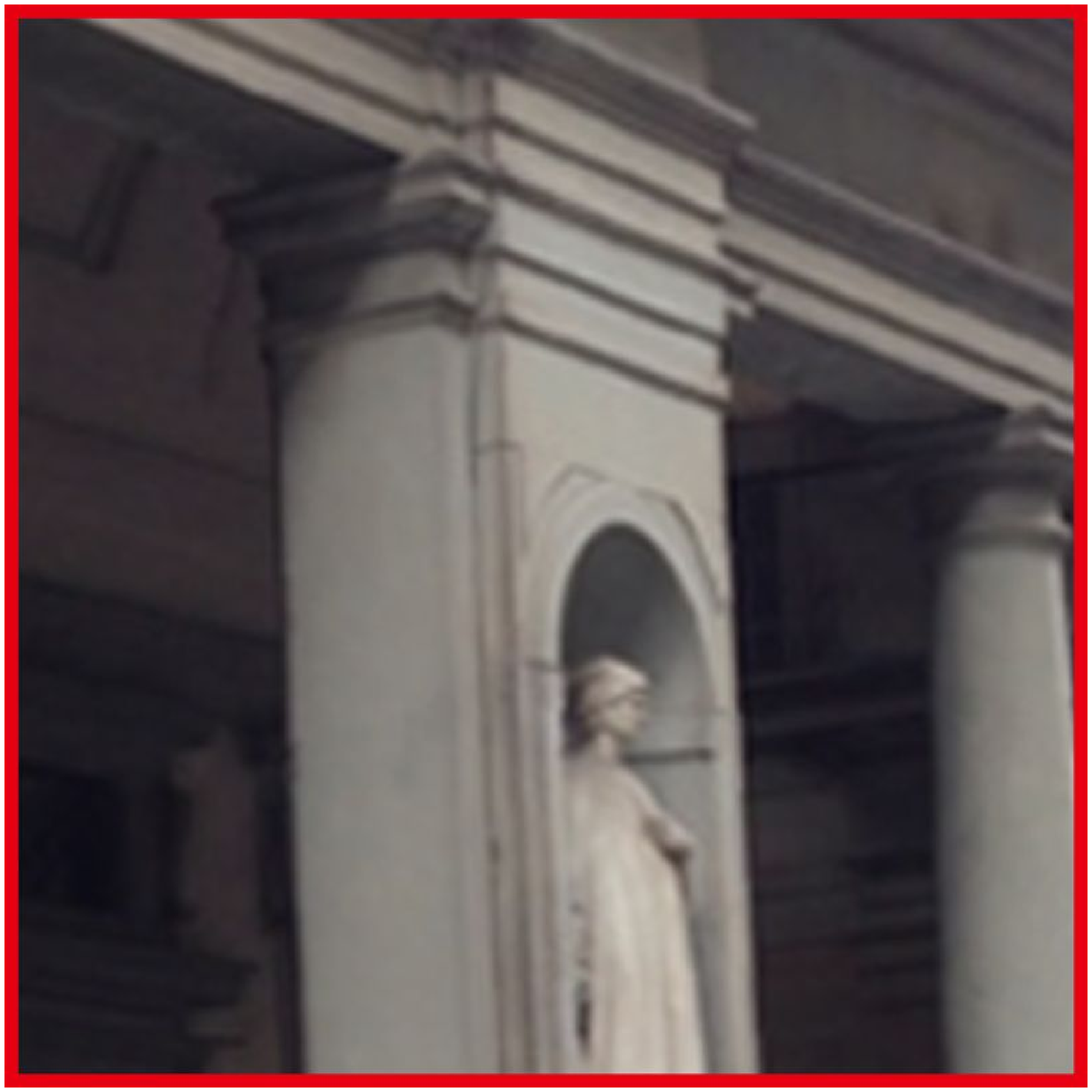} & \includegraphics[width=0.13\textwidth]{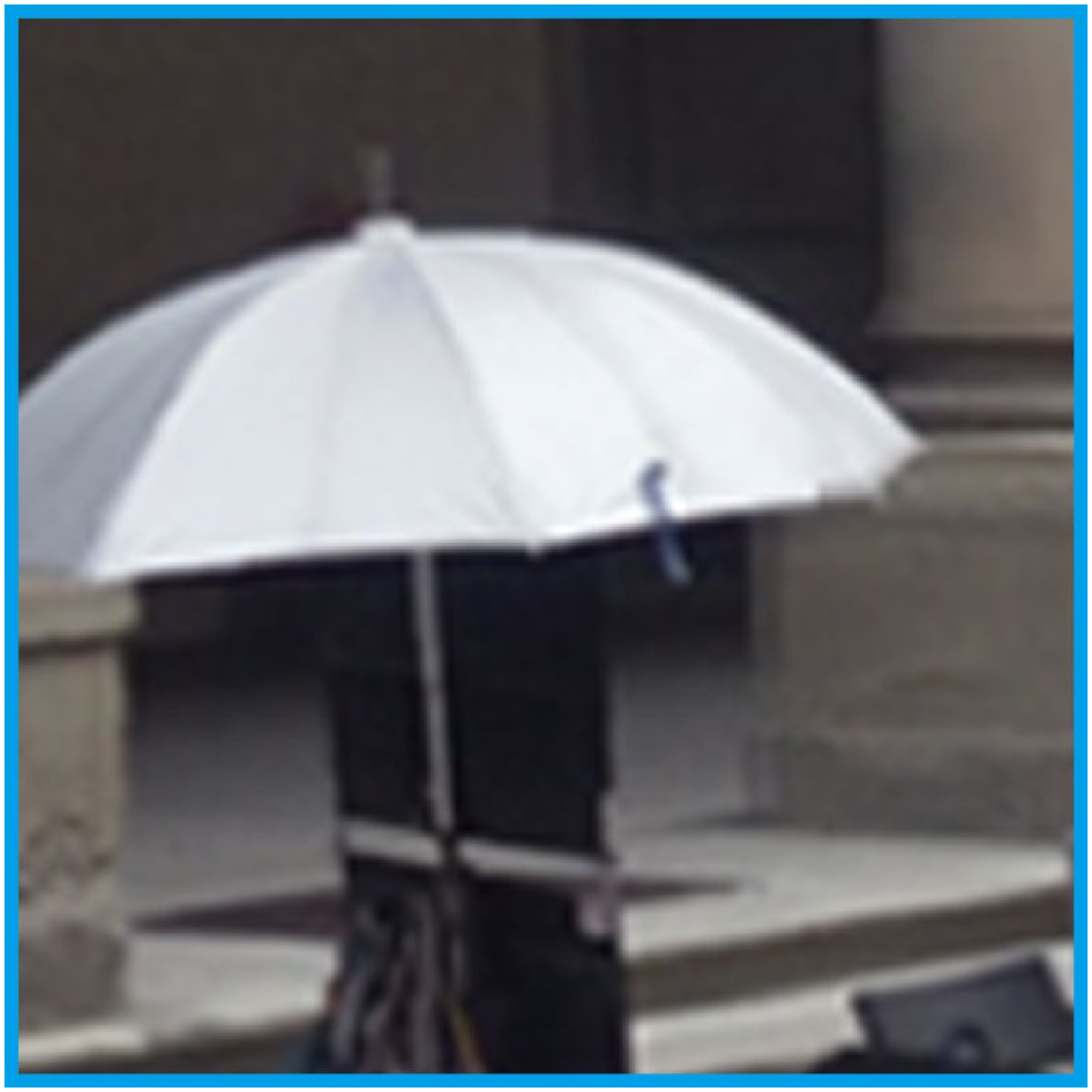}\\
\end{tabular}
\caption{Comparisons with three geometric-based methods: global homography, CPW~\cite{liu2009content} and DF-W~\cite{li2015dual}. Yellow arrow indicates the misalignment and ghost artifacts. With the photometric guidance, our GCPW method outperforms other three methods.}\label{fig:result_experimentB1}\vspace{-15pt}
\end{center}
\end{figure}

Figure~\ref{fig:result_experimentB1} presents two groups of result that intuitively compare our method with global homography, CPW, and DF-W. As we can see, the global homography fails to produce satisfactory results because of the model limitation. The CPW and the DF-W still suffer obvious misalignments as they only use geometric constraints to align images locally. In contrast, our method combines geometric constraint with photometric constraint in the proposed GCPW framework. The experimental results show that our method produces the best alignment quality compared with these three geometric-based methods.

\begin{figure}[h]
\begin{center}
\renewcommand{\tabcolsep}{1pt}
\begin{tabular} {m{0.02\textwidth} m{0.195\textwidth} m{0.13\textwidth} m{0.13\textwidth} m{0.005\textwidth} m{0.195\textwidth} m{0.13\textwidth} m{0.13\textwidth}}
 \rotatebox{90}{\small{MF-W}} &\includegraphics[width=0.195\textwidth]{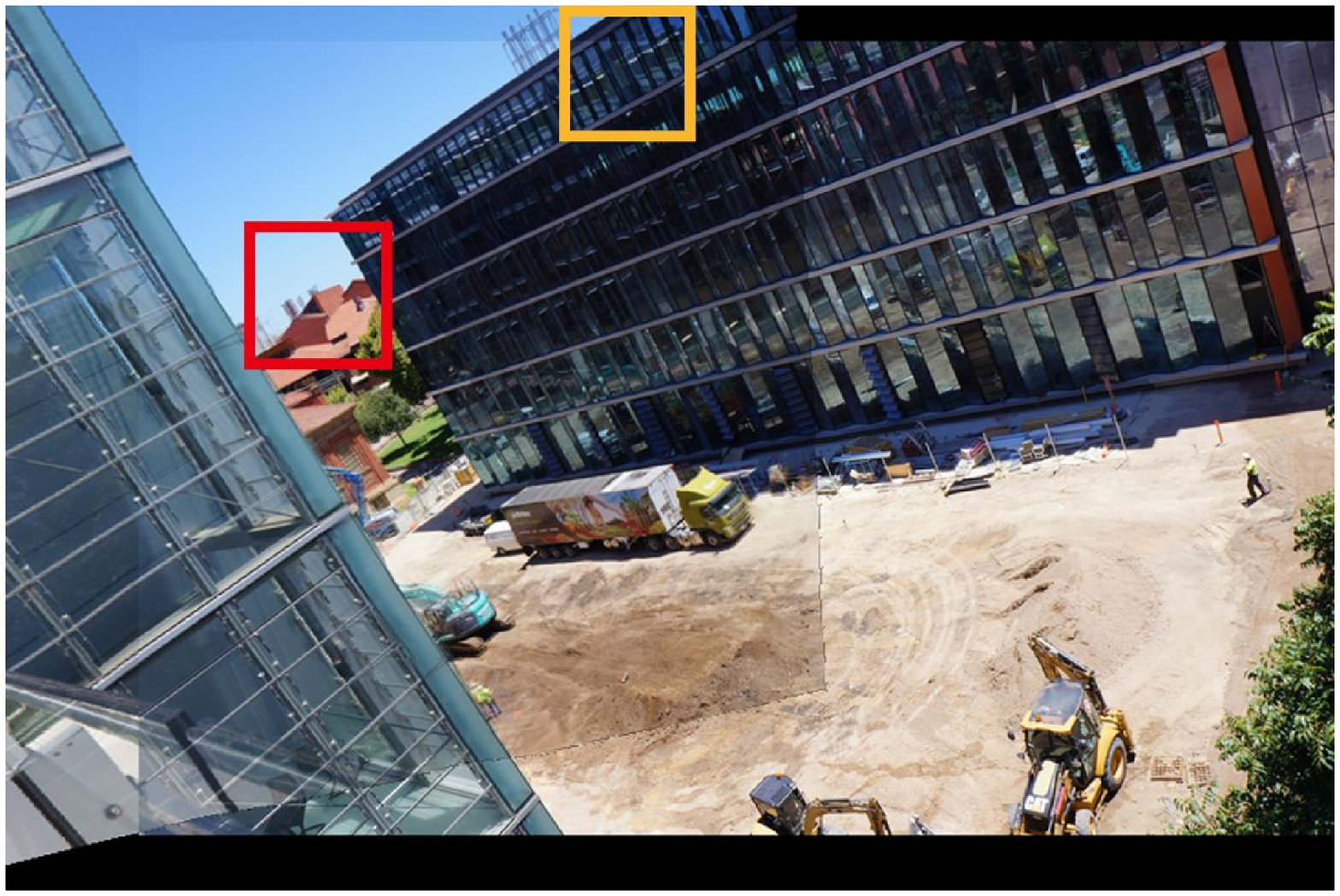}&\includegraphics[width=0.13\textwidth]{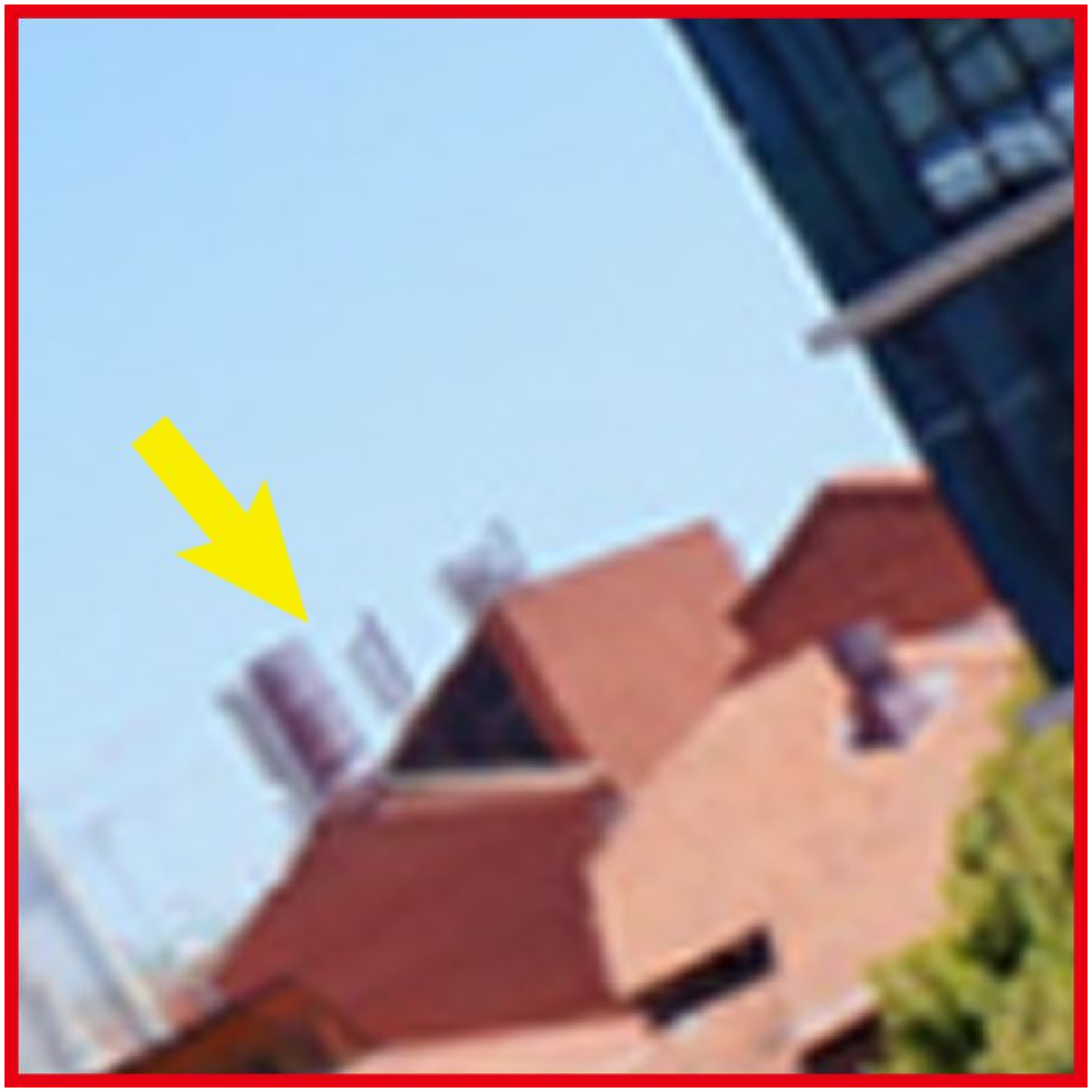}
      & \includegraphics[width=0.13\textwidth]{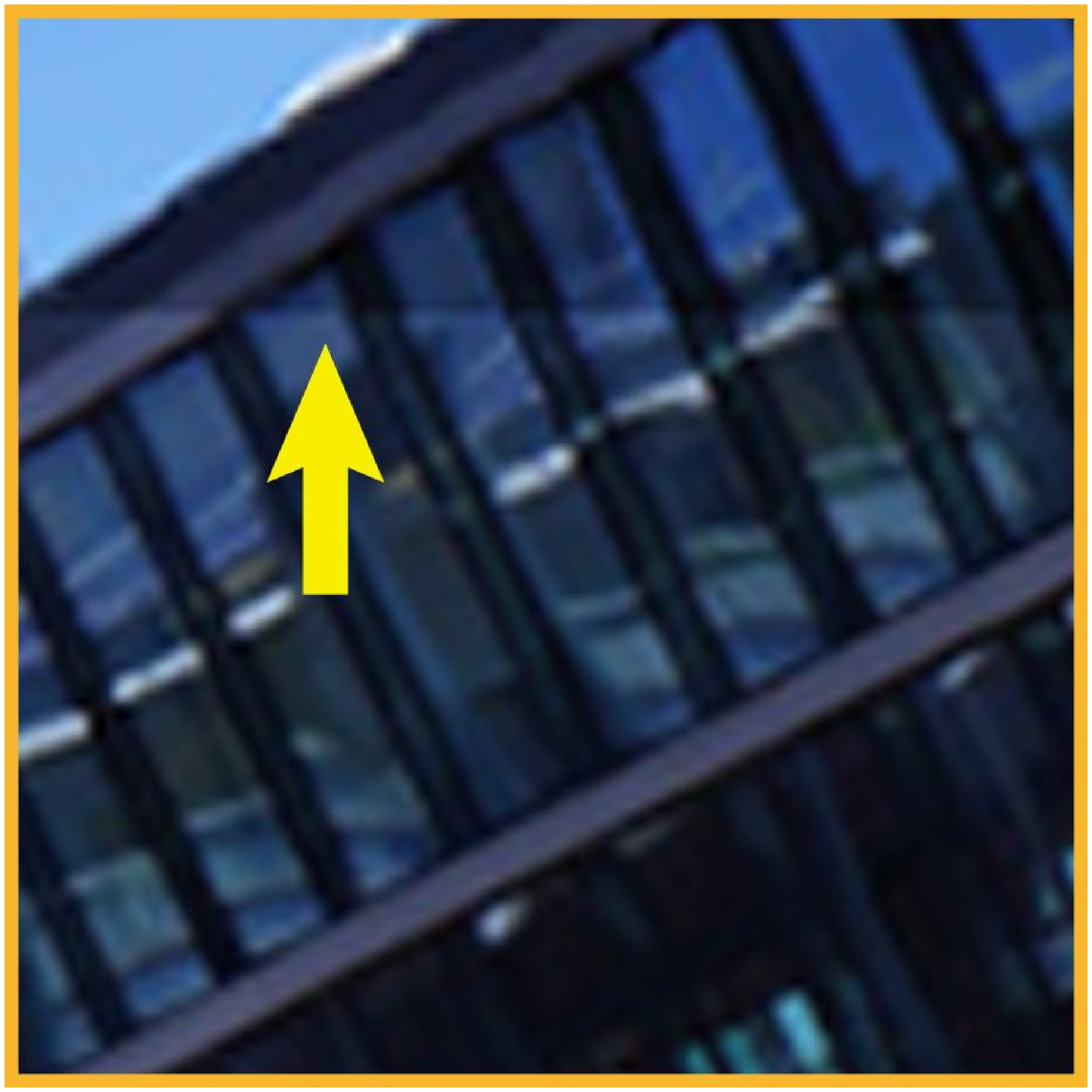} & & \includegraphics[width=0.195\textwidth]{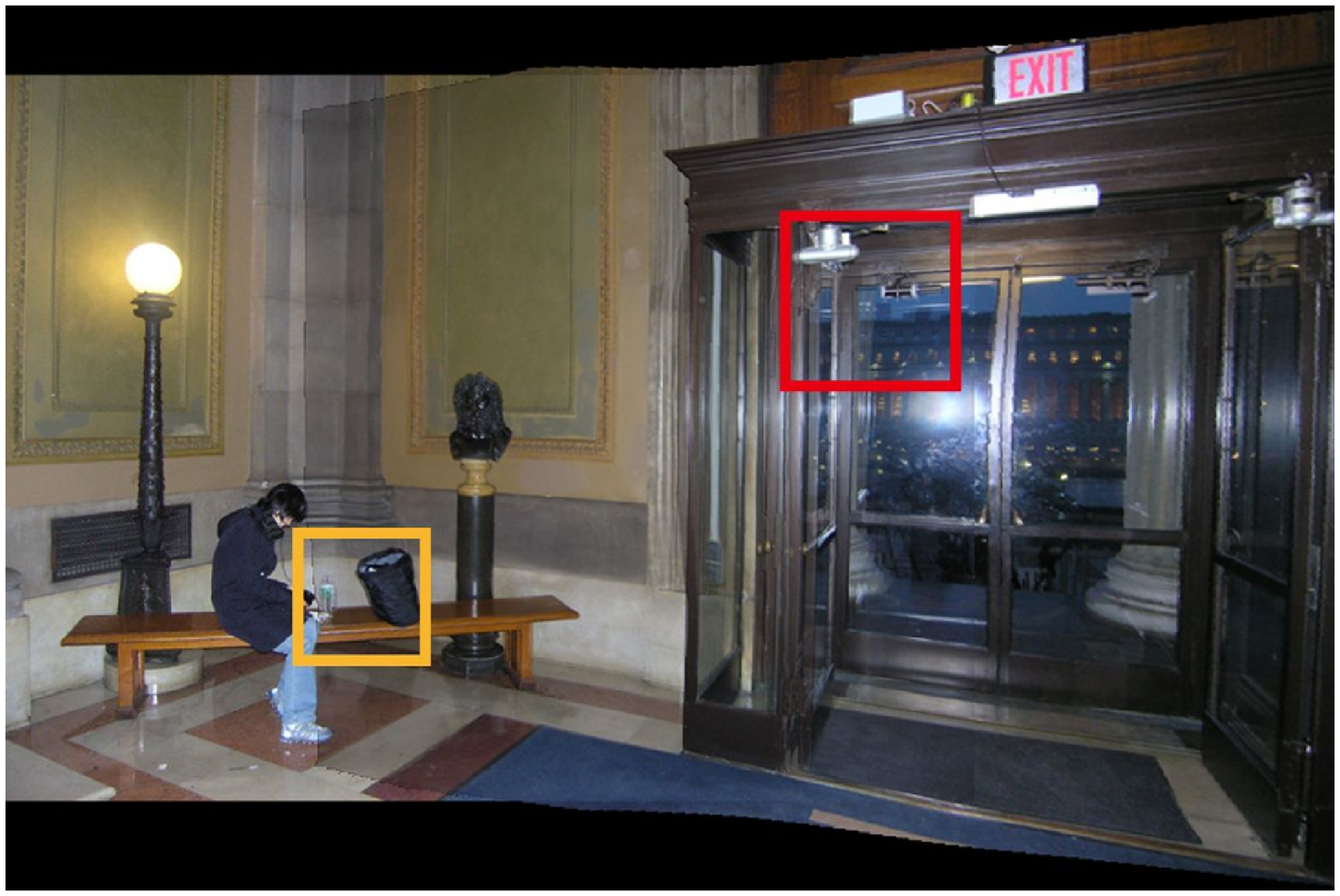}
      & \includegraphics[width=0.13\textwidth]{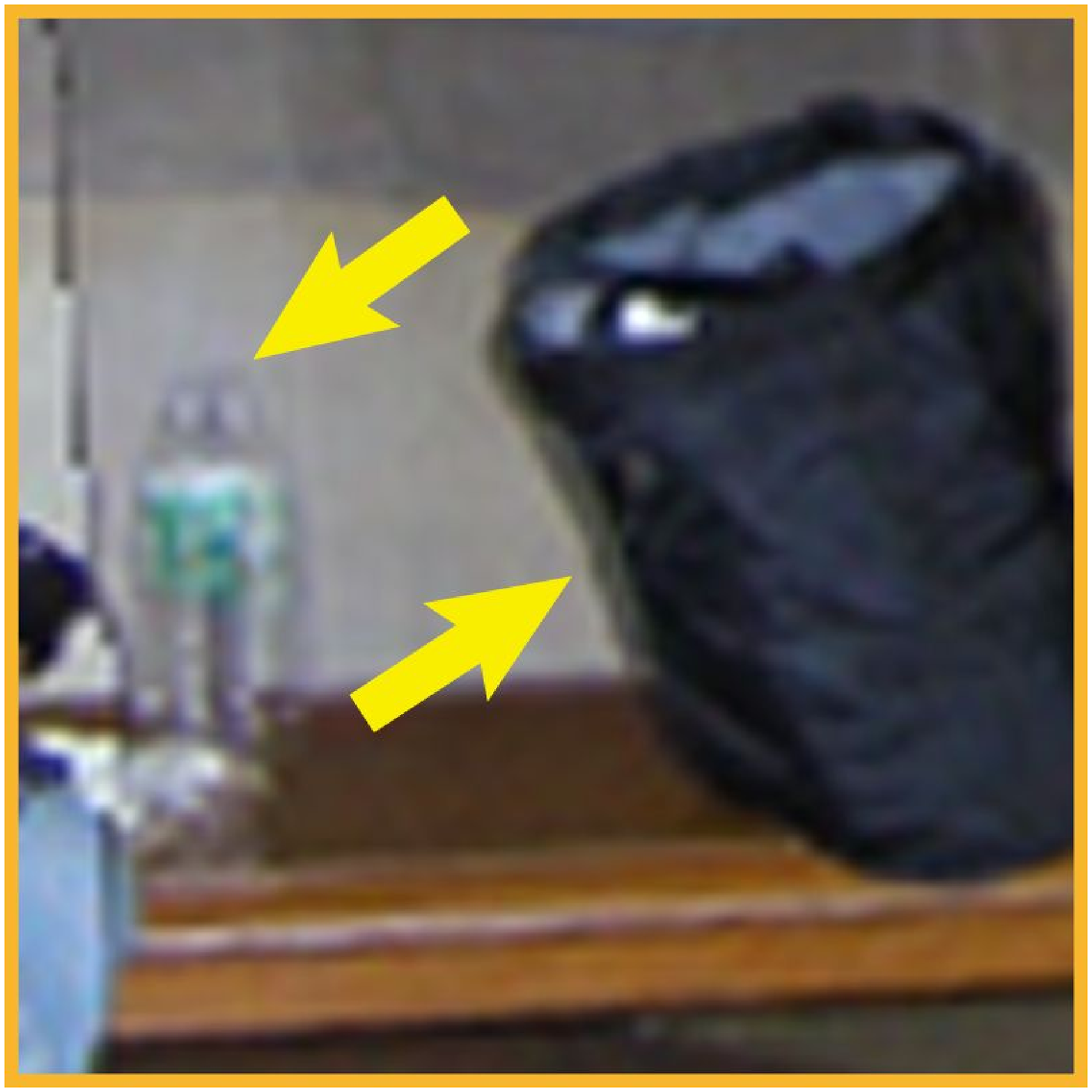} & \includegraphics[width=0.13\textwidth]{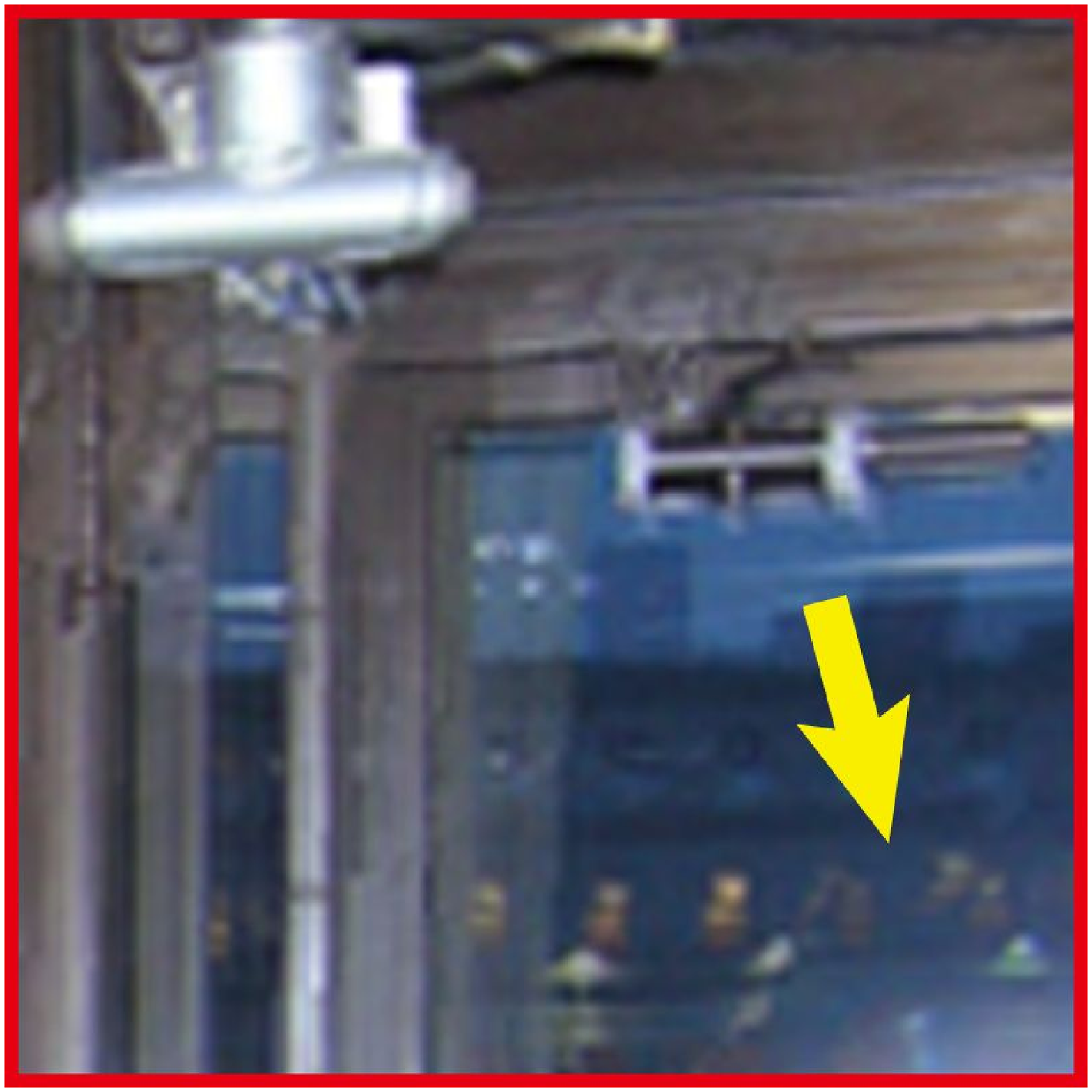}\\
 \rotatebox{90}{\small{LSH+MF-W}}&\includegraphics[width=0.195\textwidth]{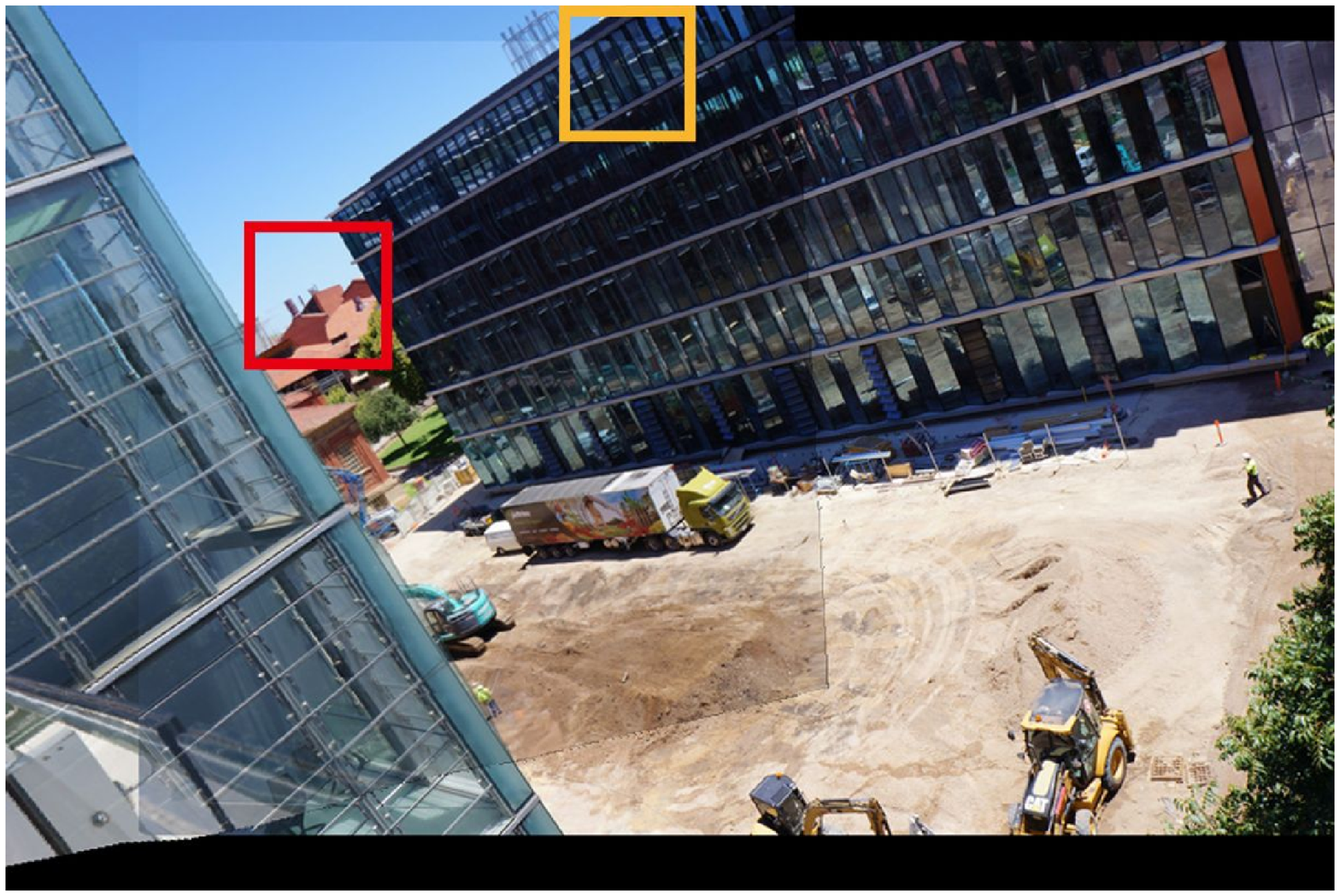}&\includegraphics[width=0.13\textwidth]{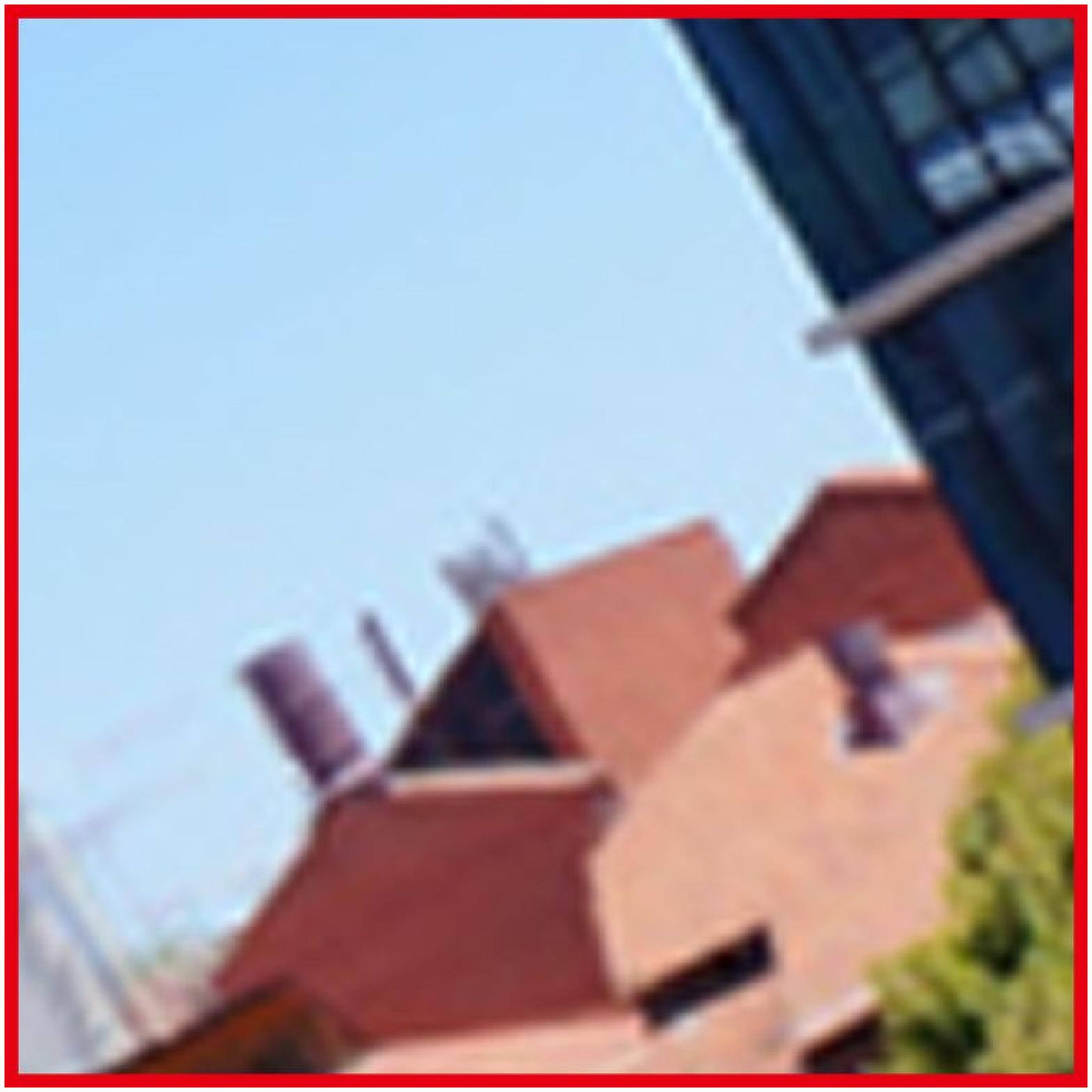}
      & \includegraphics[width=0.13\textwidth]{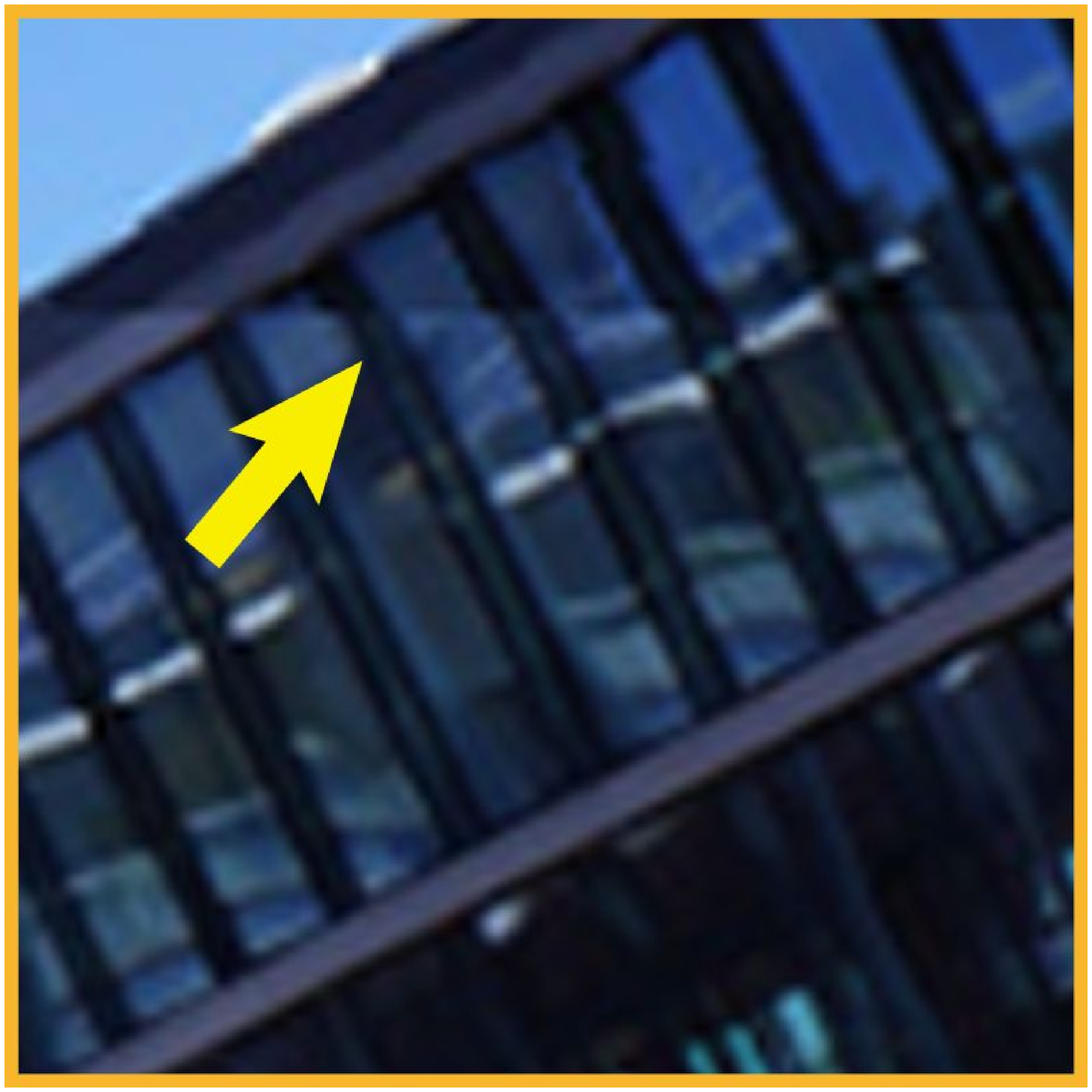} & & \includegraphics[width=0.195\textwidth]{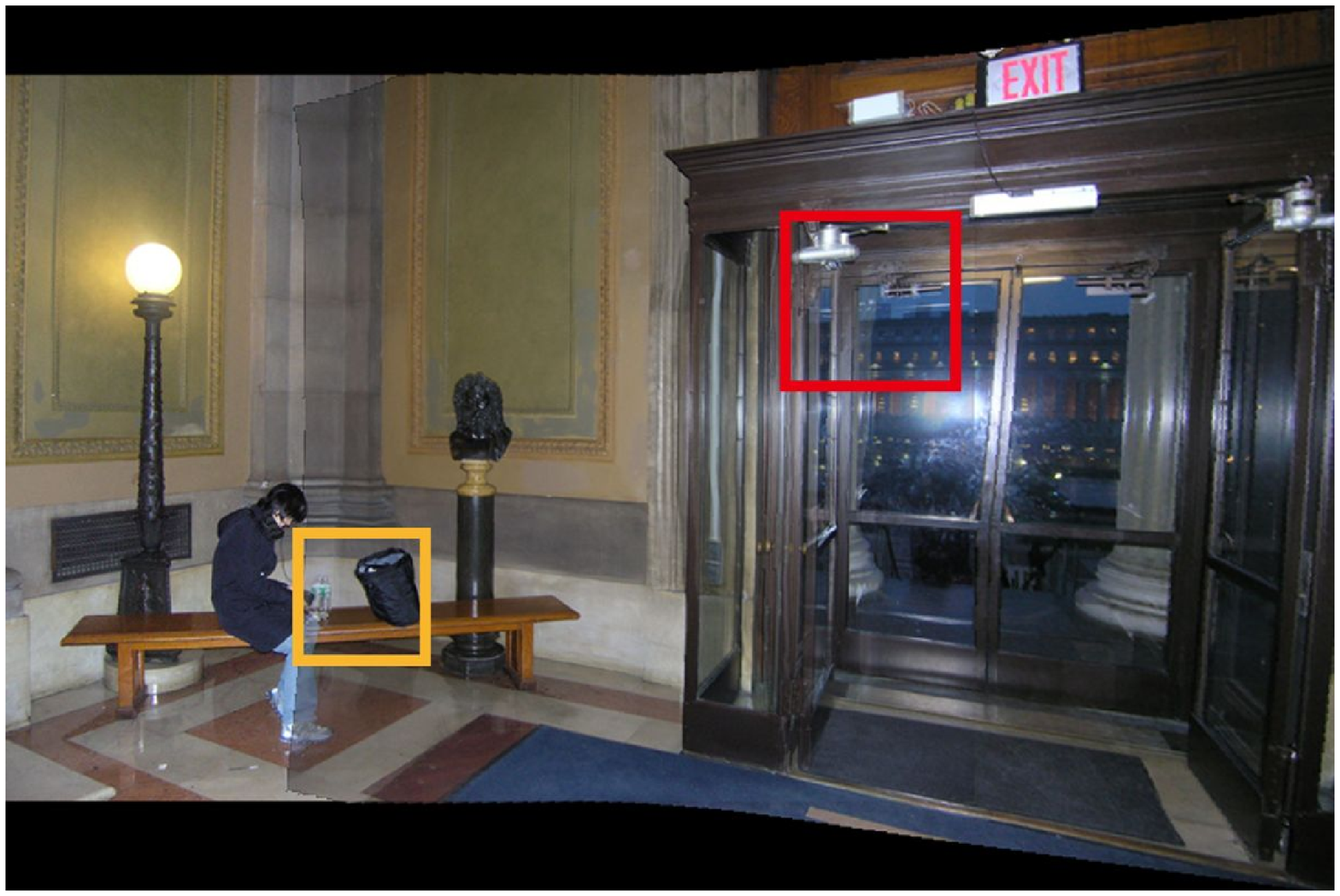}
      & \includegraphics[width=0.13\textwidth]{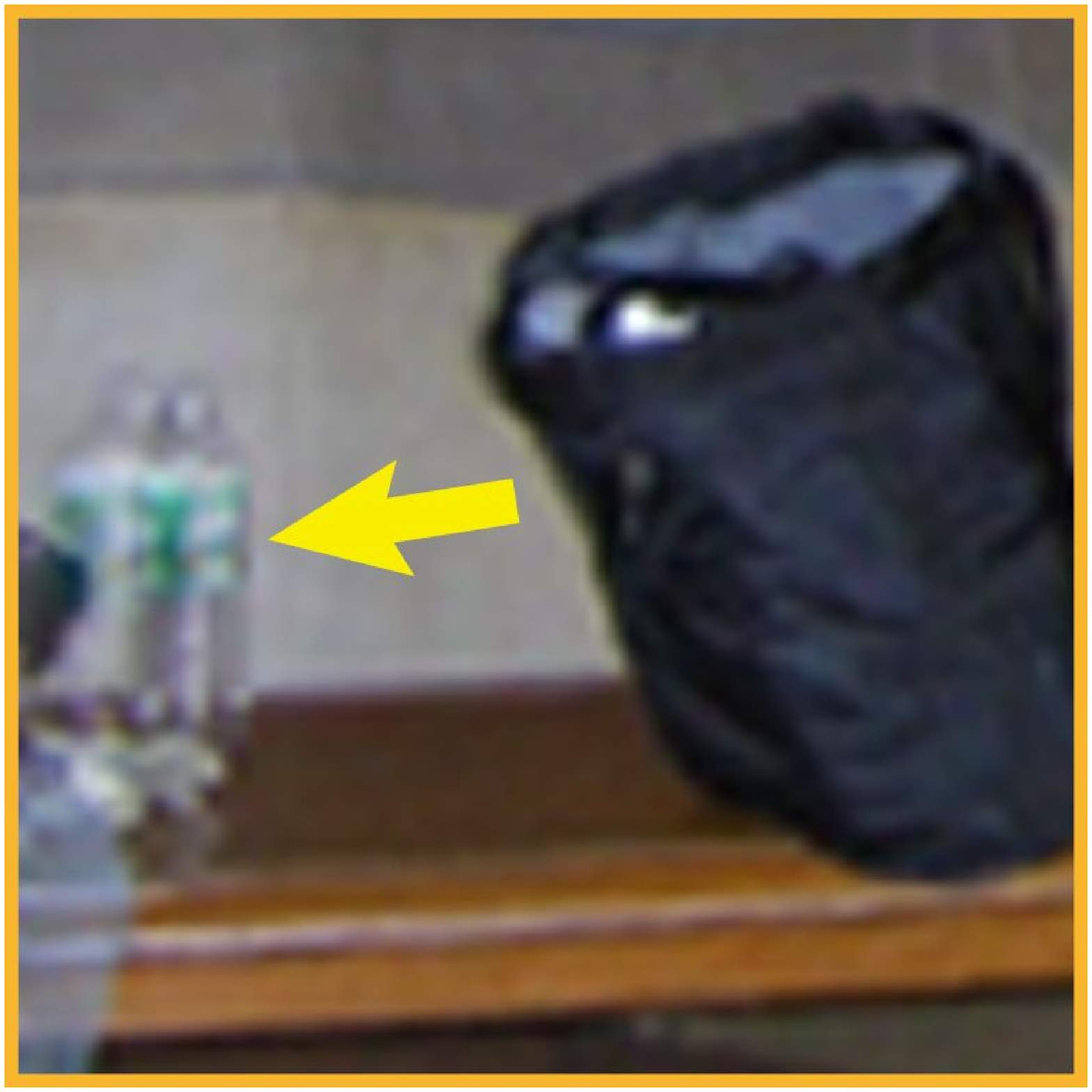} & \includegraphics[width=0.13\textwidth]{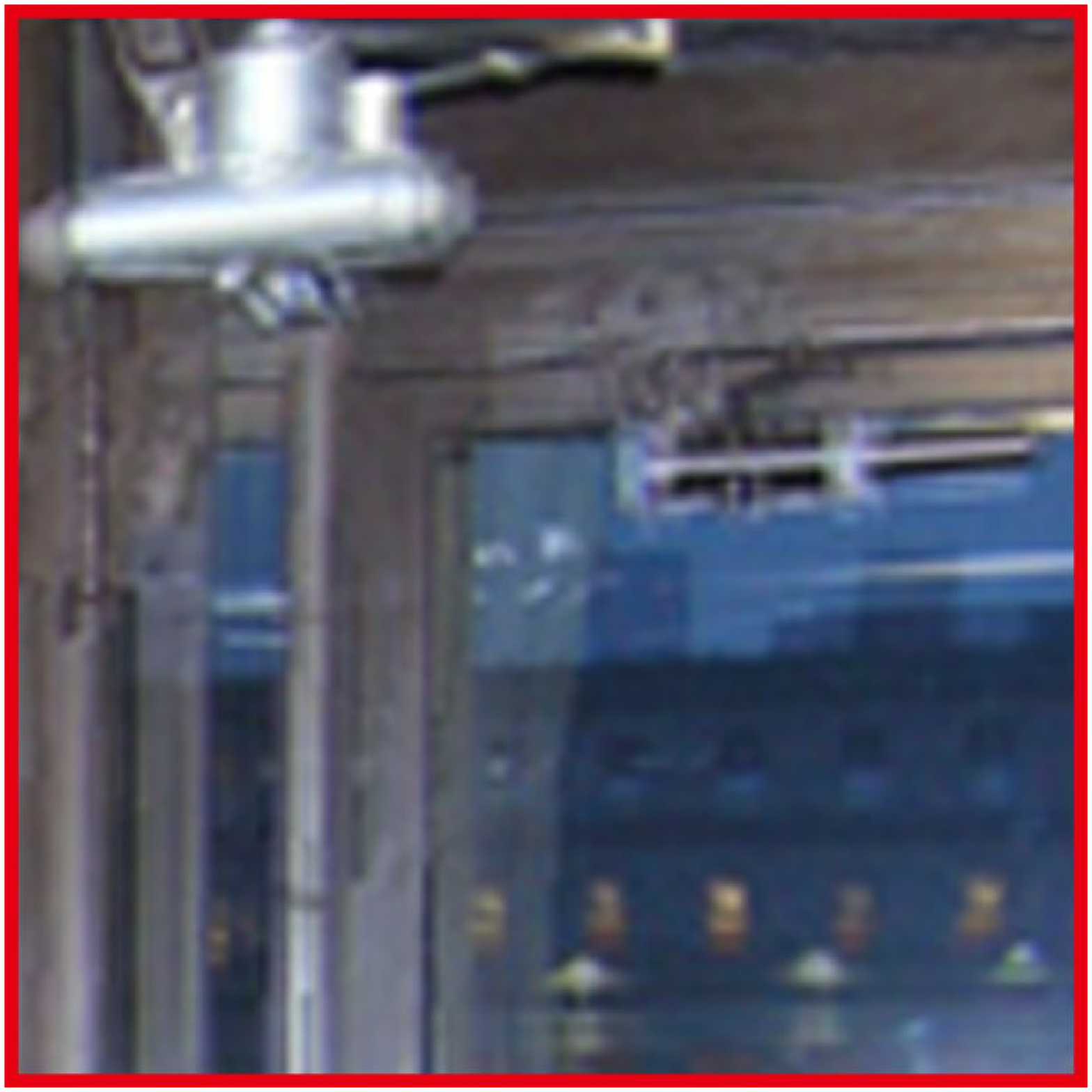}\\
 \rotatebox{90}{\small{ROS+MF-W}}&\includegraphics[width=0.195\textwidth]{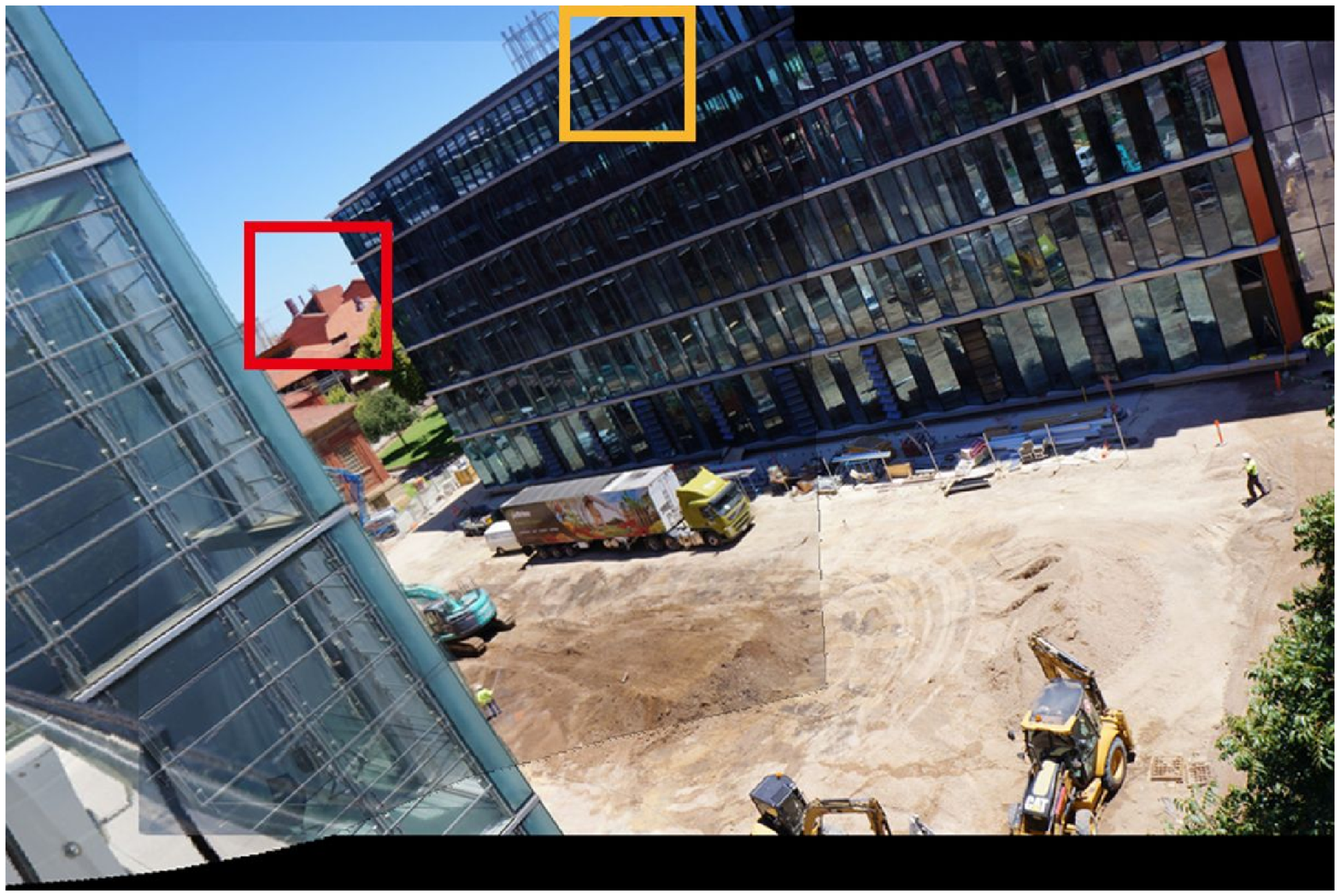}&\includegraphics[width=0.13\textwidth]{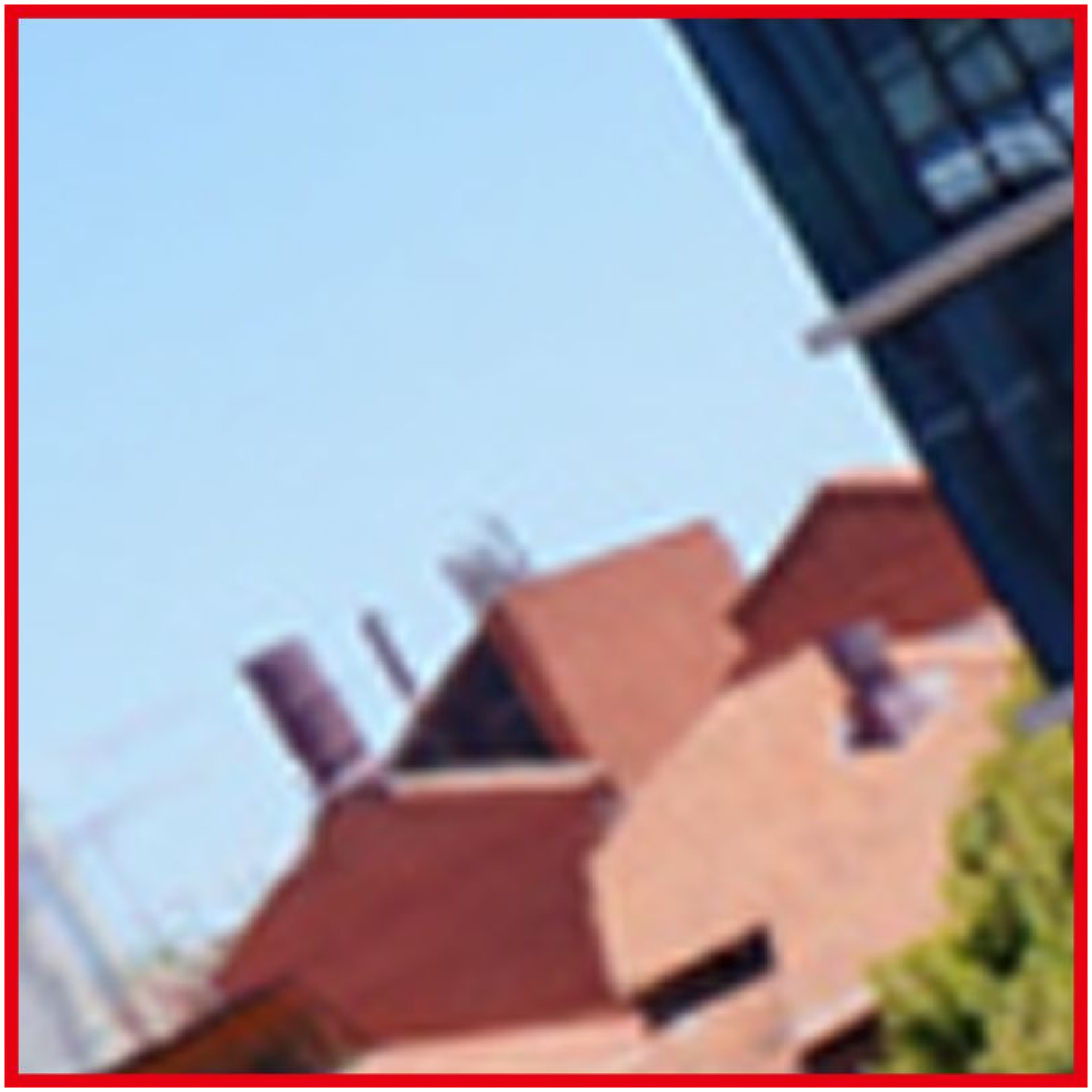}
      & \includegraphics[width=0.13\textwidth]{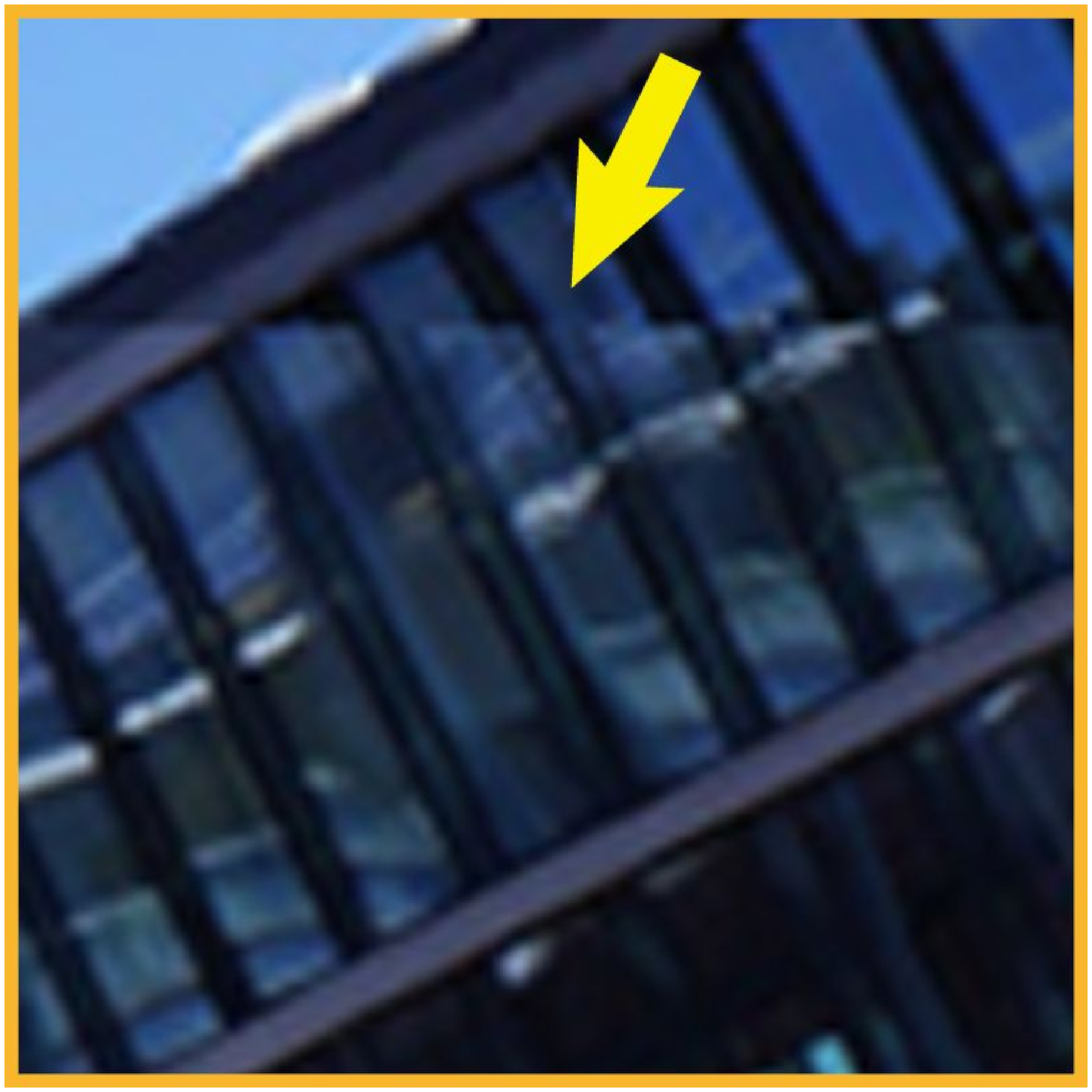} & & \includegraphics[width=0.195\textwidth]{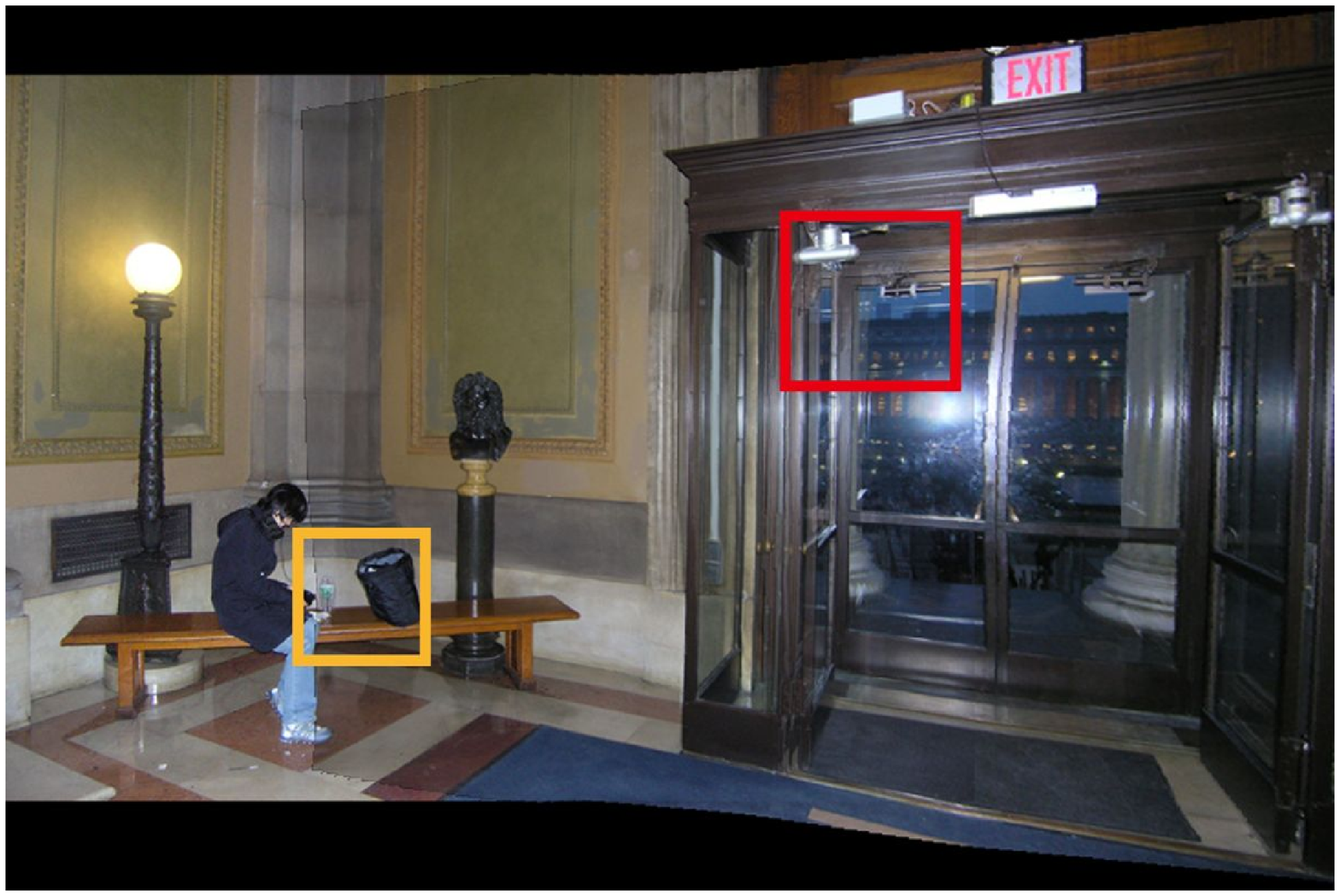}
      & \includegraphics[width=0.13\textwidth]{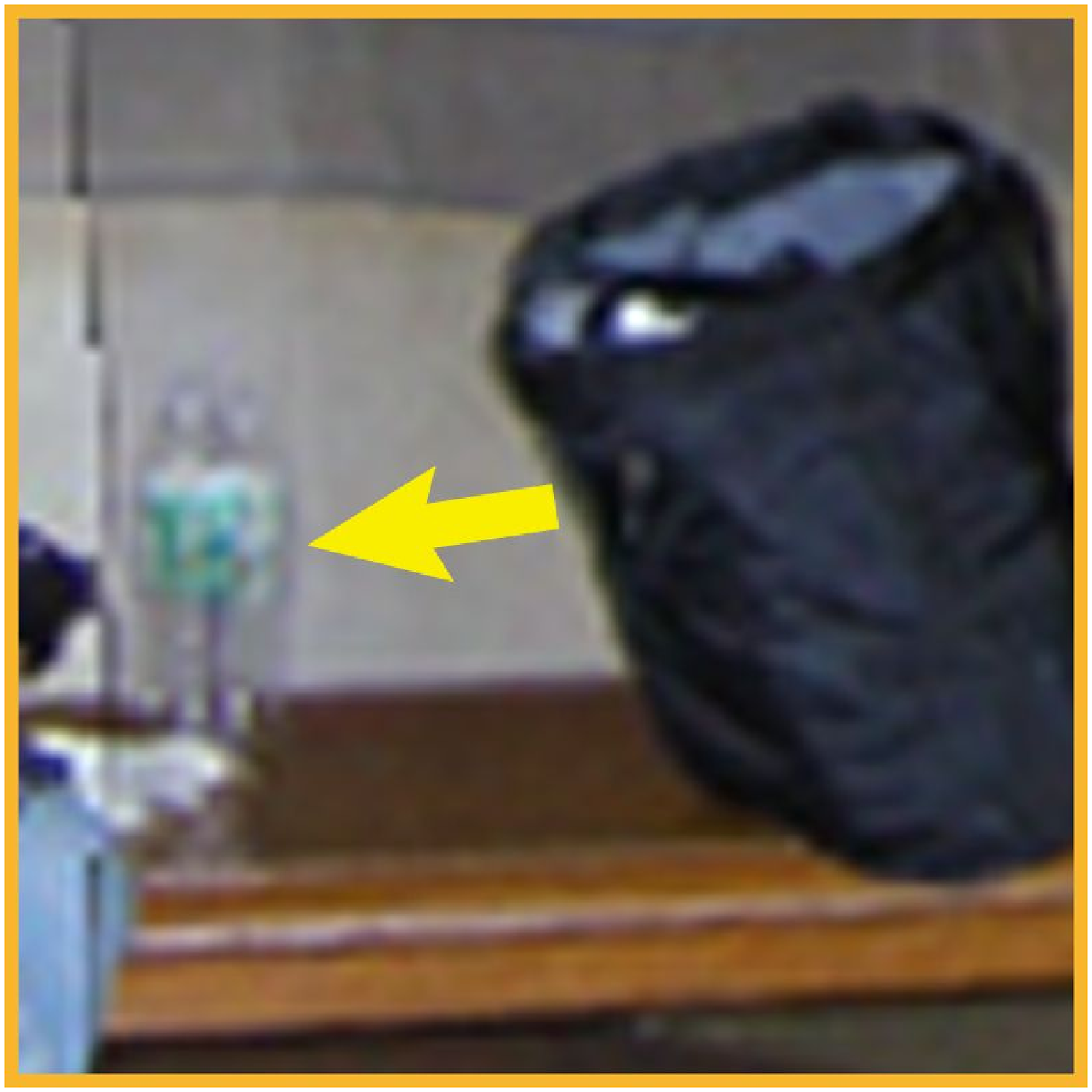} & \includegraphics[width=0.13\textwidth]{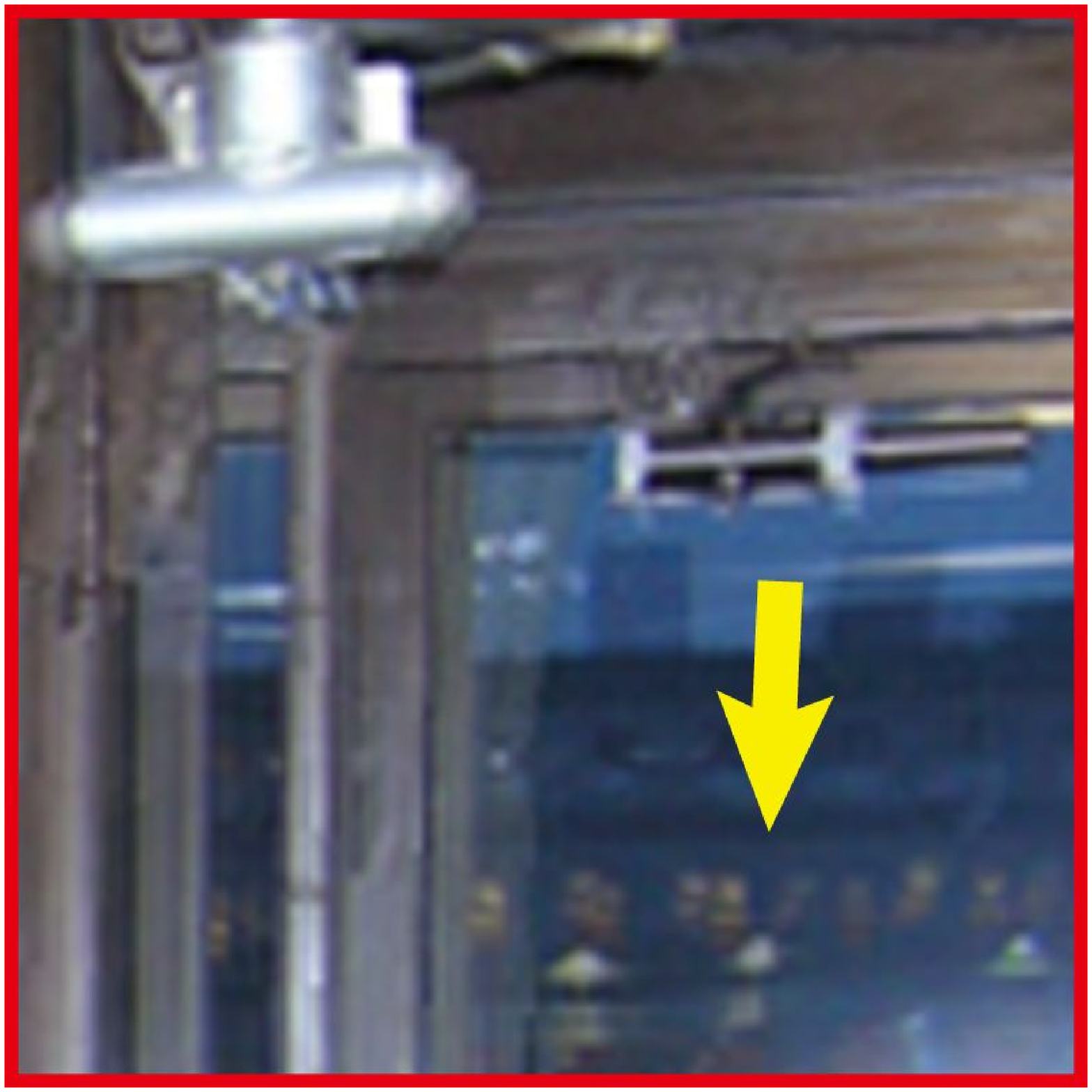}\\
 \rotatebox{90}{\small{GCPW}}  & \includegraphics[width=0.195\textwidth]{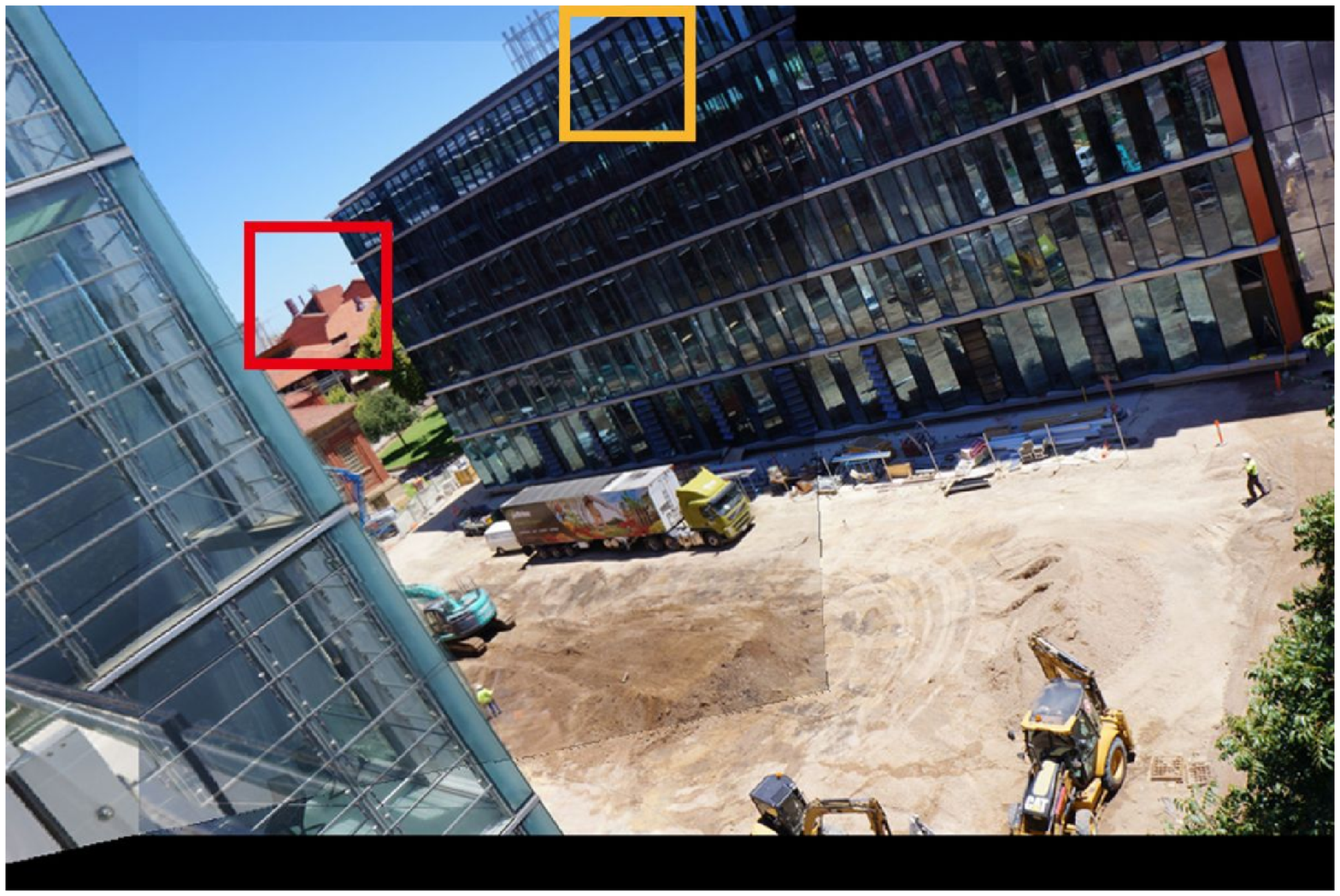}&\includegraphics[width=0.13\textwidth]{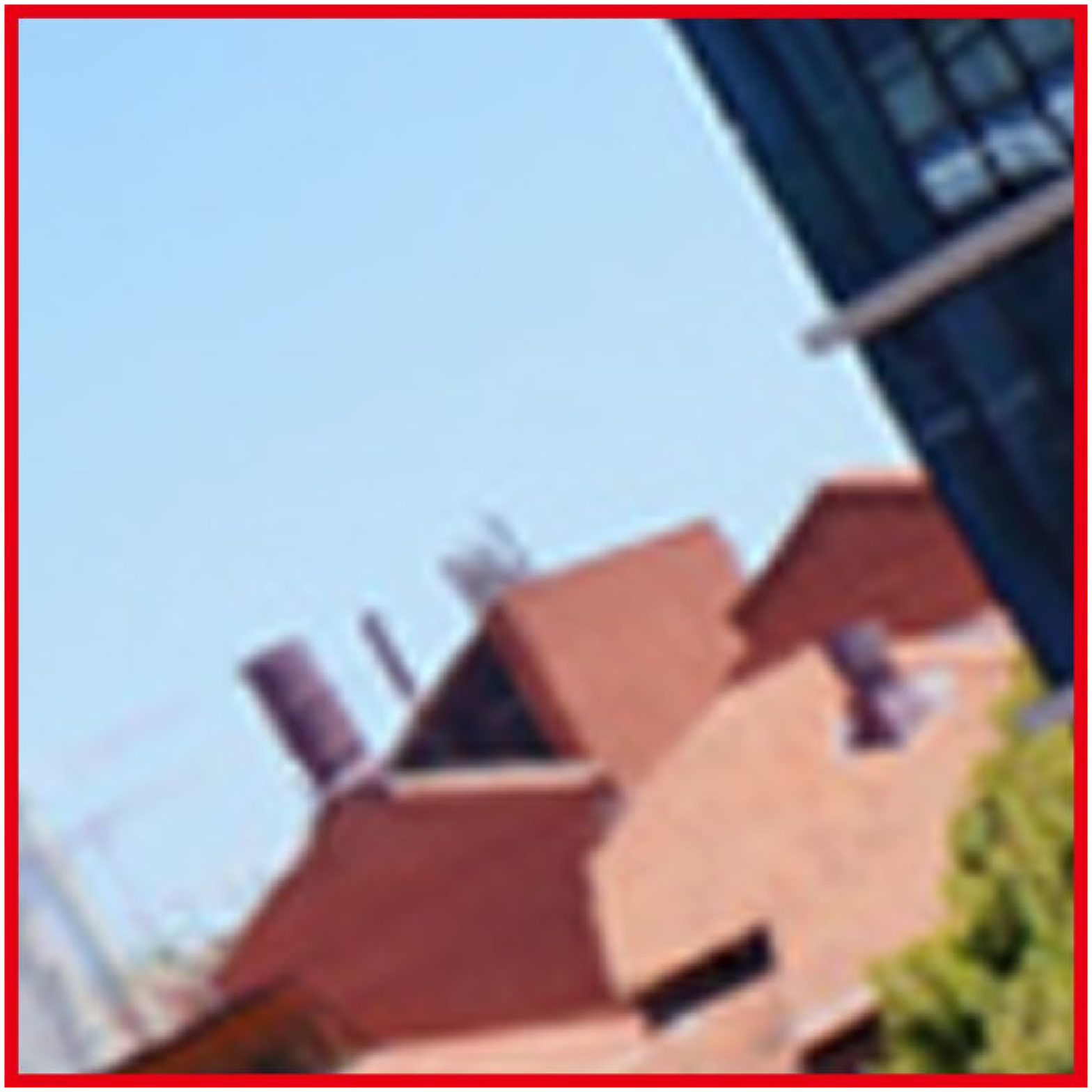}
      & \includegraphics[width=0.13\textwidth]{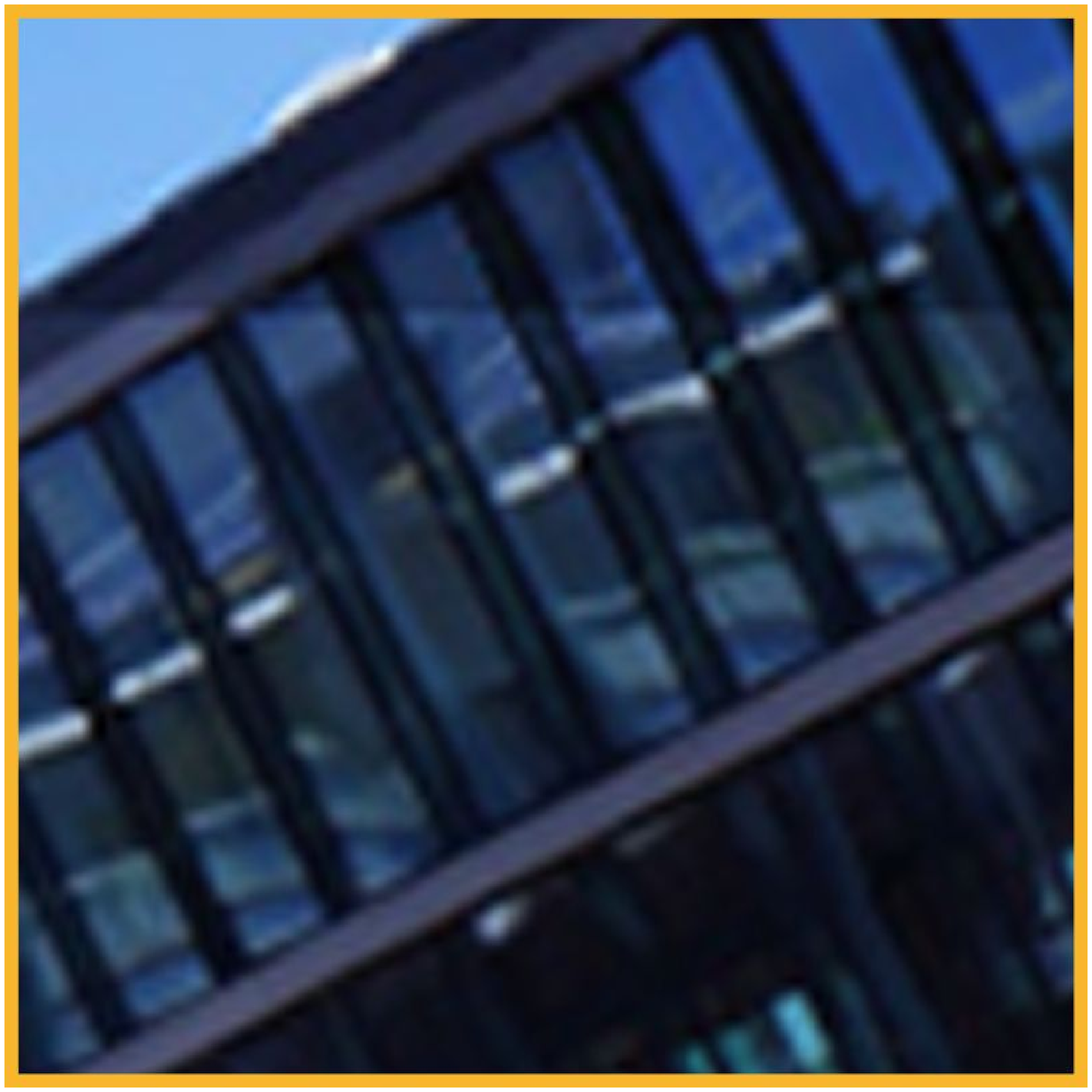} & & \includegraphics[width=0.195\textwidth]{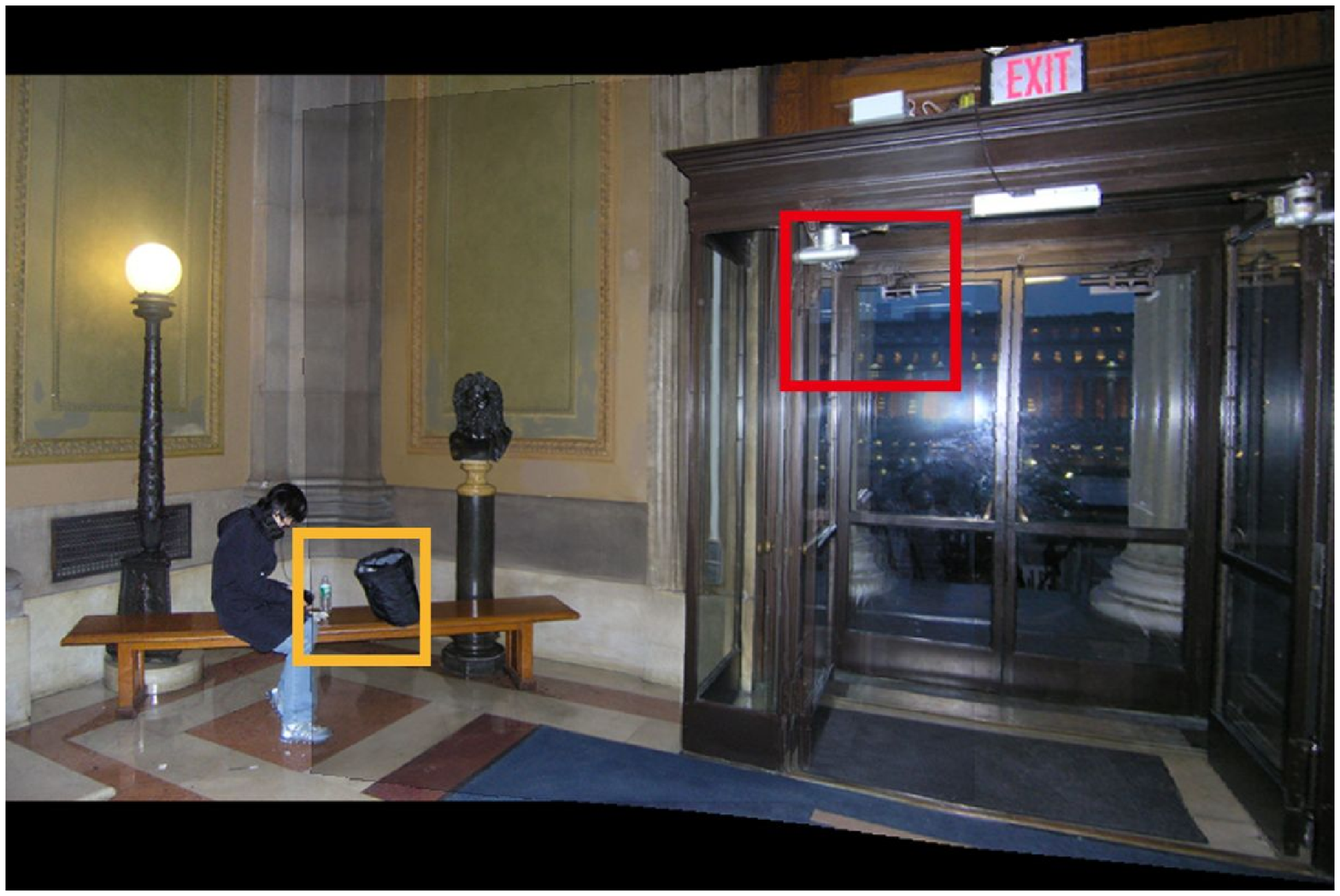}
      & \includegraphics[width=0.13\textwidth]{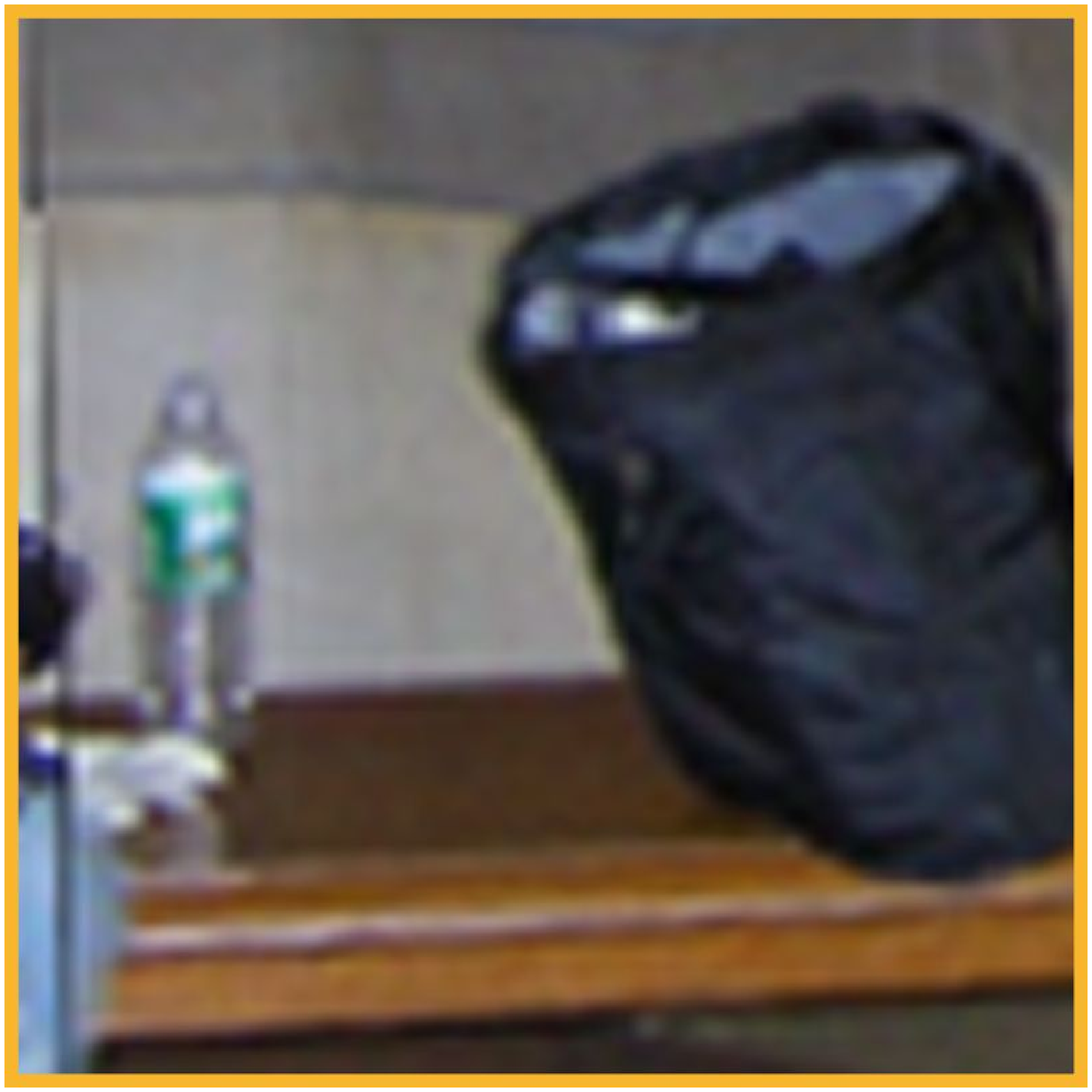} & \includegraphics[width=0.13\textwidth]{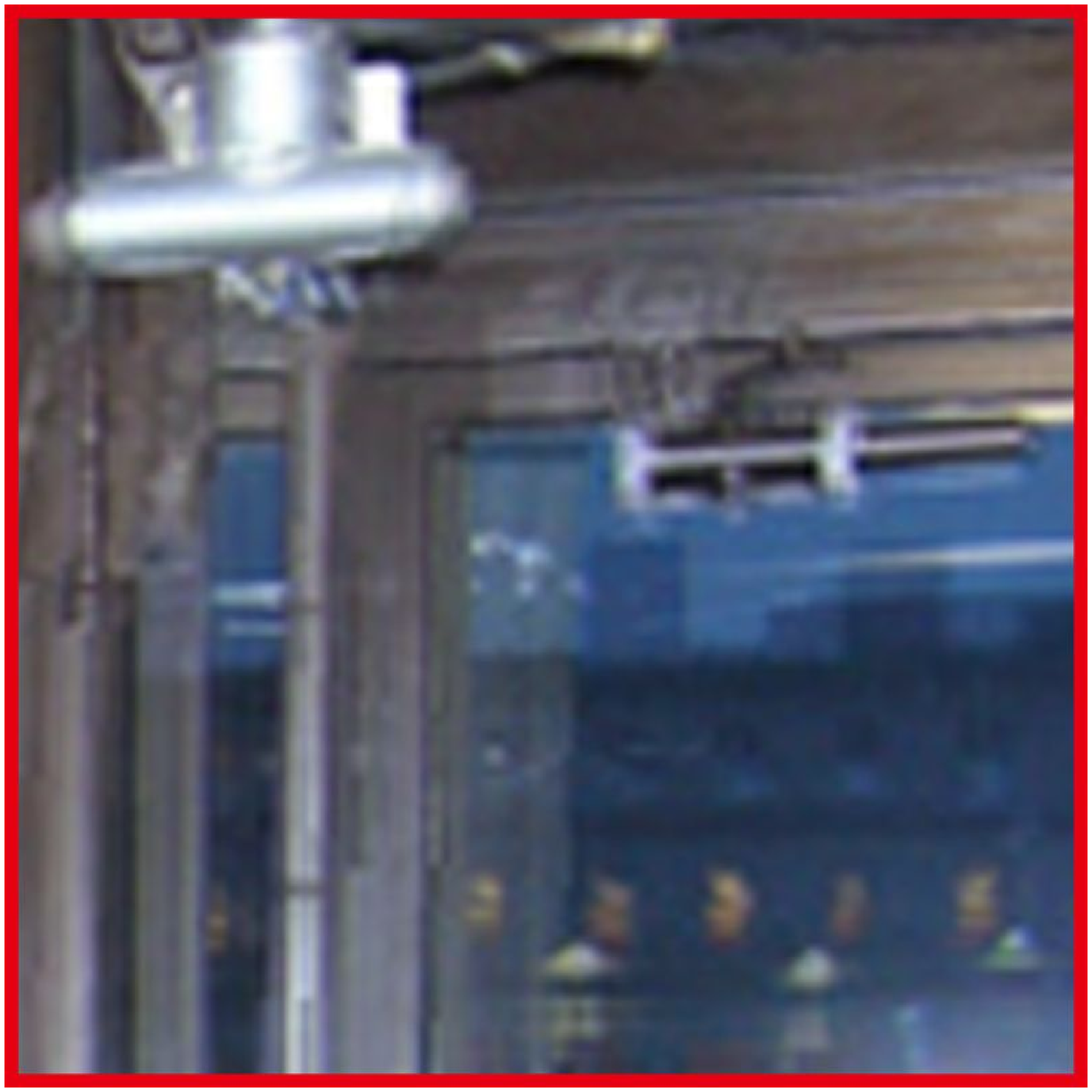}\\
\end{tabular}
\caption{Comparisons with MF-W, LSH+MF-W and ROS+MF-W on real images. Our method produces better stitching result than other three methods.}\label{fig:result_experimentB2}
\end{center}
\end{figure}

In order to further demonstrate the superiority of our method, Figure~\ref{fig:result_experimentB2} presents another two groups of result that compare the proposed GCPW with MF-W, LSH+MF-W, and ROS+MF-W. We can observe that in results produced by MF-W, LSH+MF-W and ROS+MF-W, there are severe ghosts and misalignments. The MF-W fails because it cannot cope with images with huge colour variations. Although LSH+MF-W and ROS+MF-W overcome this limitation through some pre-processing operations, nevertheless, they usually do not compensate the colour difference fully. Therefore, these two methods still suffer from severe misalignment artifacts. In contrast, our method does not adopt any pre-processing operation. It models the colour transformation in the GCPW framework and jointly estimates a local colour model and a optimal grid mesh at the same time. The results demonstrate that our method outperforms other methods when stitching images with significant colour variations. More comparative results are presented in our supplementary material.

\vspace{-15pt}
\section{Conclusion}
\label{sec:conclusion}

In this paper, we propose the GCPW framework to stitch images with colour variations. The GCPW extends the original CPW by appending a local colour model that expresses the colour transformation from the source image to the target image. This extension makes the photometric constraint robust to significant colour variations and thus can be applied to improve alignment quality. To stitch images with GCPW, we roughly align them with a global homography. Next, we combine geometric and photometric constraints in the proposed GCPW framework, in which the warped mesh vertexes and the affine colour model are optimized jointly. The images are aligned locally based on the warped grid mesh. We conducted experiments on both synthetic and real images. For synthetic experiment, images with different degrees of colour variation are tested, and results demonstrate that our method is robust to significant colour difference. For experiment on real images, although the colour model that applied in GCPW is simple, it is flexible enough to express complex colour transformations that occur in real world. Comparative results show that our method stably produces better stitching results than other state-of-the-art methods.

%
%
%

\bibliography{egbib}

\end{document}